\documentclass[9pt,a4paper,twoside]{tau}
\usepackage[english]{babel}
\usepackage{tauenvs}

\usepackage{subcaption}
\usepackage{makecell}
\usepackage[frozencache,cachedir=minted-cache]{minted}
\usepackage{xcolor} 
\usepackage{multirow}
\definecolor{lightgray}{gray}{0.95}

\setminted[json]{
  bgcolor=gray!10, 
  framesep=2mm 
}
\setminted[yaml]{
  bgcolor=gray!10, 
  framesep=2mm 
}

\setminted[python]{
  bgcolor=gray!10, 
}

\title{DSDL: Data Set Description Language \\ for Bridging Modalities and Tasks in AI Data}

\author[]{Bin Wang}
\author[]{Linke Ouyang} 
\author[]{Fan Wu} 
\author[]{Wenchang Ning} 
\author[]{Xiao Han}
\author[]{\\Zhiyuan Zhao}
\author[]{Jiahui Peng}
\author[]{Yiying Jiang}
\author[]{Dahua Lin}
\author[]{Conghui He\thanks{Corresponding author.}}

\affil[1]{Shanghai Artificial Intelligence Laboratory}

\professor{heconghui@pjlab.org.cn}

\institution{Shanghai Artificial Intelligence Laboratory}
\etal{Bin et al.}

\begin{abstract}
In the era of artificial intelligence, the diversity of data modalities and annotation formats often renders data unusable directly, requiring understanding and format conversion before it can be used by researchers or developers with different needs. To tackle this problem, this article introduces a framework called Dataset Description Language (DSDL) that aims to simplify dataset processing by providing a unified standard for AI datasets. DSDL adheres to the three basic practical principles of generic, portable, and extensible, using a unified standard to express data of different modalities and structures, facilitating the dissemination of AI data, and easily extending to new modalities and tasks. The standardized specifications of DSDL reduce the workload for users in data dissemination, processing, and usage. To further improve user convenience, we provide predefined DSDL templates for various tasks, convert mainstream datasets to comply with DSDL specifications, and provide comprehensive documentation and DSDL tools. These efforts aim to simplify the use of AI data, thereby improving the efficiency of AI development.
\end{abstract}

\begin{document}
		
    \maketitle\thispagestyle{firststyle}\tauabstract
    \doublespacing
    \tableofcontents
    \singlespacing


\newpage

\section{Introduction}
\taustart{I}n the era of artificial intelligence, large-scale, diverse, and high-quality datasets are indispensable for driving technological advancement. As more research institutions and individuals share their high-quality datasets~\cite{deng2009imagenet,lin2014microsoft,kuznetsova2020open,gupta2019lvis,schuhmann2022laion,raffel2020exploring,lin2023parrot,changpinyo2021conceptual,wang2024unimernet,he2023wanjuan}, deep learning~\cite{krizhevsky2017imagenet,he2016deep,vaswani2017attention} and large model~\cite{devlin2018bert,radford2018improving,radford2019language,brown2020language,liu2024visual,zhang2023internlm,dong2024internlm,chen2024far,wang2024vigc} technologies have seen significant enhancements. However, the diversity of modalities and tasks in datasets, along with different representations of the same content, has led to exceedingly complex data formats. When dealing with multimodal and multitask model training, researchers and developers often need to undergo cumbersome format conversions, not only increasing preprocessing workloads but also potentially leading to data storage waste. For instance, different research institutions or even the same institution might store multiple copies of data due to different format conversion methods.

To address the complexity of AI datasets, we propose a new approach: using the concept of programming languages to simplify the description and utilization of datasets. Under this approach, although different researchers might use varying styles and naming conventions, they all follow the same language specification to describe their datasets. This unified approach ensures that dataset parsing and usage are simplified and standardized, regardless of the individual styles and conventions used. Based on this idea, this paper introduces the Data Set Description Language (DSDL)—a framework designed to simplify dataset handling by providing a unified standard for AI datasets, significantly reducing the workload involved in data retrieval, preprocessing, and usage.

The core design principles of DSDL include being generic, portable, and extensible, ensuring its broad applicability and ease of use across different environments. By adopting widely used data exchange formats such as YAML and JSON, DSDL not only enhances compatibility but also facilitates the integration of datasets from various sources. This flexibility and user-friendliness make DSDL a powerful tool for simplifying dataset handling.

The main contributions of this paper are as follows:
\begin{itemize}
    \item We introduce DSDL, a standardized language that provides a unified description for multitask and multimodal AI datasets. It allows users to define different modalities and types of tasks according to a unified specification, and even stylistically diverse DSDL descriptions can be accurately parsed by the DSDL parser.
    
    \item  We provide predefined DSDL templates for individuals and institutions wishing to publish data using DSDL, enabling users to quickly standardize dataset descriptions with only a basic understanding of DSDL syntax.

    \item To reduce users' learning curves, we offer mainstream datasets described according to DSDL specifications on the OpenDataLab platform~\cite{conghui2022opendatalab}, allowing users to download and utilize these datasets directly.
    
    \item We also offer comprehensive tools~\footnote{dsdl-sdk: https://github.com/opendatalab/dsdl-sdk} and documentation~\footnote{dsdl-docs: https://opendatalab.github.io/dsdl-docs/}, supporting users in visualizing, training, inferring, and annotating datasets without needing to concern themselves with the underlying logic.
    
\end{itemize}    
\section{DSDL Overview}

Data is the cornerstone of artificial intelligence. The efficiency of data acquisition, exchange, and application directly impacts the advances in technologies and applications. Over the long history of AI, a vast quantity of data sets have been developed and distributed. However, these datasets are defined in very different forms, which incurs significant overhead when it comes to exchange, integration, and utilization -- it is often the case that one needs to develop a new customized tool or script in order to incorporate a new dataset into a workflow.

To overcome such difficulties, we develop \textbf{Data Set Description Language (DSDL)}.

\subsection{Design Goals}

The design of \textbf{DSDL} is driven by three goals, namely generic, portable, extensible. We refer to these three goals together as \textbf{GPE}.

\noindent\textbf{Generic.}This language aims to provide a unified representation standard for data in multiple fields of artificial intelligence, rather than being designed for a single field or task. It should be able to express data sets with different modalities and structures in a consistent format.

\noindent\textbf{Portable.}
Write once, distribute everywhere.

Dataset descriptions can be widely distributed and exchanged, and used in different environments without modification of the source files. The achievement of this goal is crucial for creating an open and thriving ecosystem. To this end, we need to carefully examine the details of the design, and remove unnecessary dependencies on specific assumptions about the underlying facilities or organizations.

\noindent\textbf{Extensible.}
One should be able to extend the boundary of expression without modifying the core standard. For a programming language such as C++ or Python, its application boundaries can be significantly extended by libraries or packages, while the core language remains stable over a long period. Such libraries and packages form a rich ecosystem, making the language stay alive for a very long time.

\subsection{Design Overview}

A data set is essentially a data structure stored in persistent storages. In general, it comprises unstructured objects, e.g. images, videos, and texts, together with associated annotations. Such elements are aggregated in certain ways into a data set.

It is noteworthy that the unstructured objects mentioned above usually contain large volume of data. To facilitate quick distribution of data sets, our design separates the structured description of a dataset from the content of unstructured objects.

Below is an overall summary of the language design:

\noindent\textbf{Basic data model.}
DSDL describes a data set with a collection of basic elements organized via containers such as structs, lists, and sets.

\begin{itemize}
    \item \textbf{Basic elements} are individual units in a data set description, which include not only the primitives such as numbers, strings, but also those elements that facilitate the expression of object locations, annotations, etc.
    \item \textbf{Unstructured objects} such as images, videos, and texts, are special basic elements, as they are indivisible in a data set description. In particular, an unstructured object is represented by an object locator which tells where it is stored instead of being embedded into the description entirely. Optionally, additional descriptors can be used to provide additional information about the object, e.g. the format or resolution of an image.
    \item \textbf{Aggregates} are used to organize basic elements into a data structure. DSDL provides list and struct types to express aggregate data structures. In particular, individual samples are represented by a struct (an aggregate of multiple fields), while a data set consists of a list of samples.
    
\end{itemize}

\noindent\textbf{Extensible type system.}
All elements and structured units in DSDL have types. DSDL adopts a simple yet extensible type systems. Specifically, there are three kinds of types in DSDL:
\begin{itemize}
    \item \textbf{Primitive types} are the types of primitive values such as booleans, numbers, and strings. DSDL provides a large collection of primitive types, which serve as the basic building blocks of the description. Note that different primitive types in DSDL can be expressed in the same form. For example, object locators and time stamps both use strings as their expression form, but the string values will be interpreted diffferently depending on the underlying types.

    \item \textbf{Unstructured object classes} are the abstractions for unstructured objects, such as images, videos, audios, point clouds, texts, etc. Such objects, despite their rich internal structures, are considered as indivisible units in data set definitions. DSDL provides a collection of pre-defined unstructured object classes to cover common applications, while allowing 3rd parties to extend this collection by registering new unstructured object classes via a minimal set of interfaces.

    \item \textbf{Struct classes} are the abstractions for aggregate data structures in DSDL. Each instance of a \textbf{struct} class is called a struct, which contains multiple fields, each with its own type. An important application of structs are to represent data samples. DSDL comes with a collection of predefined struct classes for commonly seen tasks in the standard library, while allowing users to define their own struct classes for special tasks.
\end{itemize}

\begin{tcolorbox}[title=Note]
Types need to be defined (either builtin, by 3rd parties, or by the user) before they are used. Circular references are not allowed in this version of DSDL.
\end{tcolorbox}

\noindent\textbf{Object locators.}
As mentioned, unstructured objects are not embedded entirely into the data set description. Instead, they are referred to by object locators. In particular, an object locator is a string with special format, which will be converted into an actual address by the DSDL interpreter when the corresponding object is to be loaded.

The introduction of object locators is the key to separating the structured description of the data set from the unstructured media content. This way not only enables light-weight distribution of data set descriptions without moving the large volume of media data, but also allows quick manipulation of a data set, e.g. combining multiple sets, merging properties, or taking a subset.

\noindent\textbf{Based on JSON or YAML.}
DSDL is a \textbf{domain-specific language} based on popular data exchange languages: \href{https://www.json.org/json-en.html}{JSON} or \href{https://yaml.org/}{YAML}. Note that the elements in JSON or YAML  are not associated with specific meanings at the language level. By endowing such elements with semantics, DSDL can describe a data set in a meaningful manner.

This design choice allows one to leverage the rich tool systems already available for JSON and YAML. With such tools, one can readily build a full-fledged system that fully supports interpretation, validation, and query, and as well as the interoperability with the Internet ecosystem.

\subsection{DSDL Core Architecture}

\begin{figure}[H]
    \centering
    \includegraphics[width=0.55\textwidth]{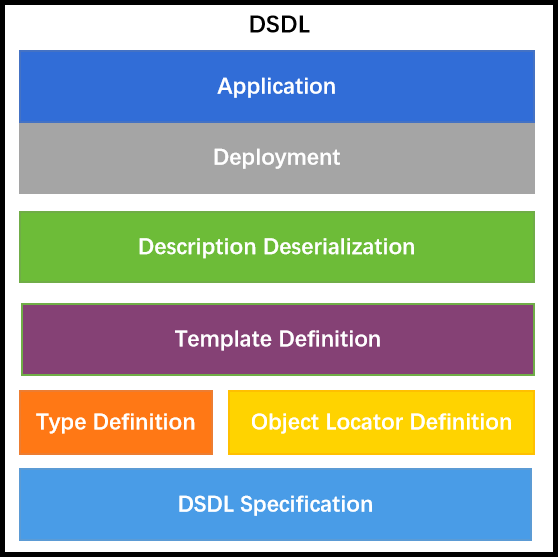}
    \caption{Schematic of the DSDL Framework}
    \label{fig:dsdl_arch}
\end{figure}

As illustrated in the~\cref{fig:dsdl_arch}, DSDL adopts a bottom-up overall architecture. We first defined the DSDL Specification, which serves as the foundation for syntax rules. Building upon this, we further defined DSDL data types and the Object Locator. Utilizing these building blocks, researchers can design dataset description files suitable for their AI tasks. The introduction of the Object Locator greatly simplifies the reading and usage of data from multiple sources, whether the data is stored locally, in a cluster, or in the cloud. To further facilitate researchers, we have predefined data templates for mainstream tasks to simplify the use of datasets. A comprehensive syntax parser can convert DSDL descriptions into Python objects, thereby realizing the full functionality of the dataset description language. In addition, we have developed a set of DSDL toolkits, so that users can easily download mainstream datasets and their DSDL description files through the OpenDataLab CLI, further simplifying data visualization and model training processes.

\subsection{DSDL Target Users}

DSDL serves AI researchers and developers at different levels:

\begin{itemize}
\item \textbf{AI Beginners}: They can quickly understand and use mainstream datasets without delving into data formats and content. DSDL provides clear metadata and annotation information.
\item \textbf{Researchers in Specific Fields}: DSDL provides a unified dataset function interface, simplifying the use of multiple datasets.
\item \textbf{Large Model Researchers}: They can efficiently combine and process a large amount of data from different tasks and modalities through DSDL, accelerating the model training and validation process.
\end{itemize}
\section{DSDL Specification}

\subsection{Get Started}

In DSDL, a data set is described by a data set description file. Below is an example that illustrates a typical data set description file. The data set description file can be in either JSON or YAML format.

\noindent\textbf{JSON Format:}
\begin{minted}{json}
  {
        "$dsdl-version": "0.5.0",
        "meta": {
            "name": "my-dataset",
            "creator": "my-team",
            "dataset-version": "1.0.0"
        },
        "defs": {
            "MyClassDom": {
                "$def": "class_domain",
                "classes": [
                    "dog",
                    "cat",
                    "fish",
                    "tiger"
                ]
            },
            "ImageClassificationSample" : {
                "$def": "struct",
                "$fields": {
                    "image": "Image",
                    "label": "Label[dom=MyClassDom]"
                }
            }
        },
        "data": {
            "sample-type": "ImageClassificationSample",
            "samples": [
                { "image": "xyz/0001.jpg", "label": "cat" },
                { "image": "xyz/0002.jpg", "label": "dog" }
            ]
        }
    }
\end{minted}

\noindent\textbf{YAML Format:}
\begin{minted}{yaml}
    $dsdl-version: "0.5.0"
    meta:
        name: "my-dataset"
        creator: "my-team"
        dataset-version: "1.0.0"
    defs:
        MyClassDom:
            $def: class_domain
            classes:
                - dog
                - cat
                - fish
                - tiger
        ImageClassificationSample:
            $def: struct
            $fields:
                image: Image
                label: Label[dom=MyClassDom]
    data:
        sample-type: ImageClassificationSample
        samples:
            - { image: "xyz/0001.jpg", label: "cat" }
            - { image: "xyz/0002.jpg", label: "dog" }
\end{minted}

Both JSON and YAML formats can express exactly the same data structure. Due to YAML's more concise form and that it allows comments, we will use YAML as the default format in later examples of this document, which can be easily translated into JSON format.

At the top level, the file consists of four parts:

- \textbf{header} specifies about how the description file should be interpreted;

- \textbf{meta section} provides meta information about the data set;

- \textbf{defs section} provides global definitions, e.g. user-defined types;

- \textbf{data section} describes the data contained in the data set.

\begin{tcolorbox}[title=Note]
- The property names with a prefix \$ are reserved by DSDL for special meaning.

- DSDL version (the property with name \$dsdl-version in the header) must be explicitly specified. It is crucial for the DSDL interpreter to know the language version in order to interpret the description correctly.

- The definition for common types are often provided in standard or extended libraries. In most cases, users don't need to define their own types. In this example, we define ImageClassificationSample just for the purpose of illustration and being self-contained.

\end{tcolorbox}

\subsection{Basic Types}
\textbf{Basic types} are the types for basic elements in DSDL. The instances of basic types serve as the basic building blocks of a data set description.

\begin{tcolorbox}[title=Note]
The underlying language, namely JSON and YAML, provides several primitive literals, such as boolean, number, string. While using such literals to express values, DSDL maintains its own basic types. It is important to note that association between DSDL basic types and JSON primitive types is \textbf{NOT} one-to-one. Different DSDL basic types can adopt the same primitive type for expressing their values. For example, object locators and labels are of different types (Loc and Label in this example) in DSDL, but they both use strings for expressing values.
\end{tcolorbox}

\subsubsection{Generic basic types}

DSDL defines four generic basic types. The values of such types are simply interpreted, without special meaning.
\begin{itemize}
    \item \colorbox{lightgray}{Bool}: boolean type, which can take either of the two values: \colorbox{lightgray}{true} and \colorbox{lightgray}{false}.

    \item \colorbox{lightgray}{Int}: integer type, which can take any integral values, such as \colorbox{lightgray}{12}, \colorbox{lightgray}{-3}, or \colorbox{lightgray}{0}. When a number has the \colorbox{lightgray}{Int} type, the DSDL interpreter should verify if it is actually an integer.

    \item \colorbox{lightgray}{Num}: general numeric type, which can take any numeric values, such as \colorbox{lightgray}{12.5}, \colorbox{lightgray}{-13}, \colorbox{lightgray}{1.25e-6}.
    
    \item \colorbox{lightgray}{Str}: string type, which can take arbitrary strings, such as \colorbox{lightgray}{"hello"}, \colorbox{lightgray}{"a"}, \colorbox{lightgray}{""}.
\end{itemize}

\subsubsection{Special basic types}

DSDL defines a collection of basic types with special meanings. The values of these types are also expressed as strings or other common JSON forms, but they have specific semantics and DSDL interpreter will interpret them accordingly.
\begin{itemize}
    \item \colorbox{lightgray}{Coord}: 2D coordinate in the form of \colorbox{lightgray}{[x, y]}.
    \item \colorbox{lightgray}{Coord3D}: 3D coordinate in the form of\colorbox{lightgray}{[x, y, z]}.
    \item \colorbox{lightgray}{Interval}: sequential interval in the form of \colorbox{lightgray}{[begin, end].}
    \item \colorbox{lightgray}{BBox}: bounding box in the form of \colorbox{lightgray}{[x, y, w, h]}.
    \item \colorbox{lightgray}{Polygon}: polygon represented in the form of a series of 2D coordinates as \colorbox{lightgray}{[[x1, y1], [x2, y2], ...]}.
    \item \colorbox{lightgray}{Date}: date represented by a string, according to the \href{https://strftime.org/}{strftime spec}.
    \item\colorbox{lightgray}{Time}: time represented by a string, according to the \href{https://strftime.org/}{strftime spec}.
\end{itemize}

\subsubsection{Label: class label type}

Classification is a common way to endow an object with semantic meaning. In this approach, \textbf{class labels} are often used to express the category which an object belongs to. In DSDL, class labels are strings with type \colorbox{lightgray}{Label}.

In practice, labels in different classification domains are different. DSDL introduces the concept of \textbf{class domain} to represent different contexts for classification. Each class domain provides a class list or a class hierarchy. Given a class domain, the labels are can be expressed in either of the following two forms:
\begin{itemize}
    \item \textbf{name-based}: with format \colorbox{lightgray}{"<class-domain>::<class-name>"}, e.g. \colorbox{lightgray}{"COCO::cat"} represents the class cat in the COCO domain.
    \item \textbf{index-based}: with format \colorbox{lightgray}{"<class-domain>[class-index]"}, e.g. \colorbox{lightgray}{"COCO[3]"} represents thr 3rd class in the COCO domain.
\end{itemize}

For a class domain with a multi-level class hierarchy, the class label can be expressed as a dot-delimited path, such as  \\ \colorbox{lightgray}{"MyDom::animal.dog.hound"} or \colorbox{lightgray}{"MyDom[3.2.5]"}.

\begin{tcolorbox}[title=Note]
We are working on unifying class systems for specific areas. The efforts would result in a standard classification domain. We reserve the domain name \textbf{std} for this.
\end{tcolorbox}

\subsubsection{Loc: object locator type}

Object locators are used as references to unstructured objects, such as images, videos, and texts. They are instances of the type Loc, and are represented by a specially-formatted string. Specifically, DSDL supports three ways to express an object locator:

\begin{itemize}
    \item \textbf{relative path}: the path relative to the root data path. This is the default way. When there is no special prefix, an object locator string will be treated as a relative path. For example, \colorbox{lightgray}{"abc/001.jpg"} will be interpreted as \colorbox{lightgray}{"<data-root>/abc/001.jpg"}, where data-root is the root directory where all data objects are stored and can be specified via environment configurations.

    \item \textbf{alias path}: when a data set comprises data objects stored in multiple source directories, one can use alias to simplify the expression of paths, e.g. \colorbox{lightgray}{"\$mydir1/abc/001.jpg"}, where \$ implies that mydir1 is an alias, which should be specified by either by a global variable in the description file or by an environment variable.

    \item \textbf{object id}: a string with prefix \colorbox{lightgray}{::}, e.g. \colorbox{lightgray}{"::cuhk.ie::abcd1234xyz"}, where cuhk.ie is the name of a data domain, while \colorbox{lightgray}{abcd1234xyz} is an ID string which uniquely identifies an data object in the data domain. When object ids are used, the data platform needs to provide a Key-value mapping facility to map an ID string to the corresponding actual address.

\end{itemize}

\subsubsection{Using type parameters}

From the standpoint of the DSDL interpretator, the type of an element determines how that element is intepreted and validated. In addition to the type name itself, DSDL allows one to provide \textbf{type parameters} to customize how the corresponding elements should be expressed, interpreted, and validated.

\textbf{Label type with parameters}

In the example in get\_started, the field \colorbox{lightgray}{label} of \colorbox{lightgray}{ImageClassificationSample} has the type specified as \colorbox{lightgray}{Label[dom=MyClassDom]}.

Here, \colorbox{lightgray}{Label} is a \textbf{parametric type}, which accepts a \textbf{type parameter} \colorbox{lightgray}{dom}. This {dom} parameter specifies the class domain where the label comes from.

When the domain is explicitly given (here it is given as \colorbox{lightgray}{MyClassDom}), there is no need to provide the class domain names in the values, and thus the labels can be expressed as either the class name or the index. For example, a value \colorbox{lightgray}{"cat"} indicates the fully qualified label \colorbox{lightgray}{"MyClassDom::cat"}; an integer value \colorbox{lightgray}{2} indicates the class label \colorbox{lightgray}{"MyClassDom[2]"}.

\textbf{Date and Time types with parameters}

For \colorbox{lightgray}{Date} and \colorbox{lightgray}{Time} types, when no parameters are explicitly provided, the values should conform to the ISO 8601 format. The interpreter will invoke \colorbox{lightgray}{date.fromisoformat} and \colorbox{lightgray}{time.fromisoformat} methods to parse the string.

One can also specify a customized format using the type parameter \colorbox{lightgray}{fmt}. For example, one can use a type \colorbox{lightgray}{Time[fmt="\%H:\%M"]}, which requires the value should follow the \colorbox{lightgray}{\%H:\%M} format, e.g. \colorbox{lightgray}{"15:32"}. When fmt is explicitly specified, the value of fmt will be fed to \colorbox{lightgray}{strptime} function to parse the time string. Note that this parameter also works for \colorbox{lightgray}{Date} type.

\subsubsection{List type}

DSDL provides a parametric type \colorbox{lightgray}{List} to express unordered or ordered lists. Specifically, an instance of \colorbox{lightgray}{List} is a list that contains multiple elements of a certain element type.

The parametric type \colorbox{lightgray}{List} has two parameters:

\begin{itemize}
    \item \colorbox{lightgray}{etype}: the type of each individual element. This parameter must be explicitly specified.

    \item \colorbox{lightgray}{ordered}: whether there is an sequential order among elements. This parameter is optional, and its default value is \colorbox{lightgray}{false}. This need should only be set to \colorbox{lightgray}{true} for truly sequential types, e.g. sequence of video frames or time series.
\end{itemize}

For example, for a list of integers, we can specify the type as \colorbox{lightgray}{List[Int]}; for a list of class labels within the domain \colorbox{lightgray}{MyClassDom}, we can specify the type as \colorbox{lightgray}{List[Label[MyClassDom]]}.

\subsection{Unstructured Object Classes}

\textbf{Unstructured objects}, such as images, videos, audios, point clouds, and texts, are digital representations of real-world objects. Despite their rich internal structures, they are treated as a whole in a data set description and their internal structures are not manifested.

In DSDL, an \textbf{unstructured object class} provides an abstraction for all unstructured objects of a particular kind.

\subsubsection{Pre-defined unstructured object classes}

With the standard library, DSDL provides the following unstructured object classes out of the box:

\begin{itemize}
    \item \colorbox{lightgray}{Image}: each instance is an image that can be represented as a matrix of pixels.

    \item \colorbox{lightgray}{Video}: each instance can be decoded into a sequence of frames, where each frame is an image.

    \item \colorbox{lightgray}{Audio}: each instance an audio signal that can be represented as a wave sequence.

    \item \colorbox{lightgray}{Text}: each instance is a sequence of words.

    \item \colorbox{lightgray}{PointCloud}: each instance is a set of 3D points, which represents the shape of a 3D entity.

    \item \colorbox{lightgray}{LabelMap}: each instance is a matrix of integer labels, where each label corresponds to a class.
\end{itemize}

\subsubsection{Describe an unstructured object}

In DSDL, an unstructured object can be specified with an object locator together with an optional descriptor that provides additional information about the object.

Take an image stored at \colorbox{lightgray}{abc/0001.jpg} for example. It can be expressed in either of the following ways:

\begin{itemize}
    \item \textbf{Just the object locator}: simply use the object locator \colorbox{lightgray}{"abc/0001.jpg"}. When a variable or a field has an unstructured object class type and its value is a string, then that string will be interpreted as an object locator.

    \item \textbf{With a descriptor}: if one wants to provide additional information, say the size and the color format, then it can be expressed with an JSON object with two properties \colorbox{lightgray}{\$loc} and \colorbox{lightgray}{\$descr}, like the following:
\begin{minted}{yaml}
$loc: "abc/0001.jpg"
$descr:
    size: [640, 480]
    color: "rgb"
\end{minted}
\end{itemize}

Here, the descriptive information is provided via the \colorbox{lightgray}{\$descr} field, which will be used by the object loader of the corresponding unstructured object classes.

\subsubsection{Extended unstructured object classes}

DSDL allows one to register \textbf{extended unstructured object classes} by specifying how to load the object from storage.

At the client side, this can be accomplished by defining a sub-class of an abstract base class \colorbox{lightgray}{UnstructuredObject} and implementing the load method for object loading.

Specifically, in Python, the abstract base class \colorbox{lightgray}{UnstructuredObject} are defined as follows.

\begin{minted}{python}
from abc import ABC, abstractmethod

class UnstructuredObject(ABC):
    """Abstract base class for unstructured objects."""

    @abstractmethod
    def load(file, descr):
        """Load an unstructured object from a given file-like object.

        Arguments:
            file:   file-like, from which the unstructured object is loaded.
            descr:  dict-like, which provides the descriptive information.

        Returns:
            The loaded object.
        """
        pass
\end{minted}

\begin{tcolorbox}[title=Note]
Note here that this load method accepts an file-like object, which is already open, instead of a file path. The design is based on the "Separation of Concerns" principle: it is the responsibility of the data system to interpret the object locator and construct a file reader accordingly. A specific subclass of UnstructuredObject only needs to care about how to load, interpret, and validate the object given a file reader.
\end{tcolorbox}

\subsection{Struct Classes}

Structs are the most common way to represent composite entities. For example, a typical sample in a data set is comprised of multiple elements, e.g. an image with a class label. Hence, structs are a good fit for representing data samples or its composite components.

DSDL allows one to define \textbf{struct classes} to provide an abstraction for a particular type of structs.

\subsubsection{Define a struct class}

In DSDL, one can define a customized struct class in the \colorbox{lightgray}{defs} section of a data set description file. In the example in get\_started, we defined a \colorbox{lightgray}{struct} class named \colorbox{lightgray}{ImageClassificationSample} as follows:

\begin{minted}{yaml}
ImageClassificationSample:
    $def: struct
    $fields:
        image: Image
        label: Label[dom=MyClassDom]
\end{minted}

This class definition is a JSON object, with the following properties:
\begin{itemize}
    \item \colorbox{lightgray}{\texttt{\$defs}}: its value must be "struct", indicating that it is defining a struct class.
    
    \item \colorbox{lightgray}{\$fields}: its value must be a JSON object containing a set of properties, each corresponding to a field. In particular, for each property of \texttt{\$field}, the key will be considered as the field name, while the value is the specification of the corresponding field. The field specification can be given in two ways:

    \begin{itemize}
        \item \textbf{Just the type name}: just give the type name of the field (if that type involves parameters, then the parameters will be set in the default way).
        \item \textbf{With parameters}: one can also specify certain type parameters using a JSON object, which contains a \$type property to specify the type name, and other properties to specify the settings of type parameters. See thet label field specification above.
    \end{itemize}
\end{itemize}

\subsubsection{Nested structs}

In DSDL, structs can be nested. For example, an object detection sample may be comprised of an image together with a set of "local objects", where each local object can be represented by a struct with a bounding box and a class label. For such a sample, we can define a struct class as follows:

\begin{minted}{yaml}
LocalObjectEntry:
    # Each entry refers to an individual detected object on an image.
    $def: struct
    $fields:
        bbox: BBox
        label: Label[dom=MyClassDom]

ObjectDetectionSample:
    # Each sample contains the detection results on an image.
    $def: struct
    $fields:
        image: Image
        objects: List[etype=LocalObjectEntry]
\end{minted}
Here, the structs of class \colorbox{lightgray}{LocalObjectEntry} are embedded into the struct of class \colorbox{lightgray}{ObjectDetectionSample}.

\subsubsection{Parametric struct classes}

Note that in the example above, the struct class \colorbox{lightgray}{LocalObjectEntry} uses a specific class domain \colorbox{lightgray}{MyClassDom}, while the struct class \colorbox{lightgray}{ObjectDetectionSample}, as it nests \colorbox{lightgray}{LocalObjectEntry}, also assumes the use of this particular class domain. Hence, such definitions are not generic. To use another class domain, one has to rewrite both classes.

DSDL provides \textbf{parametric struct classes} to address this problem. Specifically, a parametric struct class can be considered as a class template, which allows the setting of certain parameters when the class is used. With parametric struct classes, we can define classes in a more generic way.

Take the object detection example for instance. We can re-define the classes above as follows:

\begin{minted}{yaml}
LocalObjectEntry:
    $def: struct
    $params: ['cdom']
    $fields:
        bbox: BBox
        label: Label[dom=$cdom]

ObjectDetectionSample:
    $def: struct
    $params: ['cdom']
    $fields:
        image: Image
        objects: List[etype=LocalObjectEntry[cdom=$cdom]]

\end{minted}

Here, we introduce a property \colorbox{lightgray}{\$params} in struct class definition. When the \colorbox{lightgray}{\$params} property is explicitly given and non-empty, then the corresponding struct class is \textbf{parametric}. Note that when a parametric class is used, its parameters must be given in order to make it into a \textbf{concrete class}.

Particularly, in the \colorbox{lightgray}{LocalObjectEntry} class above, we introduce a class parameter \colorbox{lightgray}{cdom}, which are used in specifying the \colorbox{lightgray}{domain} attribute of \colorbox{lightgray}{label}. Note that when a class parameter is used, it should be enclosed by \colorbox{lightgray}{[]}. Then, the class \colorbox{lightgray}{ObjectDetectionSample} is also defined as a parametric struct class with a parameter \colorbox{lightgray}{cdom}, and the parameter is used when specifying the type of \colorbox{lightgray}{objects}.

With such class definitions, we can write the set of data samples as follows:

\begin{minted}{yaml}
data:
    sample-type: 
        $type: ObjectDetectionSample
        cdom: MyClassDom
    samples:
        - image: "abc/0001.jpg"
            objects:
            - { bbox: [1, 2, 3, 4], label: 1 }
            - { bbox: [5, 6, 7, 8], label: 2 }
        - image: "abc/0002.jpg"
            objects:
            - { bbox: [1, 2, 3, 4], label: 3 }
            - { bbox: [5, 6, 7, 8], label: 2 }
            - { bbox: [4, 3, 5, 8], label: 3 }

\end{minted}

Note that with the parameter \colorbox{lightgray}{cdom} is given as \colorbox{lightgray}{MyClassDom}, the parametric class \colorbox{lightgray}{ObjectDetectionSample} is made into a concrete class and used in sample type specification.

\subsection{Class Domain}

A class domain (\textit{class\_domain}) provides a detailed definition for the \colorbox{lightgray}{Label} field, which is a type of data annotation. Instantiation of the Label field requires assigning a specific class domain.

\subsubsection{Defining a Class Domain.}
In DSDL, users can customize a class domain within the \colorbox{lightgray}{\$def} section of the dataset description file. Consider the following example:

\noindent\textbf{Class Domain Example.}

\begin{minted}{yaml}
MyClassDom:
    $def: class_domain
    classes:
        - airplane
        - apple
        - backpack
        - banana
        - baseball bat
        - baseball glove
        - bear
\end{minted}

A class domain is defined using a YAML object, which includes the following attributes:
\begin{itemize}
    \item \colorbox{lightgray}{\$def}: Its value must be \colorbox{lightgray}{\textit{class\_domain}}, indicating that this YAML object is defining a class domain.
    \item \colorbox{lightgray}{classes}: Its value must be a YAML list, containing a series of class names, each of which is a string representing a category.
\end{itemize}

\subsubsection{Referencing a Class Domain}

Class domains are primarily used to instantiate the \colorbox{lightgray}{Label} field in the struct class, for example:

\begin{minted}{yaml}
ImageClassificationSample:
    $def: struct
    $fields:
        image: Image
        label: Label[dom=MyClassDom]
\end{minted}

Alternatively, parameters can initially be used as placeholders, as seen in structured class types:

\begin{minted}{yaml}
ImageClassificationSample:
    $def: struct
    $params: ['cdom']
    $fields:
        image: Image
        label: Label[dom=$cdom]
\end{minted}

Later, during the data section (describing the dataset), this parameter can be instantiated in the sample-type, as follows:

\begin{minted}{yaml}
data:
    sample-type: ImageClassificationSample[cdom=MyClassDom]
    sample-path: $local
    samples:
        - image: "xyz/0001.jpg"
          label: "apple"
        - image: "xyz/0002.jpg"
          label: "banana"
\end{minted}

\subsubsection{Defining Hierarchical Class Domains}

In some datasets, categories have hierarchical relationships. DSDL allows each category to specify its parent by directly defining a subclass class domain and connecting each label to the specific category of its parent with a dot \colorbox{lightgray}{.}:

\begin{minted}{yaml}
ClassDom:                        # Declaration of categories
    $def: class_domain
    classes:
        - vehicle.airplane       # Indicating airplane's parent category from ParentClassDom is vehicle
        - food.apple
        - accessory.backpack
        - food.banana
        - sports.baseball_bat    # Category names with spaces or special symbols are converted to underscores
        - sports.baseball_glove
        - animal.bear
        - furniture.bed
        - outdoor.bench
        - vehicle.bicycle
        - animal.bird
        - vehicle.boat
\end{minted}

In category definitions, the dot \colorbox{lightgray}{.} is used to separate hierarchical relationships between categories. Labels with multiple words should be connected with underscores \colorbox{lightgray}{\_}. If category labels include spaces or other special characters, it is recommended that the global-info section stores a mapping between the original label names with special characters and the labels stored in the class\_dom (for more on the definition and storage location of global-info, please refer to the data module section). The format is as follows:

\begin{minted}{yaml}
{"global-info":
    {"name_mapping":[
            {
                "name": "baseball_bat.sports"
                "original_name": "baseball bat.sports",
            },
            ...
    ]}
}
\end{minted}

\subsection{Data Section}

A dataset description file is generally divided into four main sections:

\begin{itemize}
    \item \textbf{Header}: Specifies how the dataset description file should be parsed.
    \item \textbf{Meta Section}: Provides meta-information about the dataset.
    \item \textbf{Defs Section}: Contains global definitions such as user-defined class domains and structures.
    \item \textbf{Data Section}: Describes the sample data within the dataset.
\end{itemize}

Previous chapters primarily discussed the \textbf{defs section}, clearly defining class domains and structures. Next, we need to define each specific sample, which is the role of the Data Section.

\subsubsection{Defining the Data Section}
From the Quick Start, we have the following example:

\begin{minted}{yaml}
data:
    sample-type: ImageClassificationSample
    sample-path: $local
    samples:
        - { image: "xyz/0001.jpg", label: "cat" }
        - { image: "xyz/0002.jpg", label: "dog" }
        ...
\end{minted}

A data section includes three modules:
\begin{itemize}
    \item \colorbox{lightgray}{Sample-type}: Defines the data type, typically a struct we have defined. If the struct contains parameters, they must be instantiated. Additionally, \colorbox{lightgray}{sample-type} can be any other data type (Label, Image, Int, etc.), as structs are also considered data types.
    \item \colorbox{lightgray}{Sample-path}: The storage path for samples. If it is an actual path, the contents of samples are read from that file. If it is\colorbox{lightgray}{\$local} (as in this example), it reads directly from the \colorbox{lightgray}{data.samples} field within this file.
    \item \colorbox{lightgray}{Samples}: Stores the sample data of the dataset. This field is only active when \colorbox{lightgray}{sample-path} is set to \colorbox{lightgray}{\$local}; otherwise, samples are read from the specified path.
\end{itemize}

Fields in \colorbox{lightgray}{data.samples} correspond directly to the data type defined in \colorbox{lightgray}{data.sample-type}:

\begin{minted}{yaml}
MyClassDom:
    $def: class_domain
    classes:
        - dog
        - cat
        - fish
        - tiger
ImageClassificationSample:
    $def: struct
    $fields:
        image: Image
        label: Label[dom=MyClassDom]
\end{minted}

\subsubsection{Usage of Optional}
If some fields are missing in our data, such as labels in some entries of \colorbox{lightgray}{data.samples}:

\begin{minted}{yaml}
data:
    sample-type: ImageClassificationSample
    sample-path: $local
    samples:
        - { image: "xyz/0001.jpg" }
        - { image: "xyz/0002.jpg", label: "dog" }
        - { image: "xyz/0003.jpg" }
        - { image: "xyz/0004.jpg", label: "tiger" }
\end{minted}

In defining the structure, we need to indicate which fields may be missing by placing the potentially missing field names in the $optional part of the struct type. The $optional field is a list, formatted as follows:

\begin{minted}{yaml}
ImageClassificationSample:
    $def: struct
    $fields:
        image: Image
        label: Label[dom=MyClassDom]
    $optional: ['label']
\end{minted}

\begin{tcolorbox}[title=Note]
When a label is omitted from \$optional, the corresponding field value can be missing in data.samples; otherwise, a warning will be displayed.
\end{tcolorbox}

\subsubsection{Adding a Global-info Field}
For global information about the dataset, we can add a global-info field in the data module. Global information includes but is not limited to:

\begin{itemize}
    \item Supplemental category information (such as synonyms, definitions, etc.)
    \item Supplemental data information (such as a global vocabulary for OCR tasks, keypoints connection methods for keypoint tasks, etc.)
\end{itemize}

To store global-info, we need to define its structure and add the corresponding field in the data module. Here is an example of how to use global-info:

Firstly, the following content needs to be added to the definition file:
\begin{minted}{yaml}
ExampleClassDomDescr:
    $def: struct
    $fields: 
        label: Label[dom=MyClassDom]    # Category name
        def: Str                        # Category description
        synonyms: List[Str]             # Synonyms

GlobalInfo:
    $def: struct
    $fields: 
        class-info: List[ExampleClassDomDescr]
\end{minted}

The data module is shown as follows:
\begin{minted}{yaml}
    data:
    global-info-type: GlobalInfo
    global-info-path: $local
    global-info:
      class-info:
          - label: "dog"
            def: 'a very common four-legged animal that is often kept by people as a pet or to guard or hunt.'
            synonyms: ['puppy', 'hound']
          - label: "fish"
            ...
\end{minted}

A new global-info related information is added to a data section, which mainly includes three modules:

\begin{itemize}
\item \colorbox{lightgray}{\texttt{global-info-type}}: The type definition of global information, which is generally a struct we defined. If this struct contains parameters, these parameters need to be instantiated. For more details, see section 3.4.3 Struct Classes with Parameters.
\item\colorbox{lightgray}{\texttt{global-info-path}}: The storage location of global information. If it is actually a path, the content of global-info is read from this file. If it is \colorbox{lightgray}{\$local} (as in this example), it is directly read from the \colorbox{lightgray}{\texttt{data.global-info}} field of this file.
\item \colorbox{lightgray}{\texttt{global-info}}: The specific content of the saved global information. Note that this field only takes effect when \texttt{global-info-path} is \$local, otherwise it will prefer to read from the path in \texttt{global-info-path}.
\end{itemize}

It's worth noting that the fields in \colorbox{lightgray}{\texttt{data.global-info}} correspond one-to-one with the data type given in \colorbox{lightgray}{\texttt{data.global-info-type}}.

\subsubsection{Reading from External Files}
For large datasets, it is advisable to extract data from the YAML file's data section and store it separately in a JSON file, specifying the path to this JSON file in the sample-path field of the data section. For example:

\begin{minted}{yaml}
data:
    global-info-type: GlobalInfo
    global-info-path: global_info.json  
    sample-type: ExampleSample  
    sample-path: samples.json
\end{minted}

In samples.json, we store our specific data, for example:

\begin{minted}{json}
{
    "samples":[
        {"image": "xyz/0001.jpg", "label": "cat"}, 
        {"image": "xyz/0002.jpg", "label": "dog"}, 
        {"image": "xyz/0003.jpg", "label": "dog"}, 
        {"image": "xyz/0004.jpg", "label": "tiger"},
        ...
    ]
}
\end{minted}

In global\_info.json, we store our specific global information data, for example:

\begin{minted}[breaklines=true]{json}
{
    "global-info":{
        "class-info":[
                {
                    "label": "dog", 
                    "synonyms": ["puppy", "hound"], 
                    "def": "a very common four-legged animal that is often kept by people as a pet or to guard or hunt."
                },
                ...
        ]
    }
}
\end{minted}

\subsection{Libraries}

Whereas we are already trying to simplify the design of DSDL, some efforts remain needed to learn how to define classes in DSDL. However, we understand that most AI researchers or developers don't want to learn yet another language. Hence, we introduce libraries to further simplify the process of data set description.

\subsubsection{Define and import a library}

Consider the example in get\_started, the part that defines the class ImageClassificationSample is quite generic and can be used in many data sets. Hence, we can extract it to a library file, while the data set description file can just import it.

In general, a \textbf{library file} is a file in YAML or JSON format that provides a collection of definitions.

In the example above, we can provide a \textbf{library file} named imageclass.yaml as follows:

\begin{minted}{yaml}
[file: imageclass.yaml]
# DSDL version is mandatory here.
$dsdl-version: "0.5.0"

# -- below are definitions --

MyClassDom:
    $def: class_domain
    classes:
        - dog
        - cat
        - fish
        - tiger

ImageClassificationSample:
    $def: struct
    $fields:
        image: Image
        label: Label[dom=MyClassDom]
\end{minted}

\begin{tcolorbox}[title=Note]
- This library file should be placed in the default library path so the system can find it.

- One can also supply additional library paths by setting an environment variable DSDL\_LIBRARY\_PATH.

\end{tcolorbox}

Then the data set description can be simplified by importing the library, as follows:

\begin{minted}{yaml}
$dsdl-version: "0.5.0"
$import: 
    - imageclass
meta:
    name: "my-dataset"
    creator: "my-team"
    dataset-version: "1.0.0"
data:
    sample-type: ImageClassificationSample
    samples:
        - { image: "xyz/0001.jpg", label: "cat" }
        - { image: "xyz/0002.jpg", label: "dog" }

\end{minted}

Here, we use an \colorbox{lightgray}{\$import} directive in the header section. The content of \colorbox{lightgray}{\$import} should be a list, which means that one can import multiple files.

\begin{tcolorbox}[title=Note]
When multiple library files being imported contain definitions of the same name, then the definition imported later will overwrite previous ones. In this case, the interpreter should raise a warning.
\end{tcolorbox}

\subsubsection{Better Practice of Using Libraries}

Here are some good practice for defining a DSDL library:

\noindent\textbf{Define generic classes.}
As discussed in :ref:parametric\_class, it is not a good idea to involving specific settings of parameters in a generic class definition. Hence, it is strongly suggested that one defines a parametric class if the class requires specific information related to a particular application (e.g. class domains) in order to be completed.

\noindent\textbf{Grouped definitions.}
It is advisable to put multiple definitions related to a certain area into one library file. This is makes it easier to distribute and import.

\noindent\textbf{Documentation.}
Document the definitions to make it easier for users to understand.

Below is an example where we put multiple classes related to visual recognition into a single library file:

\begin{minted}{yaml}
# file: visualrecog.yaml

ImageClassificationSample:
    # Each image classification sample contains an image and a label.
    $def: struct
    $params: ['cdom']
    $fields:
        image: Image
        label: Label[dom=$cdom]

LocalObjectEntry:
    # Each local object entry contains a bounding box and a label.
    $def: struct
    $params: ['cdom']
    $fields:
        bbox: BBox
        label: Label[dom=$cdom]

ObjectDetectionSample:
    # Each object detection sample contains an image and a list of local object entries.
    $def: struct
    $params: ['cdom']
    $fields:
        image: Image
        objects: List[etype=LocalObjectEntry[cdom=$cdom]]
\end{minted}

With this library, one can write a data set description as follows:

\begin{minted}{yaml}
$dsdl-version: "0.5.0"
$import: 
    - visualrecog
meta:
    name: "my-dataset"
    creator: "my-team"
    dataset-version: "1.0.0"
defs:
    MyClassDom:
        $def: class_domain
        classes:
            - dog
            - cat
            - fish
            - tiger
data:
    sample-type: ImageClassificationSample
    samples:
        - { image: "xyz/0001.jpg", label: "cat" }
        - { image: "xyz/0002.jpg", label: "dog" }
\end{minted}

Since MyClassDom is a specific definition only used in this dataset. Hence, it is fine to define it in the description file itself as a customized definition, while importing common classes from a library.

\begin{tcolorbox}[title=Note]
The library visualrecog is an example just for illustration. Along with DSDL, we provide a standard library named cv that contains a rich collection of definitions related to computer vision, including various types and commonly used class domains, etc.
\end{tcolorbox}

\section{DSDL Examples}

\subsection{Image Classification}
The task of image classification is to assign a class label to each image.

\noindent\textbf{Sample class definition:}
\begin{minted}{yaml}
ImageClassificationSample:
    # Each sample contains an image together with a class label (optional).
    $def: struct
    $params: ['cdom'] 
    $fields:
        image: Image
        label: Label[dom=$cdom]
    $optional: ['label']
\end{minted}

Here, the fields in the \$optional list can be omitted in the data samples. When label is omitted, the corresponding field value will be set to a null value (in Python it is None).

\noindent\textbf{Data samples:}

\begin{minted}{yaml}
data:
    sample-type: ImageClassificationSample[cdom=MyClassDom]
    samples:
        - { image: "xyz/0001.jpg", label: "cat" }
        - { image: "xyz/0002.jpg", label: "dog" }
        - { image: "xyz/0003.jpg" }
\end{minted}

\subsection{Object Detection}
The task of object detection is to detect meaningful objects on an image. Each detected object can be represented by a bounding box together with an object class label.

\noindent\textbf{Sample class definition}:

\begin{minted}{yaml}
LocalObjectEntry:
    # Each sample contains a bounding box together a class label (optional).
    $def: struct
    $params: ['cdom']
    $fields:
        bbox: BBox
        label: Label[dom=$cdom]
    $optional: ['label']

ObjectDetectionSample:
    # Each sample contains an image together with a list of local objects.
    $def: struct
    $params: ['cdom']
    $fields:
        image: Image
        objects: List[LocalObjectEntry[cdom=$cdom]]
\end{minted}

\noindent\textbf{Data samples}:
\begin{minted}{yaml}
data:
    sample-type: ObjectDetectionSample[cdom=MyClassDom]
    samples:
        - image: "xyz/0001.jpg"
            objects: 
            - { bbox: [x1, y1, w1, h1], label: 1 }
            - { bbox: [x2, y2, w2, h2], label: 2 }
            - { bbox: [x3, y3, w3, h3], label: 2 }
        - image: "xyz/0002.jpg"
            objects: 
            - { bbox: [x4, y4, w4, h4], label: 3 }
            - { bbox: [x5, y5, w5, h5], label: 4 }

\end{minted}

\subsection{Object Detection with Scene Classification}
In certain applications, scene classification (at the image level) and object detection are combined into a joint task.

For such a task, we can have two class domains, say SceneDom and ObjectDom, respectively for scene classes and object classes.

\noindent\textbf{Sample class definition}:
\begin{minted}{yaml}
SceneAndObjectSample:
    # Each samples contains an image, a scene label (optional)
    # together with a list of local objects.
    $def: struct
    $params: ['scenedom', 'objectdom']
    $fields:
        image: Image 
        sclabel: Label[dom=$scenedom]
        objects: List[LocalObjectEntry[cdom=$objectdom]]
    $optional: ['sclabel']
\end{minted}

\noindent\textbf{Data samples}:
\begin{minted}{yaml}
data:
    sample-type: SceneAndObjectSample[scenedom=SceneDom, objectdom=ObjectDom]
    samples:
        - image: "xyz/0001.jpg"
            sclabel: "street"
            objects: 
            - { bbox: [x1, y1, w1, h1], label: 1 }
            - { bbox: [x2, y2, w2, h2], label: 2 }
            - { bbox: [x3, y3, w3, h3], label: 2 }
        - image: "xyz/0002.jpg"
            sclabel: "river"
            objects: 
            - { bbox: [x4, y4, w4, h4], label: 3 }
            - { bbox: [x5, y5, w5, h5], label: 4 }

\end{minted}

\subsection{Image Segmentation}

The image segmentation task is to assign pixel-wise labels to an image. A common practice is to use a labelmap, which is stored as an unstructured object in an external file.

\noindent\textbf{Sample class definition}:
\begin{minted}{yaml}
ImageSegmentationSample:
    # Each sample contains an image together with a label map.
    $def: struct
    $params: ['cdom']
    $fields:
        image: Image
        labelmap: LabelMap[dom=$cdom]
\end{minted}

\noindent\textbf{Data samples}:
\begin{minted}{yaml}
data:
    sample-type: ImageSegmentationSample[cdom=MyClassDom]
    samples:
        - { image: "imgs/0001.jpg", labelmap: "maps/0001.ppm" }
        - { image: "imgs/0002.jpg", labelmap: "maps/0002.ppm" }
        - { image: "imgs/0003.jpg", labelmap: "maps/0003.ppm" }

\end{minted}

\section{DSDL Templates}

\subsection{Image Classification}

\subsubsection{Task Research}

\begin{figure}[h]
    \centering
    \includegraphics[width=0.8\textwidth]{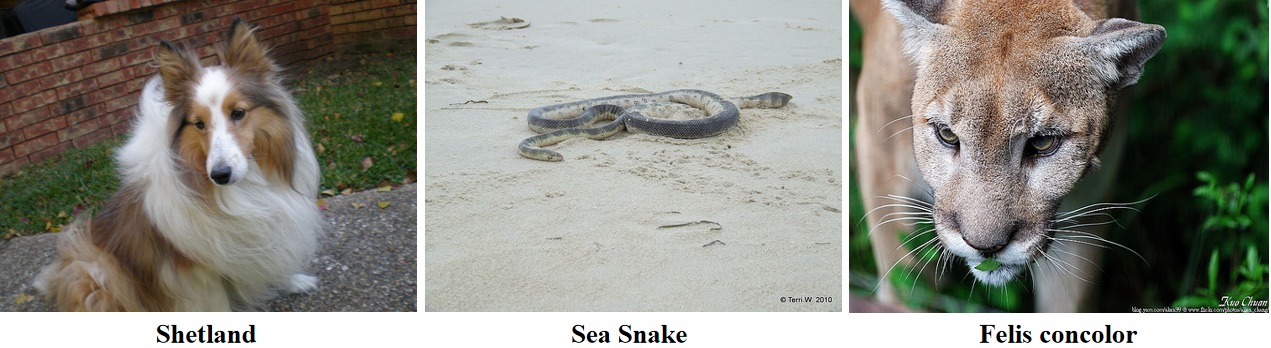}
    \caption{Example of an input image and its classification}
    \label{fig: cls1}
\end{figure}

\noindent \textbf{Task Definition.} 
Image classification refers to the process of outputting the semantic category of a given input image (~\Cref{fig: cls1}).

\begin{figure}[h]
    \centering
    \includegraphics[width=0.45\textwidth]{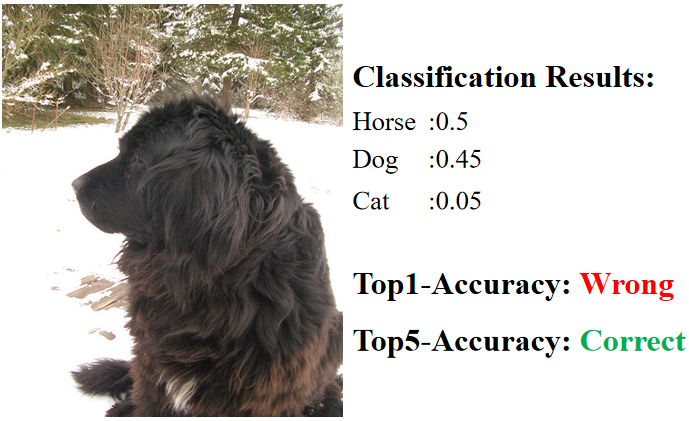}
    \caption{Illustration of top-1 and top-5 accuracy}
    \label{fig: cls2}
\end{figure}

\noindent \textbf{Evaluation Metrics.}
The evaluation metrics for the task generally include two: top-5 accuracy and top-1 accuracy, where accuracy = (number of correctly classified images) / (total number of images). Top-5 accuracy means that if one of the top five highest scoring predicted categories is correct, then the sample is considered correctly classified. Top-1 accuracy requires that the highest scoring predicted category must be correct for the sample to be considered correctly classified (~\Cref{fig: cls2}).

\noindent \textbf{Mainstream Dataset Survey.}
We have surveyed 10 mainstream classification datasets, including well-known datasets such as ImageNet-1K~\cite{russakovsky2015imagenet}, MNIST~\cite{lecun1998gradient}, and CIFAR10~\cite{krizhevsky2009learning}. To make the template more versatile and expandable, we focused on the commonalities and unique characteristics among these datasets. During the survey, we encountered fields with different names but similar or identical meanings. We treated these as the same field and standardized their naming. For example, the \textit{image\_id} field generally represents the path or ID of an image, uniquely identifying it; \textit{label\_id} indicates the category of the image, which can be represented either by numbers or strings. The complete survey results of the fields are shown in~\Cref{table:clss1}:

\begin{table}[h]
\centering
\resizebox{0.7\linewidth}{!}{ 
\begin{tabular}{@{}lcccc@{}}
\toprule
\textbf{Image Classification Dataset} & \multicolumn{2}{c}{\textbf{Shared Fields}} & \textbf{Independent Fields} \\
\cmidrule(lr){2-3}
 & \textbf{image\_id} & \textbf{label\_id} & \textbf{superclass} \\
\midrule
ImageNet-21K~\cite{russakovsky2015imagenet} & Y & Y & Y \\
ImageNet-1K~\cite{russakovsky2015imagenet} & Y & Y & Y \\
CIFAR10~\cite{krizhevsky2009learning} & Y & Y &  \\
CIFAR100~\cite{krizhevsky2009learning} & Y & Y & Y \\
MNIST~\cite{lecun1998gradient} & Y & Y &  \\
Fashion-MNIST~\cite{xiao2017fashion} & Y & Y &  \\
Caltech-256~\cite{griffin2007caltech} & Y & Y &  \\
Oxford Flowers-102~\cite{nilsback2008automated} & Y & Y &  \\
\bottomrule
\end{tabular}
}
\caption{Survey of fields across different image classification datasets}
\label{table:clss1}
\end{table}

After organizing, the shared and unique fields for classification tasks are as shown in~\Cref{table:clss2}:

\begin{table}[h]
\centering
\begin{tabular}{@{}lcc@{}}
\toprule
\textbf{Field Type} & \textbf{Field Name} & \textbf{Meaning} \\ 
\midrule
\multirow{2}{*}{Shared Fields} & image\_id & Identifies a unique image, e.g., by image name or path \\ 
 & label\_id & The category of the image, with type int for single label, List[int] for multiple labels \\ 
\midrule
Unique Fields & superclass & The superclass of the image category, e.g., "dog" might have "animal" as a superclass \\
\bottomrule
\end{tabular}
\caption{Description of shared and unique fields in classification datasets}
\label{table:clss2}
\end{table}

As can be seen, to describe a sample in a classification dataset, the most basic fields are \textit{image\_id} and \textit{label\_id}, along with unique fields like \textit{superclass}.

\subsubsection{Template Presentation}

Based on the above research results, we know that for classification tasks, the most critical attributes of a sample are the image's ID (or path) and the image's category. Therefore, in the structure of the classification task, we define two fields in the fields attribute: image and label. Furthermore, since different datasets contain different categories, it is necessary to have a formal parameter in the sample to define the category domain (in DSDL, we describe the category domain as class domain, or cdom). Lastly, considering some unsupervised or semi-supervised classification tasks, some samples may not contain category information, thus we designed the \textbf{\$optional} field and included the label field within it. Specific fields unique to some datasets can also be included in the \textbf{\$optional} field. Based on the above considerations, we have formulated a template for classification tasks, as shown below:

\begin{minted}{yaml}
$dsdl-version: "0.5.0"

ClassificationSample:
    $def: struct
    $params: ['cdom']
    $fields:
        image: Image
        label: Label[dom=$cdom]
    $optional: ['label']
\end{minted}

The meanings of some fields in the template are as follows:

\begin{itemize}
  \item \textbf{\$dsdl-version}: Describes the dsdl version corresponding to this file.
  \item \textbf{ClassificationSample}: Defines the sample format for the classification task, which includes four attributes:
    \begin{itemize}
      \item \textbf{\$def}: Indicates that ClassificationSample is a struct.
      \item \textbf{\$params}: Defines the formal parameters, here the class domain.
      \item \textbf{\$fields}: Attributes contained in the struct, specifically for classification tasks, include:
        \begin{itemize}
          \item image: Image path
          \item label: Category information
        \end{itemize}
      \item \textbf{\$optional}: Covers optional attributes of the struct, here only one field is defined, i.e., label, indicating that the presence of label in a single sample is optional. Additionally, specific fields unique to the dataset can also be included in the \textbf{\$optional} field.
    \end{itemize}
\end{itemize}

For the class domain `cdom' mentioned in the template, here is an example of the class domain for the cifar10 dataset:
\begin{minted}{yaml}
$dsdl-version: "0.5.0"

Cifar10ImageClassificationClassDom:
    $def: class_domain
    classes:
        - airplane
        - automobile
        - bird
        - cat
        - deer
        - dog
        - frog
        - horse
        - ship
        - truck
\end{minted}

The above file defines Cifar10ImageClassificationClassDom, which includes the following fields:
\begin{itemize}
  \item \$def: Describes the type of Cifar10ImageClassificationClassDom, here class\_domain.
  \item classes: Describes the categories contained within the class domain and their order in the cifar10 dataset, which are sequentially airplane, automobile, etc.
\end{itemize}

This section introduces the classification task template and cdom template, both of which can be directly called from our template library. 

\subsubsection{Usage Instructions}

This section explains how to use our templates by importing them. For example, consider the cifar10 dataset. Below is the \textit{train.yaml} template, which describes all samples in the dataset:

\begin{minted}{yaml}
$dsdl-version: "0.5.0"

$import:
    - ../defs/class-dom
    - ../defs/classification-cifar10

meta:
    name: "cifar10"
    subdata-name: "train"

data:
    sample-type: ClassificationSample[cdom=Cifar10ImageClassificationClassDom]
    sample-path: $local
    samples:
      - image: "images/000000000000.png"
        label: frog
      - image: "images/000000000001.png"
        label: truck
        ...
\end{minted}

The description file specifies:
\begin{itemize}
  \item \$dsdl-version: Version of the DSDL used.
  \item \$import: Imports required templates, including those for classification tasks and the cifar10 class domain.
  \item meta: Contains metadata about the dataset, such as its name and subset.
  \item data: Contains sample information structured according to the imported templates:
    \begin{itemize}
      \item sample-type: Specifies the type of data, using the ClassificationSample class with the cifar10 class domain.
      \item sample-path: Indicates where samples are stored. In this example, samples are read directly from this file under data.samples.
      \item samples: Lists individual sample details, active only if sample-path is set to \$local.
    \end{itemize}
\end{itemize}

\subsection{Object Detection}

\subsubsection{Task Research}

\noindent\textbf{Task Definition}
Object detection involves identifying the location (usually represented by bounding boxes) and semantic categories of objects in a given input image. An example diagram is shown in~\Cref{fig: det1}:

\begin{figure}[H]
    \centering
    \includegraphics[width=0.5\textwidth]{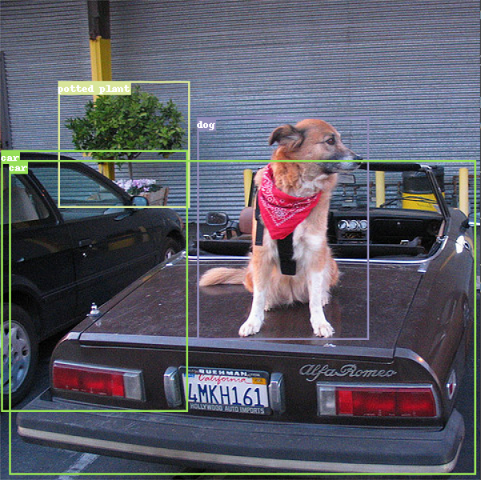}
    \caption{Example of an object detection task}
    \label{fig: det1}
\end{figure}

\noindent\textbf{Evaluation Metrics} \label{sec:detection}
The most common evaluation metric in object detection is Average Precision (AP), along with several derivatives based on AP, explained as follows:
\begin{itemize}
    \item \textbf{AP}: The area under the Precision-Recall curve.
    \item \textbf{11-point interpolation}: Simplifies AP calculation, sometimes using the 11-point method (e.g., in the PASCAL VOC dataset), which calculates the average precision at eleven recall levels (0, 0.1, 0.2, ..., 1). If a precision value is not available at a recall level, the next highest precision is used. The formula is:
    \[
    AP = \frac{1}{11} \sum_{r \in \{0, 0.1, 0.2, \ldots, 1\}} \rho_{\text{interp}}(r)
    \]
    \[
    \rho_{\text{interp}}(r) = \max_{\tilde{r} \geq r} \rho(\tilde{r})
    \]
    \item \textbf{mAP}: The mean of AP across all categories.
    \item \textbf{mAP@0.5 and mAP@0.5--0.95}: AP calculated at different thresholds. mAP@0.5 is the AP at a threshold of 0.5, while mAP@0.5--0.95 averages APs calculated at thresholds from 0.5 to 0.95. The PASCAL VOC dataset uses mAP@0.5, often referred to as AP\_50, while the COCO dataset uses mAP@0.5--0.95.
    \item \textbf{mAP\_s, mAP\_m, mAP\_l}: These metrics represent object sizes (small, medium, large) in the COCO dataset, reflecting the model's detection performance across different box sizes.
\end{itemize}

\noindent\textbf{Mainstream Dataset Research}
We researched ten object detection datasets, analyzing and summarizing their annotation fields. Fields with the same meaning are displayed using unified names. The summarized information is shown in~\Cref{table:det1}:
\begin{table}[H]
    \centering
    \resizebox{\linewidth}{!}{ 
    \begin{tabular}{c|ccc|cccccccccc|}
    \toprule
    \textbf{Dataset} & \multicolumn{3}{c|}{\textbf{Shared Fields}} & \multicolumn{9}{c}{\textbf{Unique Fields}} \\
    \midrule
    & \textbf{ImgID} & \textbf{LblID} & \textbf{BBox} & \textbf{Crowd} & \textbf{Trunc} & \textbf{Diff} & \textbf{Occ} & \textbf{Depict} & \textbf{Reflect} & \textbf{Inside} & \textbf{Conf} & \textbf{Pose} \\
    \midrule
    PASCAL VOC~\cite{everingham2011the} & Y & Y & Y & & Y & Y & Y & & & & & Y \\
    COCO~\cite{lin2014microsoft} & Y & Y & Y & Y & & & & & & & & \\
    KITTI~\cite{Geiger2012CVPR} & Y & Y & Y & & Y & & Y & & & & Y & \\
    OpenImages~\cite{kuznetsova2020open} & Y & Y & Y & Y & Y & & Y & Y & & Y & Y & \\
    Objects365~\cite{9009553} & Y & Y & Y & Y & & & & Y & Y & & & \\
    ILSVRC2015~\cite{ILSVRC15} & Y & Y & Y & & & & Y & & & & Y & \\
    LVIS~\cite{gupta2019lvis} & Y & Y & Y & & & & & & & & & \\
    \bottomrule
    \end{tabular}
    }
    \caption{Summary of shared (ImgID: Image ID, LblID: Label ID, BBox: Bounding Box) and unique fields (Crowd: Is Crowd, Trunc: Is Truncated, Diff: Is Difficult, Occ: Is Occluded, Depict: Is Depiction, Reflect: Is Reflected, Inside: Is Inside, Conf: Confidence, Pose: Pose) in object detection datasets}
\label{table:det1}
\end{table}

The meanings of shared and unique fields are further summarized in~\Cref{table:det2}:
\begin{table}[H]
    \centering
    \begin{tabular}{@{}clp{10cm}@{}} 
    \toprule
    \textbf{Field Type} & \textbf{Field Name} & \textbf{Meaning} \\
    \midrule
    \multirow{3}{*}{Shared Fields} & image\_id & Identifies a unique image, such as by name or path \\
                                   & label\_id & Category of a single object \\
                                   & bbox      & Bounding box for a single object, e.g., [xmin, ymin, xmax, ymax] \\
    \midrule
    \multirow{9}{*}{Unique Fields} & iscrowd     & Whether the object is part of a dense group, e.g., a crowd of people or a pile of apples \\
                                   & istruncated & Whether the object is truncated, i.e., partially outside the image \\
                                   & isdifficult & Whether the object is difficult to detect \\
                                   & isoccluded  & Whether the object is occluded \\
                                   & isdepiction & Whether the object is a depiction, such as a cartoon or painting, rather than a real entity \\
                                   & isreflected & Whether the object is a reflection \\
                                   & isinside    & Whether the object is inside another object, e.g., a car inside a building, a person inside a car \\
                                   & confidence  & Confidence of the detection box, usually 1 for manually annotated and typically between 0.5 and 1 for automatically generated \\
                                   & pose        & Shooting angle, values include Unspecified, Frontal, Rear, Left, Right \\
    \bottomrule
    \end{tabular}
    \caption{Detailed description of fields in object detection datasets}
\label{table:det2}
\end{table}

\subsubsection{Template Display} \label{Sec:det template}
Based on the aforementioned research, an object detection task involves an image corresponding to an unspecified number of targets, each located by a bounding box (BBox) and provided with a semantic label. The object detection template is defined as follows:

\begin{minted}{yaml}
$dsdl-version: "0.5.0"

LocalObjectEntry:
    $def: struct
    $params: ['cdom']
    $fields:
        bbox: BBox
        label: Label[dom=$cdom]

ObjectDetectionSample:
    $def: struct
    $params: ['cdom']
    $fields:
        image: Image
        objects: List[LocalObjectEntry[cdom=$cdom]]
\end{minted}

The fields in the detection template are explained below:
\begin{itemize}
    \item \$dsdl-version: Specifies the DSDL version of the file.
    \item LocalObjectEntry: A nested structure defining the description of a bounding box, containing:
        \begin{itemize}
            \item \$def: Indicates a structure type.
            \item \$params: Defines parameters, here the class domain.
            \item \$fields: Attributes of the structure, including:
                \begin{itemize}
                    \item bbox: Position of the bounding box.
                    \item label: Category of the bounding box.
                \end{itemize}
        \end{itemize}
    \item ObjectDetectionSample: Defines the structure of an object detection sample, containing:
        \begin{itemize}
            \item \$def: Indicates a structure type.
            \item \$params: Defines parameters, here the class domain.
            \item \$fields: Attributes of the structure, including:
                \begin{itemize}
                    \item image: Path of the image.
                    \item objects: Annotation information, a list of LocalObjectEntry.
                \end{itemize}
        \end{itemize}
\end{itemize}

\subsubsection{Complete Example}
Using the PASCAL VOC dataset as an example, we present the specific content of an object detection dataset DSDL description file.

\vspace{10pt}
\noindent\textbf{DSDL Syntax Describing Category Information}

\begin{minted}{yaml}
$dsdl-version: "0.5.0"

VOCClassDom:
    $def: class_domain
    classes:
        - horse
        - person
        - bottle
        - tvmonitor
        - chair
        - diningtable
        - pottedplant
        - aeroplane
        - car
        - train
        - dog
        - bicycle
        - boat
        - cat
        - sofa
        - bird
        - sheep
        - motorbike
        - bus
        - cow
\end{minted}

The file above defines VOCClassDom, which includes the following fields:
\begin{itemize}
    \item \$def: Describes the DSDL type of VOCClassDom, here a class\_domain.
    \item classes: Lists the categories within this class domain, in the PASCAL VOC dataset, these include horse, person, etc.
\end{itemize}

\newpage
\noindent\textbf{Dataset YAML File Definition}
\begin{minted}{yaml}
$dsdl-version: "0.5.0"

$import:
    - object-detection
    - voc-class-domain

meta:
    dataset_name: "PASCAL VOC2007"
    sub_dataset_name: "train"
    
data:
    sample-type: ObjectDetectionSample[cdom=VOCClassDom]
    sample-path: $local
    samples: 
      - image: "media/000000000000.jpg"
        objects:
          - {bbox: [4.0, 36.0, 496.0, 298.0], label: 12}
      - image: "media/000000000002.jpg"
        objects:
          - {bbox: [440.0, 161.0, 60.0, 81.0], label: 14}   
          - {bbox: [97.0, 159.0, 121.0, 67.0], label: 14}
          - {bbox: [443.0, 116.0, 57.0, 101.0], label: 15}
          - {bbox: [104.0, 113.0, 65.0, 106.0], label: 15}
      ...
\end{minted}

In the description file above, the version of DSDL is first defined, followed by the import of two template files, including the task template and the class domain template. The meta and data fields are then used to describe the dataset, with specific field explanations as follows:
\begin{itemize}
    \item \$dsdl-version: DSDL version information.
    \item \$import: Template import information, importing the detection task template and VOC's class domain.
    \item meta: Displays some meta-information about the dataset, such as the dataset name, creator, etc. Users can add other notes if desired.
    \item data: Contains the sample information saved according to the previously defined structure, specifically:
        \begin{itemize}
            \item sample-type: Type definition of the data, here using the ObjectDetectionSample class imported from the detection task template, specifying the class domain as VOCClassDom.
            \item sample-path: Path where samples are stored. If it is an actual path, the content of samples is read from that file; if it is \$local (as in this example), it is directly read from the data.samples field of this file.
            \item samples: Saves the sample information of the dataset. Note that this field is only effective when sample-path is \$local; otherwise, samples will preferentially read from the path specified in sample-path.
        \end{itemize}
\end{itemize}

\begin{tcolorbox}[title=Note]
If the current dataset contains a large number of images, storing all annotation information in the yaml file may lead to slow data loading. In such cases, users can provide an external file, such as specifying sample-path: train.json in train.yaml, to store data in a more efficiently readable file.
\end{tcolorbox}

\subsection{Image Segmentation}

To develop a template for describing datasets for image segmentation tasks, we have surveyed mainstream image segmentation datasets. Unlike classification and detection tasks, image segmentation usually includes three sub-tasks: semantic segmentation, instance segmentation, and panoptic segmentation. We discuss the objectives and common annotation fields of these sub-tasks, organize shared and unique fields, and based on this, create a universal template for dataset descriptions for these sub-tasks.

\subsubsection{Task Survey}

\noindent\textbf{Task Definition.}
Image segmentation involves semantically labeling each pixel in an image. Depending on slight variations in the labeling objectives, it can be divided into three sub-tasks:

\begin{itemize}
\item Semantic Segmentation: The basic form of image segmentation, which involves labeling each pixel of the image with a semantic category.
\item Instance Segmentation: Focuses on countable objects (such as people, cars, etc.), distinguishing and labeling different instances within the image.

\item Panoptic Segmentation: A combination of semantic and instance segmentation, often used in autonomous driving. It classifies objects into countable entities (things, e.g., people, cats, cars) and uncountable entities (stuff, e.g., sky, roads). Countable entities are labeled for both semantics and instances, while uncountable entities are labeled only semantically.
\end{itemize}

\begin{figure}[H]
\begin{center}
\includegraphics[width=0.7\textwidth]{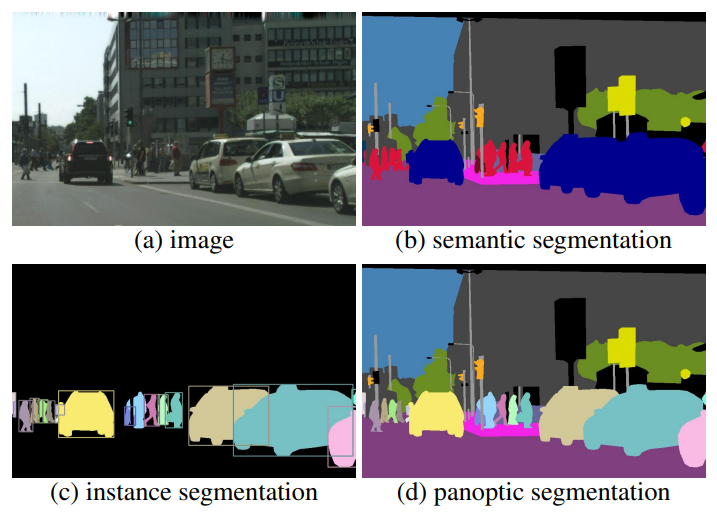}
\caption{Illustration of Image Segmentation}
\label{fig:seg1}
\end{center}
\end{figure}

Visual results of the three types of image segmentation are shown in~\Cref{fig:seg1} (image from the CVPR 2019 paper: \href{https://arxiv.org/pdf/1801.00868.pdf}{Panoptic Segmentation}).

\noindent\textbf{Evaluation Metrics.}
The evaluation metrics for the three sub-tasks of image segmentation slightly differ, as detailed below:

\textit{Semantic Segmentation.}
The evaluation metrics for semantic segmentation typically include mean Intersection over Union (mIoU) and mean Pixel Accuracy (mPA). Unlike detection, mIoU for semantic segmentation often involves irregular areas rather than rectangular boxes. Assuming there are \(k+1\) classes (including \(k\) object classes and 1 background class), where \(p_{ij}\) denotes the number of pixels that belong to class \(i\) but are predicted as class \(j\), with \(p_{ii}\) representing true positives, \(p_{ij}\) false positives, and \(p_{ji}\) false negatives, the formulas for mIoU and mPA are:

\begin{align*}
\text{mIoU} &= \frac{1}{k+1} \sum_{i=0}^{k} \frac{p_{ii}}{\sum_{j=0}^{k}p_{ij} + \sum_{j=0}^{k}p_{ji} - p_{ii}} \\
\text{mPA} &= \frac{1}{k+1} \sum_{i=0}^{k} \frac{p_{ii}}{\sum_{j=0}^{k}p_{ij}}
\end{align*}

\textit{Instance Segmentation.}
The evaluation metric for instance segmentation typically uses mean Average Precision (mAP), similar to object detection. The mIoU calculation involves the IoU between the predicted mask and the true mask, rather than the bounding boxes used in detection. For more information on mAP calculation in detection, refer to~\Cref{sec:detection}.

\textit{Panoptic Segmentation}
Panoptic segmentation uses a new metric called Panoptic Quality (PQ), which considers both things and stuff. PQ calculation involves two steps:

\begin{enumerate}
\item Segmentation Matching: A prediction is matched with a truth if their Intersection over Union (IoU) is over 0.5.
\item PQ Calculation: After matching, each segmentation prediction can be a True Positive (TP), False Positive (FP), or False Negative (FN). Given TP, FP, and FN, PQ for each category is calculated as follows:

\[
PQ = \frac{\sum_{(p,g) \in TP} \text{IoU}(p,g)}{|TP| + \frac{1}{2}|FP| + \frac{1}{2}|FN|}
\]

PQ can also be seen as the product of segmentation quality (SQ) and recognition quality (RQ):

\[
PQ = \underbrace{\frac{\sum_{(p,g) \in TP} \text{IoU}(p,g)}{|TP|}}_{\text{segmentation quality (SQ)}} \times \underbrace{\frac{|TP|}{|TP| + \frac{1}{2}|FP| + \frac{1}{2}|FN|}}_{\text{recognition quality (RQ)}}
\]

For more information on the PQ formula, refer to the original paper: \href{https://arxiv.org/pdf/1801.00868.pdf}{Panoptic Segmentation}.
\end{enumerate}

\subsubsection{Survey of Mainstream Datasets}
We have surveyed mainstream image segmentation datasets, including common datasets such as COCO~\cite{lin2014microsoft}, VOC~\cite{everingham2011the}, CityScapes~\cite{cordts2016cityscapes}, and ADE20K~\cite{zhou2017scene}. Considering that some datasets include different segmentation subtasks, we have divided the survey into three subtasks, focusing only on the annotation content relevant to each subtask. Additionally, during the survey, we encountered fields with different names but similar or identical meanings, which we treated as the same field and uniformly referred to them. For instance, \textit{image\_id} represents a unique image identifier, sometimes also represented by the image path. \textit{category\_id} also encompasses the meaning of \textit{category\_name}. Lastly, considering the unique annotation format of image segmentation (segmentation map, which annotates the original image in the form of an image, where each pixel value represents the annotation information at that position), \textit{category\_id} is sometimes not explicitly annotated but is derived from the pixel values in the segmentation map. Therefore, we have conceptualized a hypothetical field \textit{pixel\_mapping}, which describes the mapping from pixel values to actual \textit{label\_id} and \textit{instance\_id}, marked with an asterisk to indicate its difference, and this field appears bound with the \textit{segmentation\_map} field (some pixel values equal to \textit{label\_id} or \textit{instance\_id} can be considered as identity mapping).

\noindent\textbf{Semantic Segmentation Datasets.}
We surveyed three datasets for semantic segmentation tasks: VOC~\cite{everingham2011the}, CityScapes~\cite{cordts2016cityscapes}, and ADE20K~\cite{zhou2017scene}. The summarized information is shown in~\Cref{table:seg1}:

\begin{table}[H]
\centering
\begin{tabular}{@{}lccc@{}}
\toprule
\textbf{Semantic Segmentation Dataset} & \multicolumn{2}{c}{\textbf{Shared Fields}} & \textbf{Independent Field} \\
\cmidrule(lr){2-3}
 & \textbf{image\_id} & \textbf{segmentation\_map} & \textbf{pixel\_mapping*} \\
\midrule
VOC~\cite{everingham2011the} & Y & Y & Y \\
CityScapes~\cite{cordts2016cityscapes} & Y & Y & Y \\
ADE20K~\cite{zhou2017scene} & Y & Y & Y \\
\bottomrule
\end{tabular}
\caption{Survey results of semantic segmentation datasets}
\label{table:seg1}
\end{table}

In the table above, an asterisk (*) indicates a hypothetical field, meaning the information is present but not provided through an annotation file, as mentioned previously. The meanings of the other fields are as described in~\Cref{table:seg2}:

\begin{table}[H]
\centering
\begin{tabular}{@{}lcl@{}}
\toprule
\textbf{Field Type} & \textbf{Field Name} & \textbf{Meaning} \\
\midrule
\multirow{2}{*}{Shared Fields} & image\_id & Identifies a unique image, e.g., by image name or path \\
 & segmentation\_map & Segmentation map, the semantic mask for semantic segmentation tasks \\
\midrule
Independent Field & pixel\_mapping* & Mapping from pixel values in the segmentation map to semantic id/instance id \\
\bottomrule
\end{tabular}
\caption{Description of fields in semantic segmentation datasets}
\label{table:seg2}
\end{table}

\noindent\textbf{Instance Segmentation Datasets.}
We surveyed VOC~\cite{everingham2011the}, COCO~\cite{lin2014microsoft}, CityScapes~\cite{cordts2016cityscapes}, and ADE20K~\cite{zhou2017scene} datasets for instance segmentation tasks. The summarized information is shown in~\Cref{table:seg3}. The meanings of the various fields are shown in~\Cref{table:seg4}.

\begin{table}[H]
\centering
\begin{tabular}{@{}lccccccccccc@{}}
\toprule
\textbf{Dataset} & \textbf{Img ID} & \multicolumn{10}{c}{\textbf{Ind. Fields}} \\
\cmidrule(lr){3-12}
 & & \textbf{Seg Map} & \textbf{Px Map*} & \textbf{Poly/RLE} & \textbf{Cat ID} & \textbf{Inst ID} & \textbf{Area} & \textbf{Crowd} & \textbf{Occ} & \textbf{Parts} & \textbf{Scenes} \\
\midrule
VOC~\cite{everingham2011the} & Y & Y & Y & & & & & & & & \\
COCO~\cite{lin2014microsoft} & Y & & & Y & Y & Y & Y & Y & & & \\
CityScapes~\cite{cordts2016cityscapes} & Y & Y & Y & Y & Y & Y & & & & & \\
ADE20K~\cite{zhou2017scene} & Y & Y & Y & Y & Y & Y & & & Y & Y & Y \\
\bottomrule
\end{tabular}
\caption{Survey results of instance segmentation datasets. Abbreviations: Img ID (image\_id), Seg Map (segmentation\_map), Px Map (pixel\_mapping), Poly/RLE (polygon/rle\_polygon), Cat ID (category\_id), Inst ID (instance\_id), Crowd (iscrowd), Occ (occluded).}
\label{table:seg3}
\end{table}

\begin{table}[H]
\centering
\begin{tabular}{@{}lcl@{}}
\toprule
\textbf{Field Type} & \textbf{Field Name} & \textbf{Meaning} \\
\midrule
Shared Field & image\_id & Identifies a unique image, e.g., by image name or path \\
\midrule
\multirow{10}{*}{Independent Fields} & segmentation\_map & Mapping from pixel values in the segmentation map to semantic id/instance id \\
 & pixel\_mapping* & Segmentation map, the semantic mask for semantic segmentation tasks \\
 & polygon/rle\_polygon & Collection of vertex coordinates for polygonal annotations of individual objects \\
 & category\_id & Category id of the individual object \\
 & instance\_id & Instance id of the individual object \\
 & area & Area of the segmented region of the object \\
 & iscrowd & Indicates whether the object is a group of objects, such as a crowd of people, a bunch of apples, etc. \\
 & occluded & Indicates whether the object is occluded \\
 & parts & Part information, i.e., whether the object contains a part or is a part of another object \\
 & scenes & Scene category of each image \\
\bottomrule
\end{tabular}
\caption{Description of fields in instance segmentation datasets}
\label{table:seg4}
\end{table}

\noindent\textbf{Panoptic Segmentation Datasets}

We surveyed the panoptic segmentation tasks in the COCO and CityScapes datasets. The  summarized information is shown in~\Cref{table:segx}.

\begin{table}[H]
\centering
\caption{Instance Segmentation Datasets (Img ID: Image ID, Seg Map: Segmentation Map, Pix Map: Pixel Mapping, Poly/RLE: Polygon/RLE Polygon, Cat ID: Category ID, Inst ID: Instance ID, Crowd: Is Crowd, BBox: Bounding Box, Thing: Is Thing, Supercat: Supercategory)}
\begin{tabular}{@{}lccccccccccc@{}}
\toprule
\multirow{2}{*}{Dataset} & \multicolumn{2}{c}{Shared} & \multicolumn{9}{c}{Independent} \\
\cmidrule(lr){2-3} \cmidrule(lr){4-12}
 & Img ID & Seg Map & Pix Map* & Poly/RLE & Cat ID & Inst ID & Area & Crowd & BBox & Thing & Supercat \\
\midrule
COCO~\cite{lin2014microsoft}        & Y & Y & Y & Y & Y & Y & Y & Y & Y & Y & Y \\
CityScapes~\cite{cordts2016cityscapes}  & Y & Y & Y & Y & Y & Y &   &   &   &   &   \\
\bottomrule
\label{table:segx}
\end{tabular}
\end{table}

\begin{tcolorbox}[title=Note]
The fields `Is Thing' and `Supercategory' describe class domains, i.e., they describe categories rather than individual annotations.
\end{tcolorbox}

\begin{table}[h]
\centering
\caption{Detailed Descriptions of Fields}
\begin{tabular}{@{}llp{10cm}@{}}
\toprule
Field Type & Field Name & Description \\
\midrule
\multirow{2}{*}{Shared Fields} 
& image\_id & Identifies a unique image, e.g., by image name or path. \\
& segmentation\_map & Maps pixel values in the segmentation image to semantic/instance IDs. \\
\midrule
\multirow{8}{*}{Independent Fields} 
& pixel\_mapping* & Segmentation image, semantic segmentation task corresponds to the segmentation mask. \\
& polygon/rle\_polygon & Collection of vertex coordinates for polygon annotations of individual objects. \\
& category\_id & Category ID of an individual object. \\
& instance\_id & Instance ID of an individual object. \\
& area & Area of the segmented region of an object. \\
& iscrowd & Indicates if the object is part of a group, such as a crowd of people or a bunch of apples. \\
& bbox & Bounding box of the object. \\
& isthing & Indicates whether the object is a 'thing' (countable, for instance annotation) or 'stuff' (uncountable, for semantic annotation only). \\
& supercategory & Parent category of the category, e.g., 'animal' might be the supercategory for 'cat'. \\
\bottomrule
\end{tabular}
\end{table}

\noindent\textbf{Research on Segmentation Maps.}
From the previous discussion, it is evident that segmentation maps are the primary annotation tools in image segmentation, but different datasets have different formats for these maps. We researched the segmentation maps and their corresponding pixel mappings to facilitate the standardization of segmentation map formats. Common scenarios include:

\textbf{Basic Form:}
In this form, the segmentation map is single-channel with integer pixel values. For semantic segmentation, each pixel represents the semantic category at that location. For instance segmentation, each pixel represents the instance category. The VOC dataset uses this approach for both its semantic and instance segmentation tasks.

\begin{center}
\includegraphics[width=0.7\textwidth]
{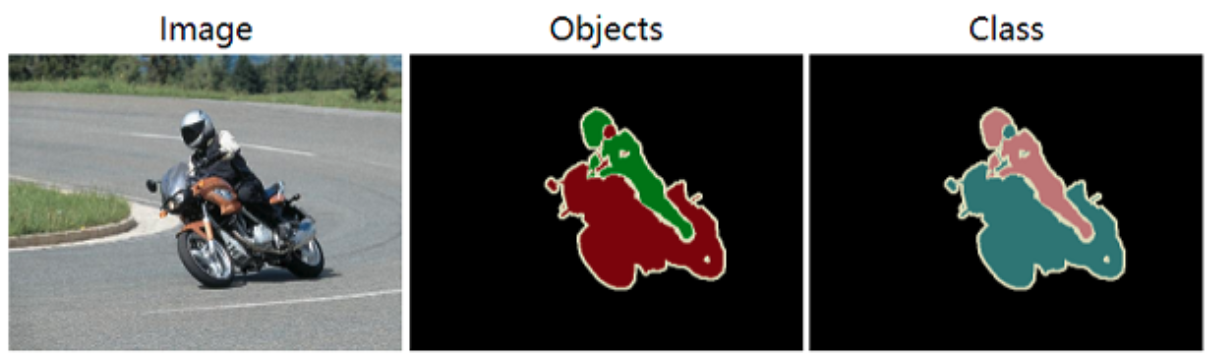}
\end{center}

\textbf{Semantic or Instance Categories Derived Through a Mapping:}
   Here, the segmentation map might be single or multi-channel, and the categories are computed through a certain relationship, as seen in CityScapes, ADE20K, and COCO Panoptic datasets:
   
\noindent - CityScapes:
\begin{minted}{python}
original_mask[original_mask < 24000] = 0
pix_values = np.unique(original_mask)[1:]
ins_attr_list = []

if pix_values.shape[0] > 0:
    for index, pix_value in enumerate(np.nditer(pix_values), 1):
        category = pix_value // 1000
        ins_attr_list.append(
        {
            "instance_id": int(index),
            "category_id": _CITYSCAPES_SEGMENTATION_CLASSES[category][1]
        }
    )
\end{minted}

\noindent - ADE20K:
\begin{minted}{python}
# Obtain the segmentation mask, built from the RGB channels of the _seg file
R = seg[:, :, 0]
G = seg[:, :, 1]
B = seg[:, :, 2]

ObjectClassMasks = (R/10).astype(np.int32)*256 + G.astype(np.int32)

# Obtain the instance mask from the blue channel of the _seg file
Minstances_hat = np.unique(B, return_inverse=True)[1]
Minstances_hat = np.reshape(Minstances_hat, B.shape)
ObjectInstanceMasks = Minstances_hat
\end{minted}

\noindent - COCO Panoptic:
\begin{minted}{python}
def rgb2id(color):
    if isinstance(color, np.ndarray) and len(color.shape) == 3:
        if color.dtype == np.uint8:
            color = color.astype(np.int32)
    return int(color[0] + 256 * color[1] + 256 * 256 * color[2])
\end{minted}

\textbf{Multiple Segmentation Maps for Instance Segmentation:}
   Sometimes, in instance segmentation tasks, an image may provide multiple instance maps, each representing one instance in single-channel format, where 0 indicates background and x indicates the instance with its semantic category, as seen in ADE20K:

   \includegraphics[width=0.7\textwidth]{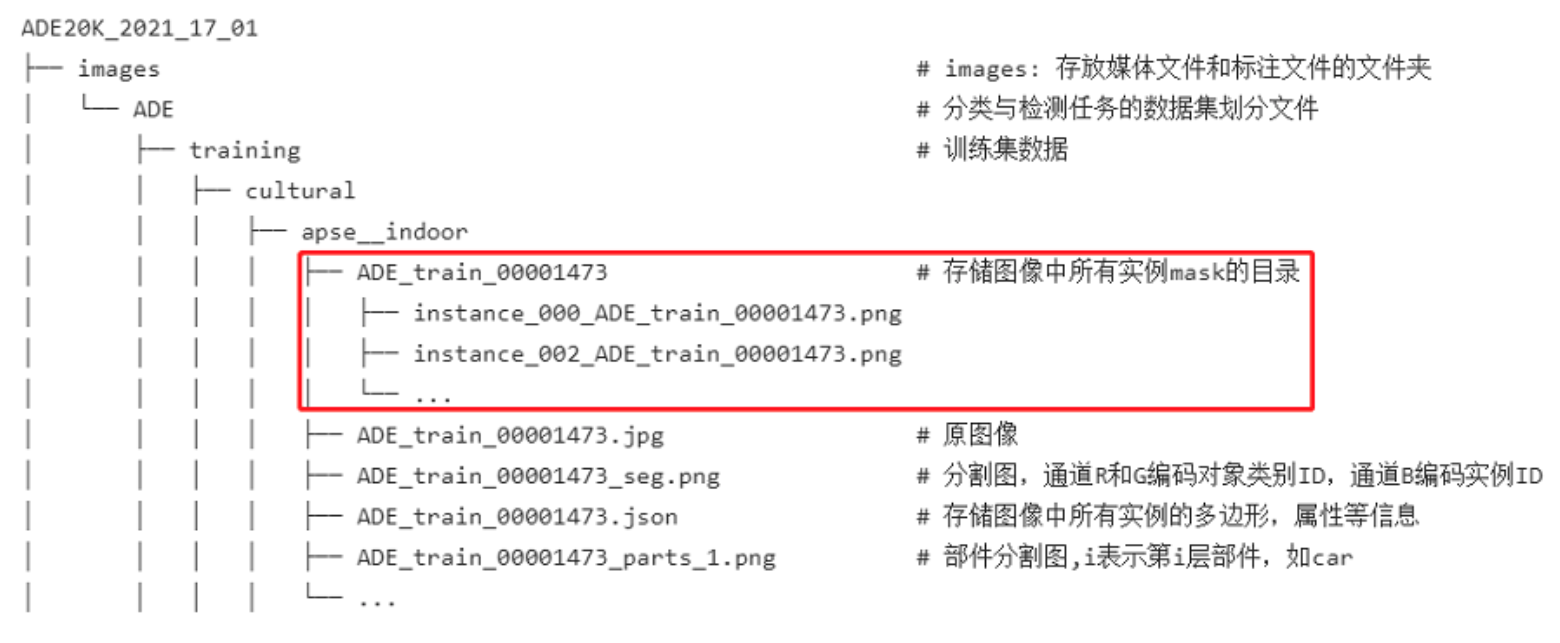}

\subsubsection{Template Display}

Based on the research above, we can preliminarily outline several scenarios for the annotation types of segmentation tasks, as shown in~\Cref{table:seg5}. This allows for the formulation of templates.

\begin{table}[H]
\centering
\caption{Types of Image Segmentation and Their Corresponding Annotations}
\label{table:seg5}
\resizebox{0.8\linewidth}{!}{ 
\setlength{\tabcolsep}{10pt} 
\renewcommand{\arraystretch}{1.5} 
\begin{tabular}{@{}lcc@{}}
\toprule
\textbf{Type of Image Segmentation} & \multicolumn{2}{c}{\textbf{Annotations}} \\
\cmidrule{2-3}
& \textbf{Method 1} & \textbf{Method 2} \\
\midrule
Semantic Segmentation & semantic\_map & - \\
Instance Segmentation & instance\_map & polygon \\
Panoptic Segmentation & instance\_map + semantic\_map & polygon + semantic\_map \\
\bottomrule
\end{tabular}
}
\end{table}

\noindent\textbf{Standards for Segmentation Maps.}
Before establishing task templates, it is essential to define the templates for segmentation maps. Based on the research on segmentation maps, we have established two standards to accommodate both semantic segmentation and instance segmentation tasks (panoramic segmentation is considered a combination of these two tasks):
\begin{itemize}
    \item \textbf{LabelMap}: Represents a semantic map. LabelMap is a single-channel, int32 type image that requires a class domain parameter. The pixel values in LabelMap represent the index of the class in the class domain, thus achieving mapping from pixels to semantic categories.
    \item \textbf{InstanceMap}: Represents an instance map. InstanceMap is also a single-channel, int32 type image. The matrix elements represent the location of instances, while semantic information is obtained through the corresponding LabelMap (InstanceMap itself does not require a class domain parameter).
\end{itemize}

\textit{Note: Both LabelMap and InstanceMap represent a single-channel image, but to reduce the size of the annotation files, the image can be saved and the paths to these standard-compliant image files can be stored in LabelMap and InstanceMap.}

With these standards for segmentation maps, we can establish guidelines for three types of segmentation tasks.

\noindent\textbf{Template for Semantic Segmentation}

From the analysis above, we know that the annotation for semantic segmentation only involves the LabelMap, and one image corresponds to one LabelMap. Hence, the template for semantic segmentation is as follows:

\begin{minted}{yaml}
SemanticSegmentationSample:
    $def: struct
    $params: ['cdom']
    $fields:
        image: Image
        semantic_map: LabelMap[dom=$cdom]
\end{minted}

\noindent\textbf{Template for Instance Segmentation} \label{Sec:seg template}

The annotation for instance segmentation typically includes instance map and polygon types. Therefore, the templates for instance segmentation are defined as follows:

\begin{itemize}
    \item For instance map type:
\end{itemize}

\begin{minted}{yaml}
InstanceSegmentationSample:
    $def: struct
    $params: ['cdom']
    $fields:
        image: Image
        instance_map: InstanceMap
        semantic_map: LabelMap[dom=$cdom]
\end{minted}

\begin{itemize}
    \item For polygon type:
\end{itemize}

\begin{minted}{yaml}
LocalInstanceEntry:
    $def: struct
    $params: ['cdom']
    $fields:
        label: Label[dom=$cdom]
        bbox: BBox
        polygon: Polygon

InstanceSegmentationSample:
    $def: struct
    $params: ['cdom']
    $fields:
        image: Image
        instances: List[LocalInstanceEntry[cdom=$cdom]]
\end{minted}

Users can choose the appropriate template based on their specific needs.

\noindent\textbf{Template for Panoramic Segmentation.}

Similar to instance segmentation, the instance information in panoramic segmentation can also be given in two ways. The templates are defined as follows:

\begin{itemize}
    \item Instance information through instance map:
\end{itemize}

\begin{minted}{yaml}
PanopticSegmentationSample:
    $def: struct
    $params: ['cdom']
    $fields:
        image: Image
        instance_map: InstanceMap
        semantic_map: LabelMap[dom=$cdom]
\end{minted}

\begin{itemize}
    \item Instance information through polygon:
\end{itemize}

\begin{minted}{yaml}
LocalInstanceEntry:
    $def: struct
    $params: ['cdom']
    $fields:
        label: Label[dom=$cdom]
        bbox: BBox
        polygon: Polygon

PanopticSegmentationSample:
    $def: struct
    $params: ['cdom']
    $fields:
        image: Image
        semantic_map: LabelMap[cdom=$cdom]
        instances: List[LocalInstanceEntry[cdom=$cdom]]
\end{minted}

Users can choose the appropriate template based on their specific needs.

Additionally, because panoramic segmentation distinguishes between 'things' and 'stuff' categories, it is necessary to provide additional information on the class domain, as shown below:

\begin{minted}{yaml}
$dsdl-version: "0.5.0"

COCOClassDom:
    $def: class_domain
    classes:
        - stuff.sky
        - things.horse
        - things.person
        - things.bottle
        - ...
\end{minted}

\subsection{Key Point Detection}

\subsubsection{Task Research}

\noindent\textbf{Task Definition.}
The goal of the keypoint detection task is to identify critical parts of an object, while the goal of the pose estimation task is to estimate the posture of an object (usually humans or animals), that is, the connections between keypoints. Keypoint detection and pose estimation are often discussed together because the connections between body parts, such as those of humans and animals, are fixed. Once the keypoints of the human body are detected, the results of pose estimation can be obtained (whether there is pose estimation depends on whether there are connections between keypoints).

\begin{figure}[H]
    \centering
    \includegraphics[width=0.8\textwidth]{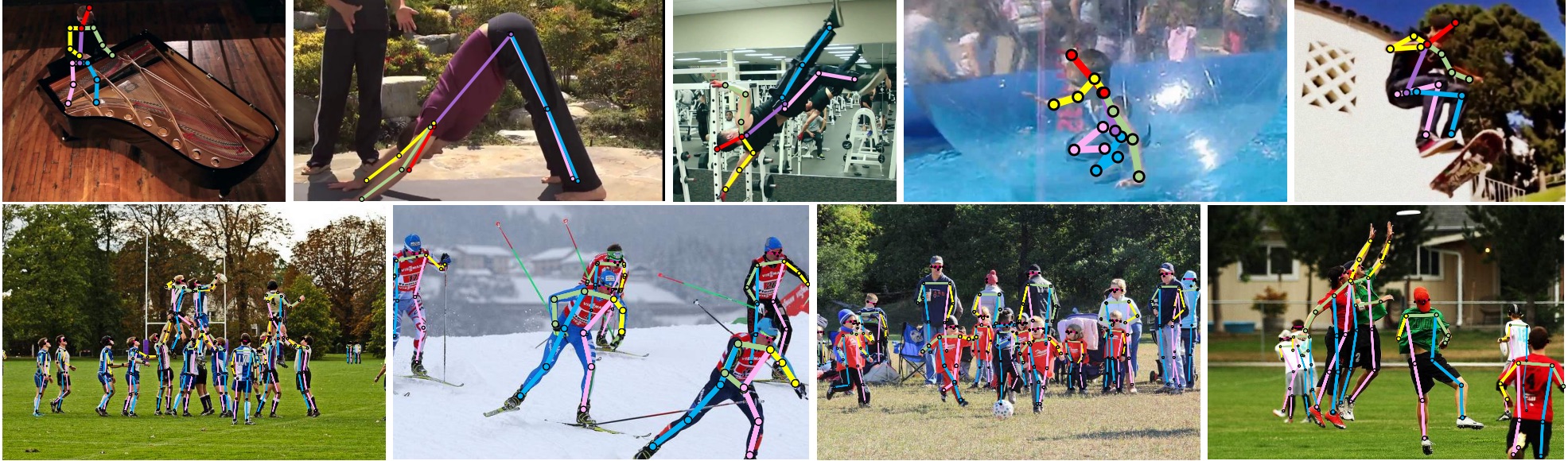}
    \caption*{Image source: Deep High-Resolution Representation Learning for Human Pose Estimation, CVPR19}
\end{figure}

\noindent \textbf{Evaluation Metrics.}
The evaluation metric typically used is the COCO-style mAP. The calculation of mAP for keypoint detection is similar to that in COCO object detection. For all detected objects and their keypoints, the OKS metric is first used to classify detections into TP, FP, and FN. After classification, by changing the score threshold, a Precision-Recall (P-R) curve is computed, and the area under this P-R curve represents the AP value.

The major difference from object detection mAP calculation is that object detection uses the IOU between detection boxes to measure instance similarity, whereas keypoint detection uses the OKS distance between object keypoints. The calculation of OKS is as follows:

\begin{figure}[H]
    \centering
    \includegraphics[width=0.8\textwidth]{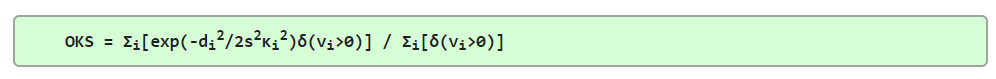}
    \caption*{Image source: \url{https://cocodataset.org/\#keypoints-eval}}
\end{figure}

OKS represents the similarity between all detected keypoints (predictions) and the actual annotations (ground truth) for an object. Here, \(d_i\) represents the Euclidean distance between the \(i\)-th detected keypoint and its corresponding ground truth, \(s\) is the pixel area of the object, and \(k\) is a normalization factor for the \(i\)-th type of keypoint (e.g., nose), calculated across all instances of that keypoint in the dataset. A higher \(k\) value indicates poorer annotation quality for that keypoint, making it harder to detect, while a lower \(k\) value indicates better annotation quality and thus easier detection. With the OKS distance, it is possible to compute the AP metric at different OKS thresholds. The COCO keypoint detection metrics are defined similarly to those in object detection:

\begin{figure}[H]
    \centering
    \includegraphics[width=0.8\textwidth]{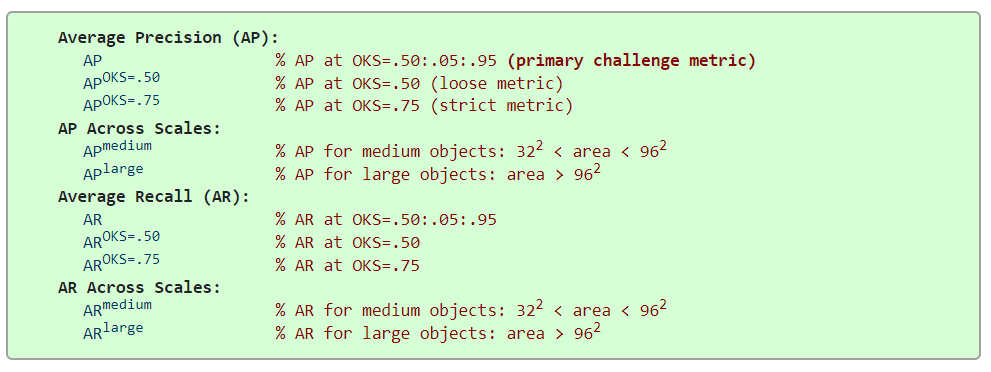}
    \caption*{Image source: \url{https://cocodataset.org/\#keypoints-eval}}
\end{figure}

\noindent \textbf{Mainstream Dataset Research.}
Based on the type of target, the annotation format of datasets for pose estimation (keypoint detection) varies. Datasets for pose estimation (keypoint detection) can be categorized by target type as follows:

\begin{table}[ht]
\centering
\caption{Datasets categorized by target type for pose estimation and keypoint detection.}
\resizebox{0.9\linewidth}{!}{
\begin{tabular}{ccc}
\toprule
\textbf{Target Type} & \textbf{Task Type} & \textbf{Representative Dataset} \\
\midrule
Human Body & Human Body Keypoint Detection & COCO~\cite{lin2014microsoft}, MPII~\cite{andriluka20142d}, MPII-TRB~\cite{duan2019trb}, AI Challenger~\cite{wu2017ai}, CrowdPose~\cite{li2019crowdpose}, OCHuman~\cite{zhang2019pose2seg} \\
Human Wholebody & Human Wholebody Keypoint Detection & COCO WholeBody~\cite{jin2020whole}, Halpe~\cite{fang2017rmpe} \\
Face & Face Keypoint Detection & 300W~\cite{zhu2016face}, WFLW~\cite{wu2018look}, AFLW~\cite{zhu2016unconstrained}, COFW~\cite{burgos2013robust}, COCO-WholeBody-Face~\cite{jin2020whole} \\
Hand & Hand Keypoint Detection & OneHand-10K~\cite{wang2018mask}, FreiHand~\cite{zimmermann2019freihand}, CMU Panoptic HandDB~\cite{simon2017hand} \\
Fashion & Fashion Landmark Detection & DeepFashion~\cite{jiang2022text2human} \\
Animal & Animal Keypoint Detection & Animal-Pose~\cite{cao2019cross}, AP-10K~\cite{yu2021ap}, Horse-10~\cite{mathis2021pretraining}, MacaquePose~\cite{labuguen2021macaquepose}, Vinegar Fly~\cite{graving2019fast} \\
\bottomrule
\end{tabular}
}    
\end{table}

We researched 10 mainstream pose estimation/keypoint detection datasets, covering all the different types mentioned above. The complete field research results for these datasets are shown in the following table:

\begin{table}[ht]
\centering
\caption{Field research results for mainstream pose estimation/keypoint detection datasets. Abbreviations: img\_id (Image ID), ht (Height), wd (Width), inst\_id (Instance ID), cat\_id (Category ID), crowd (Is Crowd), area, num\_kpts (Number of Keypoints), bbox (Bounding Box), seg (Segmentation), kpts (Keypoints), vis (Visibility), ctr (Center), cats (Categories), sup\_cats (Super Categories), kpt\_names (Keypoint Names), skel (Skeleton).}
\resizebox{\linewidth}{!}{
\setlength{\tabcolsep}{5pt}
\begin{tabular}{ccccccccccccccccccc}
\toprule
\textbf{Dataset} & \textbf{img\_id} & \textbf{ht} & \textbf{wd} & \textbf{inst\_id} & \textbf{cat\_id} & \textbf{crowd} & \textbf{area} & \textbf{num\_kpts} & \textbf{bbox} & \textbf{seg} & \textbf{kpts} & \textbf{vis} & \textbf{ctr} & \textbf{cats} & \textbf{sup\_cats} & \textbf{kpt\_names} & \textbf{skel} & \textbf{other} \\
\midrule
COCO~\cite{lin2014microsoft} & Y & Y & Y & Y & Y & Y & Y & Y & Y & Y & Y & Y & & Y & Y & Y & Y & \\
MPII~\cite{andriluka20142d} & Y & & & & & & & & & & Y & Y & Y & & & & & scale, person, torsoangle \\
AIC~\cite{wu2017ai} & Y & Y & Y & Y & Y & Y & Y & Y & Y & & Y & Y & & Y & Y & Y & Y & \\
CrowdPose~\cite{li2019crowdpose} & Y & Y & Y & Y & Y & Y & Y & Y & Y & & Y & Y & & Y & Y & Y & Y & crowd index \\
COCO-WholeBody~\cite{jin2020whole} & Y & Y & Y & Y & Y & Y & Y & Y & Y & & Y & Y & & Y & Y & Y & Y & face\&hand valid/kpts/bbox\\
Halpe~\cite{fang2017rmpe} & Y & Y & Y & Y & Y & Y & & Y & & & Y & Y & & Y & Y & & & Hoi \\
300W~\cite{zhu2016face} & Y & Y & Y & Y & Y & Y & Y & Y & Y & & Y & Y & Y & Y & Y & & & \\
OneHand10K~\cite{wang2018mask} & Y & Y & Y & Y & Y & Y & Y & & Y & Y & Y & Y & & Y & Y & Y & Y & \\
DeepFashion~\cite{jiang2022text2human} & Y & & & & Y & & & & Y & & Y & Y & & Y & & Y & & variation \\
AnimalPose~\cite{cao2019cross} & Y & & & & Y & & & Y & Y & & Y & Y & & Y & Y & Y & Y & \\
\bottomrule
\end{tabular}
}
\end{table}

The fields for keypoint detection and pose estimation tasks are categorized into shared fields, common across datasets, and independent fields, specific to particular datasets. Below is a table detailing these fields:

\begin{table}[ht]
\centering
\caption{Shared and Independent Fields in Keypoint Detection/Pose Estimation Tasks}
\begin{tabular}{llp{8cm}}
\toprule
\textbf{Field Type} & \textbf{Field Name} & \textbf{Description} \\
\midrule
\multicolumn{1}{c}{\multirow{3}{*}{Shared Fields}} & image\_id & Unique image identifier, such as name or path. \\
 & keypoints & Coordinates of object keypoints, formatted as [x, y] or [x, y, vis] (vis denotes visibility). \\
 & visible & Visibility of a keypoint, indicated by an integer. \\
\midrule
\multicolumn{1}{c}{\multirow{22}{*}{Independent Fields}} & height/width & Original image dimensions. \\
 & instance\_id & Unique identifier for each object. \\
 & category\_id & Identifier for object's category. \\
 & is\_crowd & Indicates single or multiple objects (0 for single, 1 for multiple). \\
 & area & Area occupied by the object in pixels. \\
 & num\_keypoints & Total number of keypoints. \\
 & bbox & Bounding box coordinates [x, y, w, h]. \\
 & segmentation & Pixel-level object segmentation, as [x, y] coordinates. \\
 & center & Center point of the object's bounding box [x, y]. \\
 & categories & Categories and their IDs in the dataset. \\
 & super\_categories & Parent categories of the object categories. \\
 & keypoint\_names & Names of the keypoints. \\
 & skeleton & Connections between keypoints. \\
 & scale & Scaling factor for bounding box (specific to MPII). \\
 & person & Number of people in the image (specific to MPII). \\
 & torsoangle & Torsion angle of the human torso (specific to MPII). \\
 & face/hand/foot valid & Indicates presence of face/hands/feet annotations (0 or 1, specific to COCO-WholeBody). \\
 & face/hand/foot kpts & Keypoint annotations for face/hands/feet. \\
 & face/hand bbox & Bounding box for face/hands [x, y, h, w]. \\
 & Hoi & Type of human-object interaction (specific to Halpe). \\
 & Variation & Pose variation (specific to DeepFashion). \\
\bottomrule
\end{tabular}
\end{table}

In summary, to describe a keypoint detection dataset, the essential fields include image\_id, keypoints, and visible. Users can add or modify dataset-specific fields as needed.

\subsubsection{Template Display}

Based on our research findings, for keypoint detection/pose estimation tasks, the most crucial attributes of a sample include the image ID (or path), the keypoint annotations for each object, and the visibility of these keypoints. Given that each image may contain multiple objects with multiple keypoint annotations, we have defined a nested structure called \texttt{KeyPointLocalObject} to represent the keypoint annotation information for individual targets (i.e., category and keypoints). The keypoint detection/pose estimation task structure's \texttt{\$fields} attribute defines two fields: \texttt{image} and \texttt{annotations}, where \texttt{annotations} is a list composed of multiple \texttt{KeyPointLocalObject} structures (an empty list indicates no objects with keypoint annotations in the image). Finally, considering the representativeness and scalability of the template, some attributes are mandatory, while others, specific to certain datasets, are optional. The template for the keypoint detection/pose estimation task is presented as follows:

\begin{minted}{yaml}
KeypointClassDom:
    $def: class_domain
    classes:
        - person
        - 
        - ...
        
KeypointDescDom:
    $def: class_domain
    classes:
       - "left eye"
       -
       - ...  
    skeleton:
       - [14, 16]
       - [5, 6]
       - [10, 12]
       - ...
\end{minted}

This section defines the task category domain file, including:

\begin{itemize}
    \item \texttt{KeypointClassDom}, which defines the domain of target categories, such as "person".

    \item \texttt{KeypointDescDom}, which defines domains pre-defined for the keypoint detection task, including keypoint names and the connections between them (\texttt{skeleton}).
\end{itemize}

\begin{minted}{yaml}
KeyPointLocalObject:   
    $def: struct
    $params: ['cdom1', 'cdom2']
    $fields:
        keypoint: Keypoint[dom=$cdom2]
        label: Label[dom=$cdom1]
    $optional: ["label"]
        
KeyPointSample:
    $def: struct
    $params: ['cdom1', 'cdom2']
    $fields:
        image: Image
        annotations: List[etype=LocalObjectEntry[cdom1=$cdom1, cdom2=$cdom2]]
        
data:
    sample-type: ObjectKeypointSample[cdom1=KeypointClassDom, cdom2=KeypointDescDom]
\end{minted}

The section above defines the YAML file for a keypoint detection sample, including:

\begin{itemize}
    \item \texttt{KeyPointSample}, which defines a sample object in keypoint detection, including the image path \texttt{image} and a list of annotated targets \texttt{annotations}.
    
    \item \texttt{KeyPointLocalObject}, which defines an annotation for a target, including:
   - The category of the target \texttt{label}, which belongs to the domain \texttt{KeypointClassDom}, i.e., the category of the target.
   - The keypoint annotation \texttt{keypoint}, which uses a list [x1, y1, v1, x2, y2, v2, ...] to represent the keypoints, where x1, y1 represent the coordinates of a keypoint, and v1 represents the visibility of the keypoint (the visibility annotation does not have a unified standard and maintains consistency with the original dataset format, considering values <=0 as invisible and unmarked, >1 as marked). The domain of \texttt{keypoint} is \texttt{KeypointDescDom}, which describes the keypoint domain.
\end{itemize}

\subsubsection{Usage Method}

This section describes how to use the previously defined template to describe a dataset, taking COCOKeypoint2017 as an example. The YAML file for the sample, named \texttt{keypoint-coco2017.yaml}, is shown below:

\begin{minted}{yaml}
$dsdl-version: "0.5.0"

KeyPointLocalObject:
    $def: struct
    $params: ["cdom0", "cdom1"]
    $fields:
        iscrowd: Int
        area: Num
        category: Label[dom=$cdom0]
        bbox: BBox
        polygon: Polygon
        num_keypoints: Int
        ann_id: Int
        keypoints: Keypoint[dom=$cdom1]

KeyPointSample:
    $def: struct
    $params: ["cdom0", "cdom1"]
    $fields:
        media: Image
        height: Int
        width: Int
        image_id: Int
        annotations: List[etype=KeyPointLocalObject[cdom0=$cdom0, cdom1=$cdom1]]
\end{minted}

The COCO2017 Keypoints dataset template includes several dataset-specific fields in addition to the mandatory fields (keypoints, visibility, and image\_id) defined in the keypoint detection task template. The meanings of some fields in the detection template are as follows:

\begin{itemize}
    \item \texttt{\$dsdl-version}: Describes the version of the DSDL file.
    \item \texttt{ObjectKeypointEntry}: Defines a nested structure for describing keypoint annotations, consisting of:
    \begin{itemize}
        \item \texttt{\$def: struct}, indicating that this is a structure type.
        \item \texttt{\$params}: Defines the parameters, here referring to the class domain.
        \item \texttt{\$fields}: Contains properties of the structure, including:
        \begin{itemize}
            \item \texttt{iscrowd}: Indicates whether the annotation is for a single object or multiple objects.
            \item \texttt{area}: The pixel area of the target instance.
            \item \texttt{category}: The category of the object.
            \item \texttt{bbox}: The bounding box annotation for the target, of type BBox.
            \item \texttt{polygon}: The instance segmentation annotation for the target, of type Polygon.
            \item \texttt{num\_keypoints}: Represents the number of keypoints for the object.
            \item \texttt{ann\_id}: The annotation ID within the dataset.
            \item \texttt{keypoints}: A list of keypoints, where each element is a [x, y, z] coordinate, x and y denote the keypoint position, and z denotes the visibility of the keypoint.
        \end{itemize}
    \end{itemize}
\end{itemize}

The YAML file describing the keypoint detection task's class domain is shown below:

\begin{minted}{yaml}
$dsdl-version: "0.5.0"

COCO2017KeypointsClassDom:
    $def: class_domain
    classes: 
        - person

COCO2017KeypointsDescDom[COCO2017KeypointsClassDom]:
    $def: class_domain
    classes:
        - nose[person]
        - left_eye[person]
        - right_eye[person]
        - ...
    skeleton:
        - [16, 14]
        - [14, 12]
        - [17, 15]
        - ...
\end{minted}

This file defines the category domain for the keypoint detection task, specifically including:
\begin{itemize}
    \item \texttt{COCO2017KeypointsClassDom}: The class domain for COCO2017 keypoint detection, containing only the category "person".
    \item \texttt{COCO2017KeypointsDescDom}: Describes the domain for the COCO2017 keypoint detection dataset, inheriting from the above class domain, including descriptions of "person" in the dataset, such as keypoint names and the connections between keypoints.
\end{itemize}

\subsection{Object Tracking}

\subsubsection{Task Research}

\begin{figure}[ht]
    \centering
    \begin{subfigure}[b]{0.4\textwidth}
        \centering
        \includegraphics[width=\linewidth]{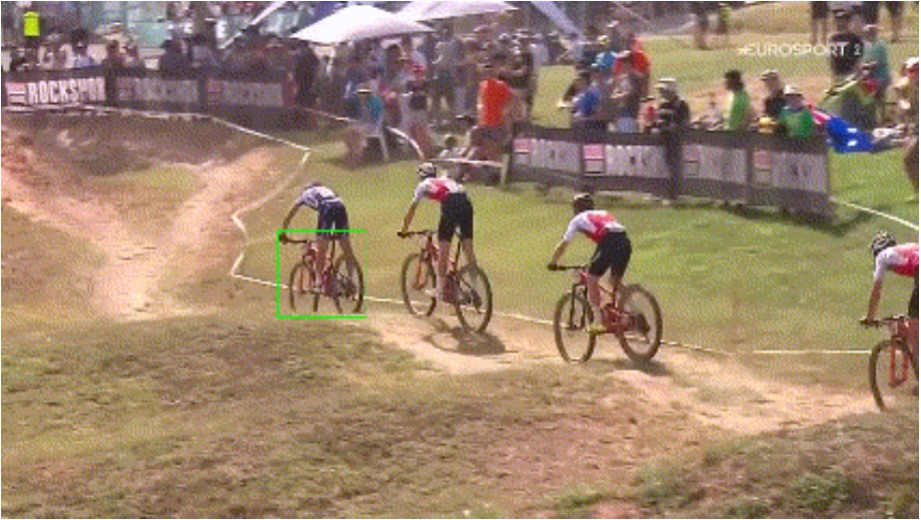}
        \caption{Single Object tracking.}
        \label{fig:track_sub1}
    \end{subfigure}
    \begin{subfigure}[b]{0.4\textwidth}
        \centering
        \includegraphics[width=\linewidth]{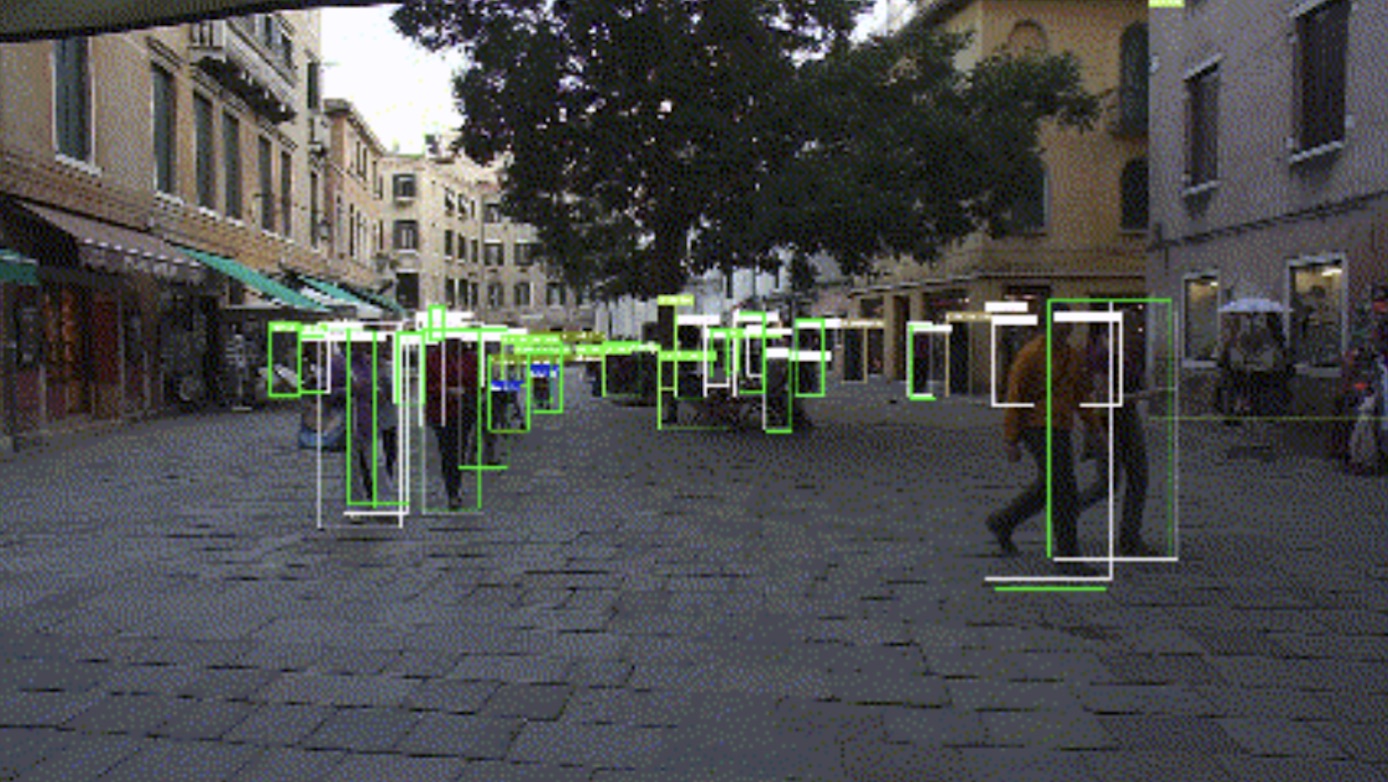}
        \caption{Multiple object tracking.}
        \label{fig:track_sub2}
    \end{subfigure}
    \caption{Illustration of Obeject Tracking.}
    \label{fig:track1}
\end{figure}

\noindent\textbf{Task Definition}
Object tracking tasks involve detecting the location of objects in images, identifying specific instances, and tracking them through a unique identification ID. It is divided into single-object tracking and multi-object tracking, and some datasets also annotate the category of each instance. The task shows in~\Cref{fig:track1}.

\noindent \textbf{Evaluation Metrics.}
In object tracking tasks, the most commonly used evaluation metrics are precision and success rate.

\begin{itemize}
    \item \textbf{Success (Success Rate/IOU Rate/AOS).}
The calculation of the Success Rate involves computing the intersection-over-union (IoU) ratio of the predicted bounding box with the annotated bounding box. The success rate is the proportion of successful instances when IoU reaches a certain threshold. The success rate will vary with different thresholds. With the threshold denoted as $x$ and the success rate as $y$, a success rate plot can be drawn. The AUC (Area under the curve) score represents the area under the success rate curve. Some papers may directly specify a threshold, and due to the median theorem, the most commonly used threshold is 0.5.
    \item \textbf{Precision.}
Precision is the proportion of successful tracking instances. To calculate the number of successful tracking instances, the Euclidean distance between the center points of the predicted bounding box and the annotated bounding box must be calculated. Typically, the threshold is set at 20 pixels, meaning that if their Euclidean distance is within 20 pixels, it is considered a successful tracking.
    \item \textbf{Normalized Precision.}
Considering that the scale of the annotated bounding box will affect the judgment of precision (for example, for a smaller annotated box, if the center points of the predicted and annotated boxes are 20 pixels apart, the intersection-over-union ratio has already dropped to a very low value), the Precision is normalized based on the size of the annotated box, resulting in Normalized Precision.
\end{itemize}

\noindent\textbf{Mainstream Dataset Research}
We conducted research on four object detection datasets, analyzed and summarized the relevant dataset description files (mainly annotation fields), and displayed the annotation fields with the same meaning with a unified naming convention. The summary information is shown in~\Cref{table:track1}.

\begin{table}[H]
    \centering
    \resizebox{\linewidth}{!}{ 
    \begin{tabular}{c|ccccc|ccccccccc|}
    \toprule
    \textbf{Dataset} & \multicolumn{5}{c|}{\textbf{Shared Fields}} & \multicolumn{8}{c}{\textbf{Unique Fields}} \\
    \midrule
    & \textbf{instance\_id} & \textbf{bbox} & \textbf{category} & \textbf{media\_path} & \textbf{frame\_id} & \textbf{width} & \textbf{height} & \textbf{frame\_rate} & \textbf{seq\_length} & \textbf{absence} & \textbf{visibility} & \textbf{truncated} & \textbf{ignore} \\
    \midrule
    TrackingNet~\cite{muller2018trackingnet} & & Y & & Y & Y & & & & & & & & \\
    GOT10k~\cite{Huang2021} & & Y & Y & Y & Y & Y & Y & Y & & Y & Y & Y & \\
    MOT17~\cite{milan2016mot16} & Y & Y & Y & Y & Y & Y & Y & Y & Y & & Y & & Y \\
    KITTI-Tracking~\cite{Geiger2012CVPR} & Y & Y & Y & Y & Y & & & & & & Y & Y & \\
    \bottomrule
    \end{tabular}
    }
    \caption{Summary of shared and unique fields in object tracking datasets}
\label{table:track1}
\end{table}

The meanings of shared and unique fields are further summarized in~\Cref{table:track2}:
\begin{table}[H]
    \centering
    \begin{tabular}{@{}clp{10cm}@{}} 
    \toprule
    \textbf{Field Type} & \textbf{Field Name} & \textbf{Meaning} \\
    \midrule
    \multirow{5}{*}{Shared Fields} & instance\_id & Identifies an unique object. The same target has a unique number throughout the entire video clip. \\
                                   & bbox      & Bounding box for a single object, e.g., [xmin, ymin, xmax, ymax] \\
                                   & category      & Category of the target. \\
                                   & media\_path      & Media file path. \\
                                   & frame\_id      & Frame number, used for ordering in video sequences. \\
    \midrule
    \multirow{8}{*}{Unique Fields} & width & The width of the frame. \\
                                   & height & The height of the frame. \\
                                   & frame\_rate & Frame rate, some datasets also refer to it as anno\_fps. \\
                                   & seq\_length  & The number of frames in the video sequence. \\
                                   & absence & Indicates whether the frame contains the object. \\
                                   & visibility & Degree of occlusion. Different datasets have different representations: it can be "visibility" (the degree of visibility of the object, with values between 0 and 1), "cover" (indicating the level of obstruction, with a range from 0 to 8),  or "occluded" (indicating whether the annotation is obstructed. 0 means "fully visible"; 1 means "partly occluded"; 2 means "largely occluded"; 3 means "unknown"). \\
                                   & truncated & Indicates whether the object being annotated is currently truncated by the image edge, with 1 indicating it is truncated. \\
                                   & ignore  & Indicates whether the current annotation is considered in the evaluation. If it is 1, then the current annotation is considered; if it is 0, then ignored. \\
    \bottomrule
    \end{tabular}
    \caption{Detailed description of fields in object tracking datasets}
\label{table:track2}
\end{table}

\subsubsection{Template Display}
Object tracking tasks are an extension of object detection tasks and also include nested structures and a class domain. However, what differs is that: based on the above research results, we know that important attributes for object tracking tasks include frame\_id, media\_path, instance\_id, bbox, and category, and these attributes belong to three levels of structures. The first level is the video, the second level is the video frame (i.e., image), and the third level is the annotation. Therefore, we need to define a three-level nested structure to describe the information of each sample in detail.

\begin{minted}{yaml}
$dsdl-version: "0.5.0"

LocalObjectEntry:  
    $def: struct   
    $params: ["cdom"]
    $fields: 
        instance_id: InstanceID
        bbox: BBox
        category: Label[dom=$cdom]

FrameSample:
    $def: struct
    $params: ["cdom"]
    $fields:
        frame_id: UniqueID
        media_path: Image
        objects: List[etype=LocalObjectEntry[cdom=$cdom]]

VideoFrame:
    $def: struct
    $params: ["cdom"]
    $fields:
        video_name: Str
        videoframes: List[etype=FrameSample[cdom=$cdom]]
\end{minted}

The fields in the tracking template are explained below:
\begin{itemize}
    \item \texttt{\$dsdl-version}: Specifies the DSDL version of the file.
    \item \texttt{LocalObjectEntry}: A nested structure defining the description of a bounding box, containing:
        \begin{itemize}
            \item \texttt{\$def}: Indicates a structure type.
            \item \texttt{\$params}: Defines parameters, here the class domain.
            \item \texttt{\$fields}: Attributes of the structure, including:
                \begin{itemize}
                    \item \texttt{instance\_id}: The unique ID for the target.
                    \item \texttt{bbox}: Position of the bounding box.
                    \item \texttt{category}: Category of the bounding box.
                \end{itemize}
        \end{itemize}
    \item \texttt{FrameSample}: Defines the structure of an object tracking sample, containing:
        \begin{itemize}
            \item \texttt{\$def}: Indicates a structure type.
            \item \texttt{\$params}: Defines parameters, here the class domain.
            \item \texttt{\$fields}: Attributes of the structure, including:
                \begin{itemize}
                    \item \texttt{frame\_id}: The unique ID for the frame.
                    \item \texttt{media\_path}: Media file path.
                    \item \texttt{objects}: Annotation information, a list of LocalObjectEntry.
                \end{itemize}
        \end{itemize}
    \item \texttt{VideoFrame}: Defines the structure of an video sample, containing:
        \begin{itemize}
            \item \texttt{\$def}: Indicates a structure type.
            \item \texttt{\$params}: Defines parameters, here the class domain.
            \item \texttt{\$fields}: Attributes of the structure, including:
                \begin{itemize}
                    \item \texttt{video\_name}: The name of the video.
                    \item \texttt{videoframes}: Annotation information, a list of FrameSample.
                \end{itemize}
        \end{itemize}
\end{itemize}

\subsubsection{Complete Example}
Using the TracingNet dataset as an example, we present the specific content of an object tracking dataset DSDL description file.

\vspace{5pt}
\noindent\textbf{Task YAML File Definition}
\begin{minted}{yaml}
$dsdl-version: "0.5.3"

LocalObjectEntry:  
    $def: struct   
    $params: ["cdom"]
    $fields: 
        instance_id: InstanceID
        bbox: BBox
        category: Label[dom=$cdom]

FrameSample:
    $def: struct
    $params: ["cdom"]
    $fields:
        frame_id: UniqueID
        media_path: Image
        objects: List[etype=LocalObjectEntry[cdom=$cdom]]

VideoFrame:
    $def: struct
    $params: ["cdom"]
    $fields:
        video_name: Str
        _folder: Str
        videoframes: List[etype=FrameSample[cdom=$cdom]]    
\end{minted}

The fields in the task yaml are explained below:
\begin{itemize}
    \item \texttt{\$dsdl-version}: Specifies the DSDL version of the file.
    \item \texttt{LocalObjectEntry}: A nested structure defining the description of a bounding box, containing:
        \begin{itemize}
            \item \texttt{\$def}: Indicates a structure type.
            \item \texttt{\$params}: Defines parameters, here the class domain.
            \item \texttt{\$fields}: Attributes of the structure, including:
                \begin{itemize}
                    \item \texttt{instance\_id}: The unique ID for the target.
                    \item \texttt{bbox}: Position of the bounding box.
                    \item \texttt{category}: Category of the bounding box.
                \end{itemize}
        \end{itemize}
    \item \texttt{FrameSample}: Defines the structure of an object tracking sample, containing:
        \begin{itemize}
            \item \texttt{\$def}: Indicates a structure type.
            \item \texttt{\$params}: Defines parameters, here the class domain.
            \item \texttt{\$fields}: Attributes of the structure, including:
                \begin{itemize}
                    \item \texttt{frame\_id}: The unique ID for the frame.
                    \item \texttt{media\_path}: Media file path.
                    \item \texttt{objects}: Annotation information, a list of LocalObjectEntry.
                \end{itemize}
        \end{itemize}
    \item \texttt{VideoFrame}: Defines the structure of an video sample, containing:
        \begin{itemize}
            \item \texttt{\$def}: Indicates a structure type.
            \item \texttt{\$params}: Defines parameters, here the class domain.
            \item \texttt{\$fields}: Attributes of the structure, including:
                \begin{itemize}
                    \item \texttt{video\_name}: The name of the video.
                    \item \texttt{\_folder}: Identify which folder the video comes from, as there are a total of 12 folders in the training set of TrackingNet. This field is not present in the template and is an additional field.
                    \item \texttt{videoframes}: Annotation information, a list of FrameSample.
                \end{itemize}
        \end{itemize}
\end{itemize}

\noindent\textbf{DSDL Syntax Describing Category Information}

\begin{minted}{yaml}
$dsdl-version: 0.5.0
TrackingNetClassDom:
  $def: class_domain
  classes:
      - object
\end{minted}

The file above defines TrackingNetClassDom, which includes the following fields:
\begin{itemize}
    \item \texttt{\$def}: Describes the DSDL type of TrackingNetClassDom, here a class\_domain.
    \item \texttt{classes}: Lists the categories within this class domain. Since TrackingNet does not have class annotations, here we use the default class "object" as a single class information.
\end{itemize}

\noindent\textbf{Dataset YAML File Definition}

\begin{minted}{yaml}
$dsdl-version: "0.5.0"

$import:
    - ../defs/object-tracking
    - ../defs/class-dom

meta:
  dataset_name: "TrackingNet"
  creator: "King Abdullah University of Science and Technology"
  home-page: "https://tracking-net.org/"  
  opendatalab-page: "https://opendatalab.com/TrackingNet"
  sub_dataset_name: "train"
  task_name: "single-object tracking" 

data:
    sample-type: VideoFrame[cdom=TrackingNetClassDom]
    sample-path: train_samples.json
\end{minted}

In the description file above, the version of DSDL is first defined, followed by the import of two template files, including the task template and the class domain template. The meta and data fields are then used to describe the dataset, with specific field explanations as follows:
\begin{itemize}
    \item \texttt{\$dsdl-version}: DSDL version information.
    \item \texttt{\$import}: Template import information, importing the task yaml and TrackingNet's class domain.
    \item \texttt{meta}: Displays some meta-information about the dataset, such as the dataset name, creator, etc. Users can add other notes if desired.
    \item \texttt{data}: Contains the sample information saved according to the previously defined structure, specifically:
        \begin{itemize}
            \item \texttt{sample-type}: Type definition of the data, here using the VideoFrame class imported from the task yaml, specifying the class domain as TrackingNetClassDom.
            \item \texttt{sample-path}: Path where samples are stored. If it is an actual path, the content of samples is read from that file.
        \end{itemize}
\end{itemize}

The \textit{train\_samples.json} should look like that:

\begin{minted}{json}
{"samples": [
    {
        "video_name": "0-6LB4FqxoE_0",
        "_folder": "TRAIN_0",
        "videoframes": [
            {
                "frame_id": "0",
                "media_path": "TRAIN_0/frames/0-6LB4FqxoE_0/0.jpg",
                "objects": [
                    {
                        "instance_id": 000000000001,
                        "bbox": [120.24, 0.32, 359.76, 596.04], 
                        "category": 1  
                    }, 
                    ...
                 ]
            },
            ...
        ]
     },
     ...  
]}    
\end{minted}

\subsection{Rotated Object Detection}

\subsubsection{Task Research}

\noindent\textbf{Task Definition}
The task of rotated object detection involves detecting the position of objects in images using rotated rectangles, quadrilaterals, or even arbitrary shapes, and identifying their categories. The schematic diagram is shown in~\Cref{fig: rotate}:

\begin{figure}[H]
    \centering
    \includegraphics[width=0.65\textwidth]{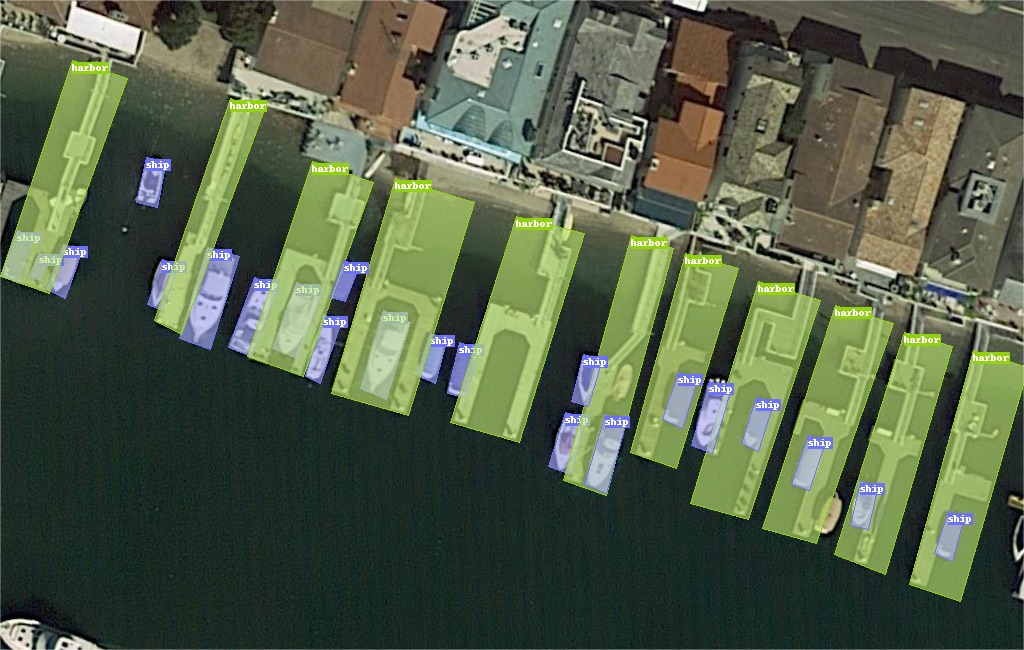}
    \caption{Example of an rotated object detection task}
    \label{fig: rotate}
\end{figure}

\noindent \textbf{Evaluation Metrics.}
The evaluation metrics for rotated object detection are the same as those for object detection, with the most commonly used metrics being mAP (mean Average Precision) and AP (Average Precision). For more detailed information, please refer to the evaluation metrics for object detection (~\Cref{sec:detection}).

\noindent\textbf{Mainstream Dataset Research}
We conducted research on five rotated object detection datasets, analyzed and summarized the relevant dataset description files (mainly annotation fields), and displayed annotation fields with the same meaning with a unified naming convention. The summary information is shown in~\Cref{table:rotate1}:

\begin{table}[H]
    \centering
    \resizebox{0.8\linewidth}{!}{ 
    \begin{tabular}{c|ccc|cccccc}
    \toprule
    \textbf{Dataset} & \multicolumn{3}{c|}{\textbf{Shared Fields}} & \multicolumn{6}{c}{\textbf{Unique Fields}} \\
    \midrule
    & \textbf{image\_id} & \textbf{label\_id} & \textbf{rbbox} & \textbf{bbox} & \textbf{istruncated} & \textbf{isdifficult} & \textbf{pose} & \textbf{theta} & \textbf{head} \\
    \midrule
    DOTAv2.0~\cite{ding2021object} & Y & Y & Y & Y & & Y & & & \\
    HRSC2016~\cite{liu2017high} & Y & Y & Y & Y & Y & Y & & & Y \\
    UCAS\_AOD~\cite{zhu2015orientation} & Y & Y & Y & Y & & & & Y & \\
    SZTAKI-INRIA & Y & Y & Y & & & & & & \\
    SSDD+~\cite{zhang2021sar} & Y & Y & Y & & Y & Y & Y & & \\
    \bottomrule
    \end{tabular}
    }
    \caption{Summary of shared and unique fields in rotated object detection datasets}
\label{table:rotate1}
\end{table}

The meanings of shared and unique fields are further summarized in~\Cref{table:rotate2}:
\begin{table}[H]
    \centering
    \begin{tabular}{@{}clp{10cm}@{}} 
    \toprule
    \textbf{Field Type} & \textbf{Field Name} & \textbf{Meaning} \\
    \midrule
    \multirow{3}{*}{Shared Fields} & image\_id & Identifies a unique image, such as by name or path \\
                                   & label\_id & Category of a single object \\
                                   & rbbox     & Rotated bounding box for a single object, e.g., [x1, y1, x2, y2, x3, y3, x4, y4] \\
    \midrule
    \multirow{9}{*}{Unique Fields} & bbox       & Bounding box for a single object, e.g., [xmin, ymin, xmax, ymax] \\
                                   & istruncated & Whether the object is truncated, i.e., partially outside the image \\
                                   & isdifficult & Whether the object is difficult to detect \\
                                   & theta  & The angle between the vector from the tail of the object to the head and the positive direction of the x-axis \\
                                   & head & The coordinates of the object's head \\
                                   & pose & The camera angle, with values being Unspecified, Frontal, Rear, Left, Right \\
    \bottomrule
    \end{tabular}
    \caption{Detailed description of fields in rotated object detection datasets}
\label{table:rotate2}
\end{table}

\subsubsection{Template Display}
Based on the aforementioned research, a single image in the rotated object detection task corresponds to an indefinite number of targets, each target is localized by a bounding box called RotatedBBox, and each RotatedBBox also provides a semantic label. Therefore, we define the rotated object detection template as follows:
\begin{minted}{yaml}
$dsdl-version: "0.5.0"

LocalObjectEntry:
    $def: struct
    $params: ['cdom']
    $fields:
        rbbox: RotatedBBox[mode="xyxy"]
        label: Label[dom=$cdom]

OrientedObjectDetectionSample:
    $def: struct
    $params: ['cdom']
    $fields:
        image: Image
        objects: List[LocalObjectEntry[cdom=$cdom]]    
\end{minted}

The fields in the template are explained below:
\begin{itemize}
    \item \texttt{\$dsdl-version}: Specifies the DSDL version of the file.
    \item \texttt{LocalObjectEntry}: A nested structure defining the description of a bounding box, containing:
        \begin{itemize}
            \item \texttt{\$def}: Indicates a structure type.
            \item \texttt{\$params}: Defines parameters, here the class domain.
            \item \texttt{\$fields}: Attributes of the structure, including:
                \begin{itemize}
                    \item \texttt{rbbox}: The position of the bounding box (bbox) is currently available in two modes:
                        \begin{itemize}
                            \item \textit{RotatedBBox[mode="xyxy"]}: Represents the annotation box as a quadrilateral, with specific annotation value examples: $[x1,y1,x2,y2,x3,y3,x4,y4]$, where $xi$, $yi$ represent the coordinates of the four vertices of the quadrilateral.
                            \item \textit{RotatedBBox[mode="xywht", measure="degree"]}: Represents the annotation box as a rotated rectangle, with specific annotation value examples: $[x, y, w, h, t]$, where $x$, $y$ are the coordinates of the rectangle's center point, $w$, $h$ are the width and height of the rectangle, and $t$ is the rotation angle of the rectangle. When measure="degree", $t$ represents the angle in degrees, with a range of $(-180, 180)$; when \textit{measure="radian"}, $t$ represents the angle in radians, with a range of $(-pi, pi)$. The default value for measure is radians. That is, \textit{RotatedBBox[mode="xywht"]} implies that $t$ should only be filled in with radian values.
                        \end{itemize}
                    \item \texttt{label}: Category of the bounding box.
                \end{itemize}
        \end{itemize}
    \item \texttt{OrientedObjectDetectionSample}: Defines the structure of an rotated object detection sample, containing:
        \begin{itemize}
            \item \texttt{\$def}: Indicates a structure type.
            \item \texttt{\$params}: Defines parameters, here the class domain.
            \item \texttt{\$fields}: Attributes of the structure, including:
                \begin{itemize}
                    \item \texttt{image}: Path of the image.
                    \item \texttt{objects}: Annotation information, a list of LocalObjectEntry.
                \end{itemize}
        \end{itemize}
\end{itemize}

\subsubsection{Complete Example}
Using the DOTAv2.0 dataset as an example, we present the specific content of a rotated object detection dataset DSDL description file.

\vspace{5pt}
\noindent\textbf{Task YAML File Definition}
\begin{minted}{yaml}
$dsdl-version: "0.5.2"

LocalObjectEntry:
  $def: struct
  $params: ['cdom']
  $fields:
    rbbox: RotatedBBox[mode="xyxy"]
    label: Label[dom=$cdom]
    isdifficult: Bool
    bbox: BBox
  $optional: ['bbox']

OrientedObjectDetectionSample:
  $def: struct
  $params: ['cdom']
  $fields:
    image: Image
    imageshape: ImageShape
    objects: List[etype=LocalObjectEntry[cdom=$cdom]]
    acquisition_dates: Str
    imagesource: Str
    gsd: Num
  $optional: ['objects','acquisition_dates', 'imagesource', 'gsd']
\end{minted}
In this template, the \$optional field is used, which contains a list of values that can only be derived from the \$fields section. Fields listed in this list may not be present in every sample.

\vspace{5pt}
\noindent\textbf{DSDL Syntax Describing Category Information}

\begin{minted}{yaml}
$dsdl-version: "0.5.0"

DOTAV2ClassDom:
    $def: class_domain
    classes:
        - large_vehicle
        - small_vehicle
        - ship
        - ground_track_field
        - soccer_ball_field
        - tennis_court
        - swimming_pool
        - harbor
        - baseball_diamond
        - plane
        - storage_tank
        - roundabout
        - basketball_court
        - bridge
        - helicopter
        - container_crane
        - airport
        - helipad


ExampleClassDomDescr:
    $def: struct
    $params: ['cdom']
    $fields:
        dsdl_name: Label[dom=$cdom]
        original_name: Str

ClassMapInfo:
    $def: struct
    $params: ['cdom']
    $fields:
        class_info: List[ExampleClassDomDescr[cdom=$cdom]]
\end{minted}

The file above defines DOTAV2ClassDom, which includes the following fields:
\begin{itemize}
    \item \texttt{\$def}: Describes the DSDL type of DOTAV2ClassDom, here a class\_domain.
    \item \texttt{classes}: Lists the categories within this class domain, in the DOTAv2.0 dataset, these include large\_vehicle, ship, etc.
\end{itemize}

From the class\_domain, you can obtain all category information for the current dataset. Users familiar with the DOTA dataset will notice that the current category names differ from the original DOTAv2.0 dataset. This is because the category naming in the DOTAv2.0 dataset contains special characters, which we have converted to standardized ones. We store the mapping relationship from the original naming to the DSDL naming in ClassMapInfo, which users can retrieve as needed.

\vspace{5pt}
\noindent\textbf{Dataset YAML File Definition}
\begin{minted}{yaml}
$dsdl-version: "0.5.0"

$import:
  - ../defs/class-dom
  - ../defs/rotated-detection

meta:
  dataset_name: "DOTAv2.0"
  subset-name: "train"
  creator: "Wuhan University·Cornell University·Huazhong University of Science and Technology"
  dataset-version: "2.0"
  home-page: "https://captain-whu.github.io/DOTA/dataset.html"
  opendatalab-page: "https://opendatalab.com/DOTA_V2.0"
  task_type: "Rotated Object Detection"

data:
    global-info-type: ClassMapInfo[cdom=DOTAV2ClassDom]
    global-info-path: ../defs/global-info.json 
    sample-type: OrientedObjectDetectionSample[cdom=DOTAV2ClassDom]
    sample-path: samples.json    
\end{minted}

In the description file above, the version of DSDL is first defined, followed by the import of two template files, including the task template and the class domain template. The meta and data fields are then used to describe the dataset, with specific field explanations as follows:
\begin{itemize}
    \item \texttt{\$dsdl-version}: DSDL version information.
    \item \texttt{\$import}: Template import information, importing the task yaml and DOTAv2.0's class domain.
    \item \texttt{meta}: Displays some meta-information about the dataset, such as the dataset name, creator, etc. Users can add other notes if desired.
    \item \texttt{data}: Contains the sample information saved according to the previously defined structure, specifically:
        \begin{itemize}
            \item \texttt{global-info-type}: The global information type definition for the dataset uses the ClassMapInfo class imported from class-dom.yaml, and specifies that the cdom used is DOTAV2ClassDom.
            \item \texttt{global-info-path}: The storage path for the global information file \textit{global-info.json}.
            \item \texttt{sample-type}: Type definition of the data, here using the OrientedObjectDetectionSample class imported from the task yaml, specifying the class domain as DOTAV2ClassDom.
            \item \texttt{sample-path}: Path where samples are stored. If it is an actual path, the content of samples is read from that file.
        \end{itemize}
\end{itemize}

Here is the example of \textit{global-info.json}:
\begin{minted}{json}
{
  "global-info": {
    "class_info": [
      {
          "dsdl_name": "roundabout",
          "original_name": "roundabout"
      },
      {
          "dsdl_name": "large_vehicle",
          "original_name": "large-vehicle"
      },
      {
          "dsdl_name": "small_vehicle",
          "original_name": "small-vehicle"
      },
      ...
    ]
  }
}    
\end{minted}

The \textit{train\_samples.json} should look like that:
\begin{minted}{json}
{
  "samples": [
    {     
      "image": "train/images/P0000.png",
      "imageshape": [
        5502,   
        3875  
      ],    
      "objects": [
        {     
          "rbbox": [
            2244.0, 
            1791.0, 
            2254.0, 
            1795.0, 
            2245.0, 
            1813.0, 
            2238.0, 
            1809.0  
          ],    
          "label": "small_vehicle",
          "isdifficult": true,
          "bbox": [
            2238.0, 
            1791.0, 
            16.0,   
            22.0  
          ]     
        },
        ...
      ],
    "acquisition_dates": "2016-05-04",
    "imagesource": "GoogleEarth",
    "gsd": 0.146343590398
    }
    ...
    ]
}    
\end{minted}

\subsection{OCR}

\subsubsection{Task Research}

\begin{figure}[H]
    \centering
    \includegraphics[width=0.65\textwidth]{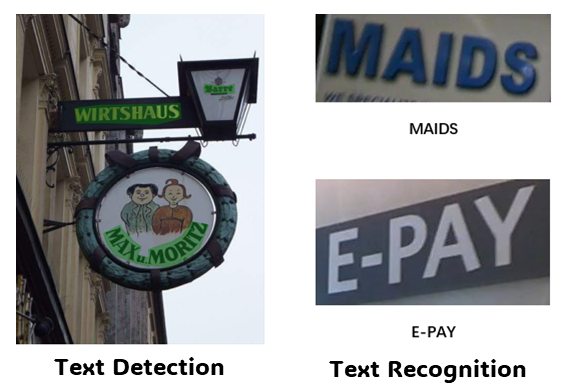}
    \caption{Example of an OCR task.}
    \label{fig: ocr}
\end{figure}

\noindent\textbf{Task Definition}
Optical Character Recognition (OCR) refers to a technology that identifies, extracts, and recognizes text from images. It detects patterns of dark and light to determine their shapes, and then translates these shapes into computer-readable text using character recognition methods. The following is illustrated in~\Cref{fig: ocr}.

OCR tasks can be divided into several parts according to the algorithm:
\begin{itemize}
    \item Text Detection/Segmentation, which refers to accurately locating the position of text in scene images, with annotation forms such as polygon, bounding box (bbox), or segment maps.
    \item Text Recognition, which refers to obtaining the text content from pure text images or the detection frames mentioned above, with annotation forms such as text.
    \item Key Information Extraction, which extracts desired categories of information from special documents such as receipts and identity cards.
    \item Other tasks include table recognition, relationship extraction, layout analysis, etc.
Based on the input images, they can be categorized into street view images, handwritten images, documents, web images, etc.
\end{itemize}

\noindent \textbf{Evaluation Metrics}
\begin{itemize}
    \item Text detection: Assess the detection based on the Intersection over Union (IOU). If the IOU is greater than a certain threshold, it is judged as accurate detection. Here, the detection and annotation boxes differ from general object detection boxes, and some are represented by polygons.
    \[
    Precision = \frac{\textit{Number of correctly detected sentences}}{\textit{Number of sentences detected by model}}
    \]
    \[
    Recall = \frac{\textit{Number of correctly detected sentences}}{\textit{Number of annotated sentences}}
    \]
    \[
    F_1 = 2 \times \frac{\text{Precision} \times \text{Recall}}{\text{Precision} + \text{Recall}}
    \]
    \item Text Recognition Accuracy: Only when the entire line of text is correctly identified does it count as accurate recognition.
    \[
    Accuracy = \frac{\textit{Number of correctly recognized sentences}}{\textit{Number of annotated sentences}}
    \]
    \item End-to-End OCR: Including text detection and text recognition.
    \[
    Precision = \frac{\textit{Number of correctly detected and recognized sentences}}{\textit{Number of sentences detected by model}}
    \]
    \[
    Recall = \frac{\textit{Number of correctly detected and recognized sentences}}{\textit{Number of annotated sentences}}
    \]
\end{itemize}

\noindent\textbf{Mainstream Dataset Research}
Considering that some datasets contain different OCR subtasks, these subtasks have been split here, and each subtask only considers the annotations related to it. To make the template more universal and extensible, we focus on the commonalities and characteristics between datasets. In addition, during the research process, we may encounter fields with different names but the same or similar meanings; we also consider these fields as the same and name them uniformly. For example, the image\_id field generally represents the path or ID of the image and is the unique identifier of the image.

1. Text Detection

We have surveyed ICDAR2013~\cite{karatzas2013icdar}, ICDAR2015~\cite{karatzas2015icdar}, SVT~\cite{wang2011end}, Total-Text~\cite{ch2017total}, MSRA-TD50~\cite{yao2012detecting}, and CUTE80~\cite{risnumawan2014robust} datasets (~\Cref{table:ocr1}).

\begin{table}[H]
    \centering
    \small
    \resizebox{0.6\linewidth}{!}{ 
    \begin{tabular}{c|ccc|cc}
    \toprule
    \textbf{Dataset} & \multicolumn{3}{c|}{\textbf{Shared Fields}} & \multicolumn{2}{c}{\textbf{Unique Fields}} \\
    \midrule
    & \textbf{image\_id} & \textbf{bbox} & \textbf{polygon} & \textbf{orientation} & \textbf{isdifficult} \\
    \midrule
    ICDAR2013~\cite{karatzas2013icdar} & Y & Y & Y & & \\
    ICDAR2015~\cite{karatzas2015icdar} & Y & & Y & & \\
    SVT~\cite{wang2011end} & Y & Y & & & \\
    Total-Text~\cite{ch2017total} & Y & & Y & Y & \\
    MSRA-TD50~\cite{yao2012detecting} & Y & Y & & Y & Y \\
    CUTE80~\cite{risnumawan2014robust} & Y & & Y & & \\
    \bottomrule
    \end{tabular}
    }
    \caption{Summary of shared and unique fields in text detection datasets}
\label{table:ocr1}
\end{table}

The meanings of shared and unique fields are further summarized in~\Cref{table:ocr2}:
\begin{table}[H]
    \centering
    \begin{tabular}{@{}clp{10cm}@{}} 
    \toprule
    \textbf{Field Type} & \textbf{Field Name} & \textbf{Meaning} \\
    \midrule
    \multirow{3}{*}{Shared Fields} & image\_id & Identifies a unique image, such as by name or path \\
                                   & bbox      & Bounding box for a single object, e.g., [xmin, ymin, xmax, ymax] \\
                                   & polygon & Polygonal bounding box for locating a single target, with a variable number of vertices. \\
    \midrule
    \multirow{2}{*}{Unique Fields} & orientation     & The rotation angle of the target bounding box, or the rotation category of the target bounding box: curved, horizontal, etc. \\
                                   & isdifficult & Whether the object is difficult to detect \\
    \bottomrule
    \end{tabular}
    \caption{Detailed description of fields in text detection datasets}
\label{table:ocr2}
\end{table}

2. Text Segmentation

There are few datasets that support text segmentation. Here, we only investigated Total Text~\cite{ch2017total}, and the results shows in~\Cref{table:ocr3}:

\begin{table}[H]
    \centering
    \small
    \resizebox{0.4\linewidth}{!}{ 
    \begin{tabular}{c|cc}
    \toprule
    \textbf{Dataset} & \multicolumn{2}{c}{\textbf{Shared Fields}} \\
    \midrule
    & \textbf{image\_id} & \textbf{segmentation\_map} \\
    \midrule
    Total-Text~\cite{ch2017total} & Y & Y \\
    \bottomrule
    \end{tabular}
    }
    \caption{Summary of shared and unique fields in text segmentation datasets}
\label{table:ocr3}
\end{table}

The meanings of shared fields are further summarized in~\Cref{table:ocr4}:
\begin{table}[H]
    \centering
    \begin{tabular}{@{}clp{10cm}@{}} 
    \toprule
    \textbf{Field Type} & \textbf{Field Name} & \textbf{Meaning} \\
    \midrule
    \multirow{2}{*}{Shared Fields} & image\_id & Identifies a unique image, such as by name or path \\
                                   & segmentation\_map      & A segmentation graph, which can be a word segmentation or a character segmentation, usually a binary graph. \\
    \bottomrule
    \end{tabular}
    \caption{Detailed description of fields in text segmentation datasets}
\label{table:ocr4}
\end{table}

3. Text Recognition

We conducted research on ICDAR2013~\cite{karatzas2013icdar}, ICDAR2015~\cite{karatzas2015icdar}, SVT~\cite{wang2011end}, Total-Text~\cite{ch2017total},IIIT-5K~\cite{mishra2012scene}, give the result as shows in~\Cref{table:ocr5}:

\begin{table}[H]
    \centering
    \small
    \resizebox{0.45\linewidth}{!}{ 
    \begin{tabular}{c|cc|c}
    \toprule
    \textbf{Dataset} & \multicolumn{2}{c|}{\textbf{Shared Fields}} & \multicolumn{1}{c}{\textbf{Unique Fields}} \\
    \midrule
    & \textbf{image\_id} & \textbf{text} & \textbf{lexicon} \\
    \midrule
    ICDAR2013~\cite{karatzas2013icdar} & Y & Y & Y \\
    ICDAR2015~\cite{karatzas2015icdar} & Y & Y & Y \\
    SVT~\cite{wang2011end} & Y & Y & Y \\
    Total-Text~\cite{ch2017total} & Y & Y & \\
    IIIT-5K~\cite{mishra2012scene} & Y & Y & Y \\
    \bottomrule
    \end{tabular}
    }
    \caption{Summary of shared and unique fields in text recognition datasets}
\label{table:ocr5}
\end{table}

The meanings of shared and unique fields are further summarized in~\Cref{table:ocr6}:
\begin{table}[H]
    \centering
    \begin{tabular}{@{}clp{10cm}@{}} 
    \toprule
    \textbf{Field Type} & \textbf{Field Name} & \textbf{Meaning} \\
    \midrule
    \multirow{2}{*}{Shared Fields} & image\_id & Identifies a unique image, such as by name or path \\
                                   & text      & Text content in the image \\
    \midrule
    \multirow{1}{*}{Unique Fields} & lexicon     & Vocabulary list, which can be one vocabulary list for one single image, or one for all images in the dataset. \\
    \bottomrule
    \end{tabular}
    \caption{Detailed description of fields in text recognition datasets}
\label{table:ocr6}
\end{table}

\subsubsection{Template Display}

We have formulated templates for different OCR tasks.

\noindent\textbf{Text Recognition}
\begin{minted}{yaml}
$dsdl-version: "0.5.0"

OCRSample:
    $def: struct
    $fields:
        image: Image
        instances: List[Bbox]     
\end{minted}

The fields in the text detection template are explained below:
\begin{itemize}
    \item \texttt{\$dsdl-version}: Specifies the DSDL version of the file.
    \item \texttt{OCRSample}: Defines the structure of an text detection sample, containing:
        \begin{itemize}
            \item \texttt{\$def}: Indicates a structure type.
            \item \texttt{\$params}: Defines parameters, here the class domain.
            \item \texttt{\$fields}: Attributes of the structure, including:
                \begin{itemize}
                    \item \texttt{image}: Path of the image.
                    \item \texttt{instances}: Annotation information, a list of bounding box or polygon.
                \end{itemize}
        \end{itemize}
\end{itemize}

\noindent\textbf{Text Segmentation}
\begin{minted}{yaml}
$dsdl-version: "0.5.0"

SegClassDom:
    $def: class_domain
    classes:
        - text

OCRSample:
    $def: struct
    $params: ["cdom"]
    $fields:
        image: Image
        word_segmap: LabelMap[dom=$cdom]
        chr_segmap: LabelMap[dom=$cdom]    
\end{minted}

The fields in the text segmentation template are explained below:
\begin{itemize}
    \item \texttt{\$dsdl-version}: Specifies the DSDL version of the file.
    
    \item \texttt{SegClassDom}:
        \begin{itemize}
            \item \texttt{\$def}: Describes the DSDL type of SegClassDom, here a class\_domain.
            \item \texttt{classes}: Lists the categories within this class domain, in the text segmentation tack, only contain one class, i.e., text.
        \end{itemize}
    \item \texttt{OCRSample}: Defines the structure of an text segmentation sample, containing:
        \begin{itemize}
            \item \texttt{\$def}: Indicates a structure type.
            \item \texttt{\$params}: Defines parameters, here the class domain.
            \item \texttt{\$fields}: Attributes of the structure, including:
                \begin{itemize}
                    \item \texttt{image}: Path of the image.
                    \item \texttt{word\_segmap}: Word-level segmentation map.
                    \item \texttt{chr\_segmap}: Character-level segmentation map.
                \end{itemize}
        \end{itemize}
\end{itemize}

\noindent\textbf{Text Recognition}
The text recognition template is used exclusively for OCR tasks and defines a new field called "Text".

\begin{minted}{yaml}
$dsdl-version: "0.5.0"

OCRSample:
    $def: struct
    $fields:
        word_image: Image
        text: Text    
\end{minted}

The fields in the template are explained below:
\begin{itemize}
    \item \texttt{\$dsdl-version}: Specifies the DSDL version of the file.
    \item \texttt{OCRSample}: Defines the structure of an text recoginition sample, containing:
        \begin{itemize}
            \item \texttt{\$def}: Indicates a structure type.
            \item \texttt{\$fields}: Attributes of the structure, including:
                \begin{itemize}
                    \item \texttt{word\_image}: Path of the image.
                    \item \texttt{text}: Text content in the image.
                \end{itemize}
        \end{itemize}
\end{itemize}

\noindent\textbf{End-to-End OCR}
Here, the end-to-end task takes text detection followed by text recognition as an example:
\begin{minted}{yaml}
$dsdl-version: "0.5.0"

LocalInstanceEntry:
    $def: struct
    $fields:
        location: Polygon/Bbox 
        text: Text

OCRSample:
    $def: struct
    $fields:
        image: Image
        instances: List[LocalInstanceEntry]
\end{minted}

The fields in the template are explained below:
\begin{itemize}
    \item \texttt{\$dsdl-version}: Specifies the DSDL version of the file.
    \item \texttt{LocalInstanceEntry}: A nested structure defining the description of a bounding box, containing:
        \begin{itemize}
            \item \texttt{\$def}: Indicates a structure type.
            \item \texttt{\$fields}: Attributes of the structure, including:
                \begin{itemize}
                    \item \texttt{location}: Position of the bounding box or polygon.
                    \item \texttt{text}: Text content of the bounding box.
                \end{itemize}
        \end{itemize}
    \item \texttt{OCRSample}: Defines the structure of an text recoginition sample, containing:
        \begin{itemize}
            \item \texttt{\$def}: Indicates a structure type.
            \item \texttt{\$fields}: Attributes of the structure, including:
                \begin{itemize}
                    \item \texttt{image}: Path of the image.
                    \item \texttt{instances}: Annotation information, a list of LocalInstanceEntry.
                \end{itemize}
        \end{itemize}
\end{itemize}

\subsubsection{Complete Example}
In this section, let's discuss how to use the task templates by importing it into the dataset yaml. Taking the SynthText end-to-end template as an example.

\vspace{5pt}
\noindent\textbf{Task YAML File Definition}
\begin{minted}{yaml}
$dsdl-version: "0.5.0"

LocalCharEntry:                       
    $def: struct 
    $fields:
        char_polygon: Polygon
        char_text: Text                

LocalInstanceEntry:                         
    $def: struct 
    $fields:        
        polygon: Polygon
        text: Text                   
        charlist: List[LocalCharEntry]

OCRSample:
    $def: struct
    $fields:
        image: Image
        instances: List[LocalInstanceEntry]   
\end{minted}

The fields in the task yaml file are explained below:
\begin{itemize}
    \item \texttt{\$dsdl-version}: Specifies the DSDL version of the file.
    \item \texttt{LocalCharEntry}: A nested structure defining the description of the annotation of every character, containing:
        \begin{itemize}
            \item \texttt{\$def}: Indicates a structure type.
            \item \texttt{\$fields}: Attributes of the structure, including:
                \begin{itemize}
                    \item \texttt{char\_polygon}: Position of the polygon of every character.
                    \item \texttt{char\_text}: Text content of the polygon.
                \end{itemize}
        \end{itemize}
    \item \texttt{LocalInstanceEntry}: A nested structure defining the description of the annotation of every word or sentance, containing:
        \begin{itemize}
            \item \texttt{\$def}: Indicates a structure type.
            \item \texttt{\$fields}: Attributes of the structure, including:
                \begin{itemize}
                    \item \texttt{polygon}: Position of the polygon of every word or sentance.
                    \item \texttt{text}: Text content of the polygon.
                    \item \texttt{charlist}: Annotation information, a list of LocalCharEntry.
                \end{itemize}
        \end{itemize}
    \item \texttt{OCRSample}: Defines the structure of a sample, containing:
        \begin{itemize}
            \item \texttt{\$def}: Indicates a structure type.
            \item \texttt{\$fields}: Attributes of the structure, including:
                \begin{itemize}
                    \item \texttt{image}: Path of the image.
                    \item \texttt{instances}: Annotation information, a list of LocalInstanceEntry.
                \end{itemize}
        \end{itemize}
\end{itemize}

\vspace{5pt}
\noindent\textbf{Dataset YAML File Definition}
\begin{minted}{yaml}
$dsdl-version: "0.5.0"

$import:
    - ../defs/OCR-SynthText

meta:
    dataset_name: "SynthText"
    creator: "University of Oxford"
    home-page: "https://www.robots.ox.ac.uk/~vgg/data/scenetext/"
    opendatalab-page: "https://opendatalab.com/SynthText"
    sub-name: "train"
    task-type: "Optical Character Recognition"

data:
    sample-type: OCRSample
    sample-path: train_samples.json    
\end{minted}

In the description file above, the version of DSDL is first defined, followed by the import of the task yaml file pre-defined. The meta and data fields are then used to describe the dataset, with specific field explanations as follows:
\begin{itemize}
    \item \texttt{\$dsdl-version}: DSDL version information.
    \item \texttt{\$import}: Template import information, importing the task yaml file.
    \item \texttt{meta}: Displays some meta-information about the dataset, such as the dataset name, creator, etc. Users can add other notes if desired.
    \item \texttt{data}: Contains the sample information saved according to the previously defined structure, specifically:
        \begin{itemize}
            \item \texttt{sample-type}: Type definition of the data, here using the OCRSample class imported from the task yaml file.
            \item \texttt{sample-path}: Path where samples are stored. If it is an actual path, the content of samples is read from that file.
        \end{itemize}
\end{itemize}

The \textit{train\_samples.json} should look like that:

\begin{minted}{json}
{"samples": [
    {
        "image": "image":"8/ballet_106_0.jpg",
        "instances": [
            {
                "polygon": [[[420,21],[512,23],[512,42],[420,40]]],
                "text": "Lines",
                "charlist":[
                	{"char_polygon":[[[423,22],[438,22],[436,40],[420,40]]],
                	 "char_text": "L"}
                	{"char_polygon":[[[440,22],[453,22],[450,40],[437,40]]],
                	 "char_text": "i"}
                	 ...
                ]
            },
            ...
        ]
    },
    ...
]}   
\end{minted}

\subsection{Image Generation}

\subsubsection{Task Research}

\begin{figure}[t]
    \centering
    \begin{subfigure}[b]{0.32\textwidth}
        \centering
        \includegraphics[width=\linewidth]{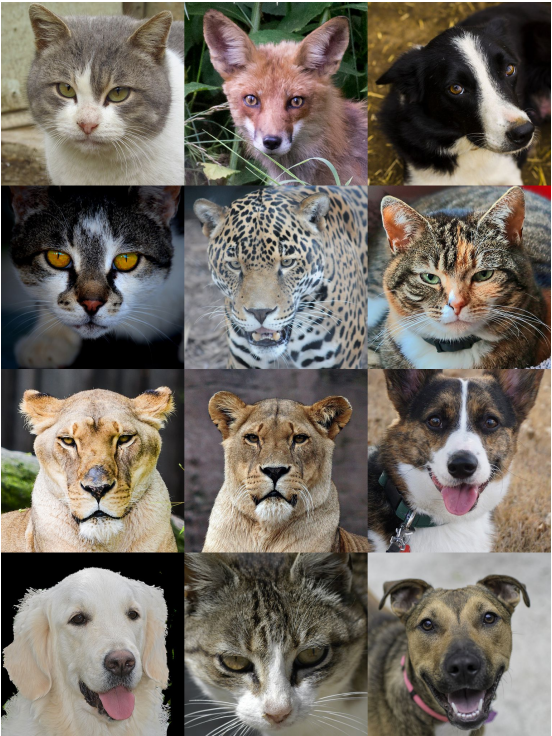}
        \caption{Real images from the AFHQv2 dataset.}
        \label{fig:gen1}
    \end{subfigure}
    \begin{subfigure}[b]{0.32\textwidth}
        \centering
        \includegraphics[width=\linewidth]{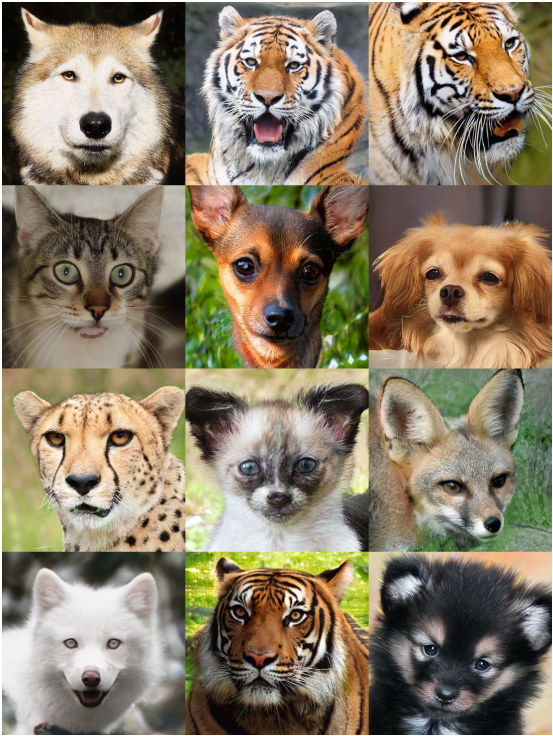}
        \caption{Images generated by StyleGANv3.}
        \label{fig:gen2}
    \end{subfigure}
    \caption{Example of image generation task.}
    \label{fig:generation}
\end{figure}

\noindent\textbf{Task Definition}
Image generation refers to the task of learning statistical patterns from an existing set of image data and creating new images that conform to the statistical patterns of the original dataset but have not appeared in it. An example shows in~\Cref{fig:generation}.

\noindent \textbf{Evaluation Metrics.}
When designing evaluation metrics for image generation models, there are generally several objectives:

1. The more realistic the generated images are, the better the evaluation score should be. This is because the main goal of most generative models is to produce more realistic images.

2. The more diverse the generated images are, the better the evaluation score should be. This primarily serves to determine whether the generative model has overfitted or experienced modal collapse.

3. For the latent variables in the generative model, it is preferable that they have some "meaning," meaning that by changing the latent variables in a certain way, the generated images should also exhibit regular and meaningful changes. The evaluation metric should be able to measure whether the latent variables in the generative model have this "meaning."

4. The changes in the evaluation score should be consistent with human perception, meaning that images with a large difference in evaluation scores should also have a large difference in visual effects perceived by humans.

Currently, there are many types of evaluation metrics for image generation models, which generally need to meet the aforementioned requirements (or some of them). Although these metrics have some limitations, they still have significant reference value when evaluating the performance of a generative model.

Below are two commonly used evaluation metrics: Inception Score (IS) and Fréchet Inception Distance (FID).
\begin{itemize}
    \item Inception Score (IS): This metric measures the quality and diversity of a generative model's output. It is based on the Inception model's predictions and calculates a score that reflects how well the generated images match the distribution of real images. A higher IS score indicates that the generated images are both realistic and diverse.
    \item Fréchet Inception Distance (FID): This metric quantifies the difference between the distributions of generated images and real images. It uses the Inception model to compute the two-sample test statistic between the two distributions. A lower FID score indicates that the generated images are closer to the real images in terms of their statistical properties, suggesting better performance of the generative model.
\end{itemize}

\noindent\textbf{Mainstream Dataset Research}
We conducted research on some image generation datasets, analyzed and summarized the relevant dataset description files (mainly annotation fields). Annotation fields with the same meaning will be displayed with a unified naming convention. The summary information is shown in~\Cref{table:gen1}. 
\begin{table}[H]
    \centering
    \resizebox{0.8\linewidth}{!}{ 
    \begin{tabular}{cc|c|ccc}
    \toprule
    \multicolumn{2}{c|}{ } & \multicolumn{1}{c|}{\textbf{Shared Fields}} & \multicolumn{3}{c}{\textbf{Unique Fields}} \\
    \midrule
    \textbf{Task Name} & \textbf{Dataset} & \textbf{image\_id} & \textbf{category\_name} & \textbf{paired\_images} & \textbf{domain} \\
    \midrule
    \multirow{5}{*}{{\makecell{Unconditional Generation}}} & LSUN~\cite{yu2015lsun} & Y & Y & & \\
    & FFHQ~\cite{karras2019style} & Y & & & \\
    & CelebA-HQ~\cite{karras2017progressive} & Y & & & \\
    & MetFaces~\cite{karras2020training} & Y & & & \\
    & AFHQ~\cite{choi2020stargan} & Y & & & Y \\
    \midrule
    \multirow{2}{*}{{\makecell{Conditional Generation}}} & ImageNet-1K~\cite{russakovsky2015imagenet} & Y & Y & & \\
    & Cifar-10~\cite{krizhevsky2009learning} & Y & Y & & \\
    \midrule
    Style Transfer(Unpaired) & Facade\_CycleGAN~\cite{zhu2017unpaired} & Y & & & \\
    \midrule
    Style Transfer(Paired) & Facade\_pix2pix & Y & & Y & Y \\
    \bottomrule
    \end{tabular}
    }
    \caption{Summary of shared and unique fields in image generation datasets}
\label{table:gen1}
\end{table}

The meanings of shared and unique fields are further summarized in~\Cref{table:gen2}:
\begin{table}[H]
    \centering
    \begin{tabular}{@{}clp{10cm}@{}} 
    \toprule
    \textbf{Field Type} & \textbf{Field Name} & \textbf{Meaning} \\
    \midrule
    \multirow{1}{*}{Shared Fields} & image\_id & Identifies a unique image, such as by name or path \\
    \midrule
    \multirow{3}{*}{Unique Fields} & category\_name     & The category name of the image. \\
                                   & paired\_images & Whether the paired image exist \\
                                   & domain & The domain of an image. It represents the stylistic characteristics of the image and can be used for tasks such as style transfer. \\
    \bottomrule
    \end{tabular}
    \caption{Detailed description of fields in image generation datasets}
\label{table:gen2}
\end{table}

\subsubsection{Template Display}

From the research results of the dataset, it can be seen that for Unconditional Generation, only the image\_id is a required field. For Conditional Generation, only the image\_id and category\_name are required fields. Therefore, there is no need to develop specific templates for these two tasks. The dataset for Conditional Generation can be described using the image classification template, while the dataset for Unconditional Generation only needs to define a struct with an Image field in the sample.

For Style Transfer tasks, different templates can be developed based on whether there are paired images (paired/unpaired). Below are two templates introduced respectively.

\vspace{5pt}
\noindent\textbf{Style Transfer(Unpaired)}

\begin{minted}{yaml}
$dsdl-version: "0.5.0"

UnpairedSample:
    $def: struct
    $params: ['cdom']
    $fields:
        image: Image
        domain: Label[dom=$cdom]   
\end{minted}

The fields in the template are explained below:
\begin{itemize}
    \item \texttt{\$dsdl-version}: Specifies the DSDL version of the file.
    \item \texttt{UnpairedSample}: Defines the structure of an unparied style transfer sample, containing:
        \begin{itemize}
            \item \texttt{\$def}: Indicates a structure type.
            \item \texttt{\$params}: Defines parameters, here the class domain.
            \item \texttt{\$fields}: Attributes of the structure, including:
                \begin{itemize}
                    \item \texttt{image}: Path of the image.
                    \item \texttt{domain}: The domain of an image.
                \end{itemize}
        \end{itemize}
\end{itemize}

\vspace{5pt}
\noindent\textbf{Style Transfer(Paired)}

\begin{minted}{yaml}
$dsdl-version: "0.5.0"

ImageMedia:
    $def: struct
    $params: ['cdom']
    $fields:
        image: Image
        domain: Label[dom=$cdom]

PairedSample:
    $def: struct
    $params: ['cdom']
    $fields:
        image_a: ImageMedia[cdom=$cdom]
        image_b: ImageMedia[cdom=$cdom]    
\end{minted}

The fields in the template are explained below:
\begin{itemize}
    \item \texttt{\$dsdl-version}: Specifies the DSDL version of the file.
    \item \texttt{ImageMedia}: Defines the structure of a labeled image, containing:
        \begin{itemize}
            \item \texttt{\$def}: Indicates a structure type.
            \item \texttt{\$params}: Defines parameters, here the class domain.
            \item \texttt{\$fields}: Attributes of the structure, including:
                \begin{itemize}
                    \item \texttt{image}: Path of the image.
                    \item \texttt{domain}: The domain of an image.
                \end{itemize}
        \end{itemize}
    \item \texttt{PairedSample}: Defines the structure of an paried style transfer sample, containing:
        \begin{itemize}
            \item \texttt{\$def}: Indicates a structure type.
            \item \texttt{\$params}: Defines parameters, here the class domain.
            \item \texttt{\$fields}: Attributes of the structure, including:
                \begin{itemize}
                    \item \texttt{image\_a / image\_b}: A paired ImageMedia sample.
                \end{itemize}
        \end{itemize}
\end{itemize}

\subsubsection{Complete Example}
Let's take the CMP Facade dataset as an example to demonstrate the specific content of the DSDL description file for style transfer dataset. 

\vspace{5pt}
\noindent\textbf{\textit{1. Style Transfer (Paired)}}

\vspace{5pt}
\noindent\textbf{Task YAML File Definition}

\begin{minted}{yaml}
$dsdl-version: "0.5.0"

ImageMedia:
    $def: struct
    $params: ['cdom']
    $fields:
        image: Image
        domain: Label[dom=$cdom]

FacadeImageSample:
    $def: struct
    $params: ['cdom']
    $fields:
        image_a: ImageMedia[cdom=$cdom]
        image_b: ImageMedia[cdom=$cdom]    
\end{minted}

The fields in the yaml are explained below:
\begin{itemize}
    \item \texttt{\$dsdl-version}: Specifies the DSDL version of the file.
    \item \texttt{ImageMedia}: Defines the structure of a labeled image, containing:
        \begin{itemize}
            \item \texttt{\$def}: Indicates a structure type.
            \item \texttt{\$params}: Defines parameters, here the class domain.
            \item \texttt{\$fields}: Attributes of the structure, including:
                \begin{itemize}
                    \item \texttt{image}: Path of the image.
                    \item \texttt{domain}: The domain of an image.
                \end{itemize}
        \end{itemize}
    \item \texttt{FacadeImageSample}: Defines the structure of an paried style transfer sample, containing:
        \begin{itemize}
            \item \texttt{\$def}: Indicates a structure type.
            \item \texttt{\$params}: Defines parameters, here the class domain.
            \item \texttt{\$fields}: Attributes of the structure, including:
                \begin{itemize}
                    \item \texttt{image\_a / image\_b}: A paired ImageMedia sample.
                \end{itemize}
        \end{itemize}
\end{itemize}

\vspace{5pt}
\noindent\textbf{DSDL Syntax Describing Category Information}

\begin{minted}{yaml}
$dsdl-version: "0.5.0"

FacadeStyleDom:
    $def: class_domain
    classes:
        - photo
        - mask
\end{minted}

The file above defines FacadeStyleDom, which includes the following fields:
\begin{itemize}
    \item \texttt{\$def}: Describes the DSDL type of FacadeStyleDom, here a class\_domain.
    \item \texttt{classes}: Lists the categories within this class domain, in the CMP Facade dataset, these include photo and mask.
\end{itemize}

\vspace{5pt}
\noindent\textbf{Dataset YAML File Definition}

\begin{minted}{yaml}
$dsdl-version: "0.5.0"

$import:
    - ../defs/image-generation-facade
    - ../defs/class-dom
meta:
    Dataset Name: "CMP Facade"
    HomePage: "https://cmp.felk.cvut.cz/~tylecr1/facade/"
    Subset Name: "base"
    Modality: "Images"
    Task: "Image Generation"
data:
    sample-type: FacadeImageSample[sdom=FacadeStyleDom]
    sample-path: train_samples.json
\end{minted}

In the description file above, the version of DSDL is first defined, followed by the import of two template files, including the task yaml and the class domain. The meta and data fields are then used to describe the dataset, with specific field explanations as follows:
\begin{itemize}
    \item \texttt{\$dsdl-version}: DSDL version information.
    \item \texttt{\$import}: Template import information, importing the CMP Facade's task yaml file and  class domain.
    \item \texttt{meta}: Displays some meta-information about the dataset, such as the dataset name, creator, etc. Users can add other notes if desired.
    \item \texttt{data}: Contains the sample information saved according to the previously defined structure, specifically:
        \begin{itemize}
            \item \texttt{sample-type}: Type definition of the data, here using the FacadeImageSample class imported from the task yaml, specifying the class domain as FacadeStyleDom.
            \item \texttt{sample-path}: Path where samples are stored. If it is an actual path, the content of samples is read from that file.
        \end{itemize}
\end{itemize}

The \textit{train\_samples.json} should look like that:

\begin{minted}{json}
{
    "samples":[
        {
            "image_a":{
                "image":"base/cmp_b0001.jpg",
                "domain":"photo",
            },
            "image_b":{
                "image":"base/cmp_b0001.png",
                "domain":"mask",
            }
        },
        ...
    ]
}
\end{minted}

\vspace{5pt}
\noindent\textbf{\textit{2. Style Transfer (Unpaired)}}
Besides, CMP Facade dataset can also be used for unpaired styple transfer task, just import the unpaired style transfter task template into the Dataset YAML File.

\vspace{5pt}
\noindent\textbf{Task YAML File Definition}
\begin{minted}{yaml}
$dsdl-version: "0.5.0"

FacadeImageSample:
    $def: struct
    $params: ['cdom']
    $fields:
        image: Image
        domain: Label[dom=$cdom]    
\end{minted}

The fields in the yaml are explained below:
\begin{itemize}
    \item \texttt{\$dsdl-version}: Specifies the DSDL version of the file.
    \item \texttt{FacadeImageSample}: Defines the structure of an unparied style transfer sample, containing:
        \begin{itemize}
            \item \texttt{\$def}: Indicates a structure type.
            \item \texttt{\$params}: Defines parameters, here the class domain.
            \item \texttt{\$fields}: Attributes of the structure, including:
                \begin{itemize}
                    \item \texttt{image}: Path of the image.
                    \item \texttt{domain}: The domain of an image.
                \end{itemize}
        \end{itemize}
\end{itemize}

The \textit{train\_samples.json} should look like that:

\begin{minted}{json}
{
    "samples":[
        {
            "image":"base/cmp_b0001.jpg",
            "domain":"photo",
        },
        {
            "image":"base/cmp_b0001.png",
            "domain":"mask",
        },
        ...
    ]
}
\end{minted}

\addcontentsline{toc}{section}{References}
\printbibliography

@inproceedings{deng2009imagenet,
  title={Imagenet: A large-scale hierarchical image database},
  author={Deng, Jia and Dong, Wei and Socher, Richard and Li, Li-Jia and Li, Kai and Fei-Fei, Li},
  booktitle={2009 IEEE conference on computer vision and pattern recognition},
  pages={248--255},
  year={2009},
  organization={Ieee}
}

@inproceedings{lin2014microsoft,
  title={Microsoft coco: Common objects in context},
  author={Lin, Tsung-Yi and Maire, Michael and Belongie, Serge and Hays, James and Perona, Pietro and Ramanan, Deva and Doll{\'a}r, Piotr and Zitnick, C Lawrence},
  booktitle={Computer Vision--ECCV 2014: 13th European Conference, Zurich, Switzerland, September 6-12, 2014, Proceedings, Part V 13},
  pages={740--755},
  year={2014},
  organization={Springer}
}

@article{kuznetsova2020open,
  title={The open images dataset v4: Unified image classification, object detection, and visual relationship detection at scale},
  author={Kuznetsova, Alina and Rom, Hassan and Alldrin, Neil and Uijlings, Jasper and Krasin, Ivan and Pont-Tuset, Jordi and Kamali, Shahab and Popov, Stefan and Malloci, Matteo and Kolesnikov, Alexander and others},
  journal={International journal of computer vision},
  volume={128},
  number={7},
  pages={1956--1981},
  year={2020},
  publisher={Springer}
}

@inproceedings{gupta2019lvis,
  title={Lvis: A dataset for large vocabulary instance segmentation},
  author={Gupta, Agrim and Dollar, Piotr and Girshick, Ross},
  booktitle={Proceedings of the IEEE/CVF conference on computer vision and pattern recognition},
  pages={5356--5364},
  year={2019}
}

@article{schuhmann2022laion,
  title={Laion-5b: An open large-scale dataset for training next generation image-text models},
  author={Schuhmann, Christoph and Beaumont, Romain and Vencu, Richard and Gordon, Cade and Wightman, Ross and Cherti, Mehdi and Coombes, Theo and Katta, Aarush and Mullis, Clayton and Wortsman, Mitchell and others},
  journal={Advances in Neural Information Processing Systems},
  volume={35},
  pages={25278--25294},
  year={2022}
}

@article{raffel2020exploring,
  title={Exploring the limits of transfer learning with a unified text-to-text transformer},
  author={Raffel, Colin and Shazeer, Noam and Roberts, Adam and Lee, Katherine and Narang, Sharan and Matena, Michael and Zhou, Yanqi and Li, Wei and Liu, Peter J},
  journal={Journal of machine learning research},
  volume={21},
  number={140},
  pages={1--67},
  year={2020}
}

@article{lin2023parrot,
  title={Parrot Captions Teach CLIP to Spot Text},
  author={Lin, Yiqi and He, Conghui and Wang, Alex Jinpeng and Wang, Bin and Li, Weijia and Shou, Mike Zheng},
  journal={arXiv preprint arXiv:2312.14232},
  year={2023}
}

@inproceedings{changpinyo2021conceptual,
  title={Conceptual 12m: Pushing web-scale image-text pre-training to recognize long-tail visual concepts},
  author={Changpinyo, Soravit and Sharma, Piyush and Ding, Nan and Soricut, Radu},
  booktitle={Proceedings of the IEEE/CVF conference on computer vision and pattern recognition},
  pages={3558--3568},
  year={2021}
}

@article{wang2024unimernet,
  title={UniMERNet: A Universal Network for Real-World Mathematical Expression Recognition},
  author={Wang, Bin and Gu, Zhuangcheng and Xu, Chao and Zhang, Bo and Shi, Botian and He, Conghui},
  journal={arXiv preprint arXiv:2404.15254},
  year={2024}
}

@article{he2023wanjuan,
  title={Wanjuan: A comprehensive multimodal dataset for advancing english and chinese large models},
  author={He, Conghui and Jin, Zhenjiang and Xu, Chao and Qiu, Jiantao and Wang, Bin and Li, Wei and Yan, Hang and Wang, Jiaqi and Lin, Dahua},
  journal={arXiv preprint arXiv:2308.10755},
  year={2023}
}

@article{krizhevsky2017imagenet,
  title={ImageNet classification with deep convolutional neural networks},
  author={Krizhevsky, Alex and Sutskever, Ilya and Hinton, Geoffrey E},
  journal={Communications of the ACM},
  volume={60},
  number={6},
  pages={84--90},
  year={2017},
  publisher={AcM New York, NY, USA}
}

@inproceedings{he2016deep,
  title={Deep residual learning for image recognition},
  author={He, Kaiming and Zhang, Xiangyu and Ren, Shaoqing and Sun, Jian},
  booktitle={Proceedings of the IEEE conference on computer vision and pattern recognition},
  pages={770--778},
  year={2016}
}

@article{vaswani2017attention,
  title={Attention is all you need},
  author={Vaswani, Ashish and Shazeer, Noam and Parmar, Niki and Uszkoreit, Jakob and Jones, Llion and Gomez, Aidan N and Kaiser, {\L}ukasz and Polosukhin, Illia},
  journal={Advances in neural information processing systems},
  volume={30},
  year={2017}
}

@article{devlin2018bert,
  title={Bert: Pre-training of deep bidirectional transformers for language understanding},
  author={Devlin, Jacob and Chang, Ming-Wei and Lee, Kenton and Toutanova, Kristina},
  journal={arXiv preprint arXiv:1810.04805},
  year={2018}
}

@article{radford2018improving,
  title={Improving language understanding by generative pre-training},
  author={Radford, Alec and Narasimhan, Karthik and Salimans, Tim and Sutskever, Ilya and others},
  year={2018},
  publisher={OpenAI}
}

@article{radford2019language,
  title={Language models are unsupervised multitask learners},
  author={Radford, Alec and Wu, Jeffrey and Child, Rewon and Luan, David and Amodei, Dario and Sutskever, Ilya and others},
  journal={OpenAI blog},
  volume={1},
  number={8},
  pages={9},
  year={2019}
}

@article{brown2020language,
  title={Language models are few-shot learners},
  author={Brown, Tom and Mann, Benjamin and Ryder, Nick and Subbiah, Melanie and Kaplan, Jared D and Dhariwal, Prafulla and Neelakantan, Arvind and Shyam, Pranav and Sastry, Girish and Askell, Amanda and others},
  journal={Advances in neural information processing systems},
  volume={33},
  pages={1877--1901},
  year={2020}
}

@article{liu2024visual,
  title={Visual instruction tuning},
  author={Liu, Haotian and Li, Chunyuan and Wu, Qingyang and Lee, Yong Jae},
  journal={Advances in neural information processing systems},
  volume={36},
  year={2024}
}

@article{zhang2023internlm,
  title={Internlm-xcomposer: A vision-language large model for advanced text-image comprehension and composition},
  author={Zhang, Pan and Wang, Xiaoyi Dong Bin and Cao, Yuhang and Xu, Chao and Ouyang, Linke and Zhao, Zhiyuan and Ding, Shuangrui and Zhang, Songyang and Duan, Haodong and Yan, Hang and others},
  journal={arXiv preprint arXiv:2309.15112},
  year={2023}
}

@article{dong2024internlm,
  title={InternLM-XComposer2: Mastering free-form text-image composition and comprehension in vision-language large model},
  author={Dong, Xiaoyi and Zhang, Pan and Zang, Yuhang and Cao, Yuhang and Wang, Bin and Ouyang, Linke and Wei, Xilin and Zhang, Songyang and Duan, Haodong and Cao, Maosong and others},
  journal={arXiv preprint arXiv:2401.16420},
  year={2024}
}

@article{chen2024far,
  title={How far are we to gpt-4v? closing the gap to commercial multimodal models with open-source suites},
  author={Chen, Zhe and Wang, Weiyun and Tian, Hao and Ye, Shenglong and Gao, Zhangwei and Cui, Erfei and Tong, Wenwen and Hu, Kongzhi and Luo, Jiapeng and Ma, Zheng and others},
  journal={arXiv preprint arXiv:2404.16821},
  year={2024}
}

@inproceedings{wang2024vigc,
  title={Vigc: Visual instruction generation and correction},
  author={Wang, Bin and Wu, Fan and Han, Xiao and Peng, Jiahui and Zhong, Huaping and Zhang, Pan and Dong, Xiaoyi and Li, Weijia and Li, Wei and Wang, Jiaqi and others},
  booktitle={Proceedings of the AAAI Conference on Artificial Intelligence},
  volume={38},
  number={6},
  pages={5309--5317},
  year={2024}
}

@misc{conghui2022opendatalab,
  author={He, Conghui and Li, Wei and Jin, Zhenjiang and Wang, Bin and Xu, Chao and Lin, Dahua},
  title={OpenDataLab: Empowering General Artificial Intelligence with Open Datasets},
  howpublished = {\url{https://opendatalab.com}},
  year={2022},
  note = {Accessed: 2023-12-22}
}

@article{russakovsky2015imagenet,
  title={Imagenet large scale visual recognition challenge},
  author={Russakovsky, Olga and Deng, Jia and Su, Hao and Krause, Jonathan and Satheesh, Sanjeev and Ma, Sean and Huang, Zhiheng and Karpathy, Andrej and Khosla, Aditya and Bernstein, Michael and others},
  journal={International journal of computer vision},
  volume={115},
  pages={211--252},
  year={2015},
  publisher={Springer}
}

@article{lecun1998gradient,
title={Gradient-based learning applied to document recognition},
author={LeCun, Yann and Bottou, L{\'e}on and Bengio, Yoshua and Haffner, Patrick},
journal={Proceedings of the IEEE},
volume={86},
number={11},
pages={2278--2324},
year={1998},
publisher={Ieee}
}

@article{krizhevsky2009learning,
  title={Learning multiple layers of features from tiny images},
  author={Krizhevsky, Alex and Hinton, Geoffrey and others},
  year={2009},
  publisher={Toronto, ON, Canada}
}

@article{xiao2017fashion,
  title={Fashion-mnist: a novel image dataset for benchmarking machine learning algorithms},
  author={Xiao, Han and Rasul, Kashif and Vollgraf, Roland},
  journal={arXiv preprint arXiv:1708.07747},
  year={2017}
}

@article{griffin2007caltech,
  title={Caltech-256 object category dataset},
  author={Griffin, Gregory and Holub, Alex and Perona, Pietro},
  year={2007},
  publisher={California Institute of Technology}
}

@inproceedings{nilsback2008automated,
  title={Automated flower classification over a large number of classes},
  author={Nilsback, Maria-Elena and Zisserman, Andrew},
  booktitle={2008 Sixth Indian conference on computer vision, graphics \& image processing},
  pages={722--729},
  year={2008},
  organization={IEEE}
}

@article{everingham2011the,
      title={The pascal visual object classes challenge 2012 (voc2012) results (2012)},
      author={Everingham, M and Gool, Van L and Williams, CKI and Winn, J and Zisserman, A},
      year={2011}
}

@INPROCEEDINGS{Geiger2012CVPR,
author = {Andreas Geiger and Philip Lenz and Raquel Urtasun},
title = {Are we ready for Autonomous Driving? The KITTI Vision Benchmark Suite},
booktitle = {Conference on Computer Vision and Pattern        Recognition (CVPR)},
year = {2012}
}

@INPROCEEDINGS{9009553,
author={Shao, Shuai and Li, Zeming and Zhang, Tianyuan and Peng, Chao and Yu, Gang and Zhang, Xiangyu and Li, Jing and Sun, Jian},
booktitle={2019 IEEE/CVF International Conference on Computer Vision (ICCV)}, 
title={Objects365: A Large-Scale, High-Quality Dataset for Object Detection}, 
year={2019},
volume={},
number={},
pages={8429-8438},
doi={10.1109/ICCV.2019.00852}
}

@article{ILSVRC15,
Author = {Olga Russakovsky and Jia Deng and Hao Su and Jonathan Krause and Sanjeev Satheesh and Sean Ma and Zhiheng Huang and Andrej Karpathy and Aditya Khosla and Michael Bernstein and Alexander C. Berg and Li Fei-Fei},
Title = {{ImageNet Large Scale Visual Recognition Challenge}},
Year = {2015},
journal   = {International Journal of Computer Vision (IJCV)},
doi = {10.1007/s11263-015-0816-y},
volume={115},
number={3},
pages={211-252}
}

@inproceedings{cordts2016cityscapes,
  title={The cityscapes dataset for semantic urban scene understanding},
  author={Cordts, Marius and Omran, Mohamed and Ramos, Sebastian and Rehfeld, Timo and Enzweiler, Markus and Benenson, Rodrigo and Franke, Uwe and Roth, Stefan and Schiele, Bernt},
  booktitle={Proceedings of the IEEE conference on computer vision and pattern recognition},
  pages={3213--3223},
  year={2016}
}

@inproceedings{zhou2017scene,
  title={Scene parsing through ade20k dataset},
  author={Zhou, Bolei and Zhao, Hang and Puig, Xavier and Fidler, Sanja and Barriuso, Adela and Torralba, Antonio},
  booktitle={Proceedings of the IEEE conference on computer vision and pattern recognition},
  pages={633--641},
  year={2017}
}

@article{yu2015lsun,
  title={Lsun: Construction of a large-scale image dataset using deep learning with humans in the loop},
  author={Yu, Fisher and Seff, Ari and Zhang, Yinda and Song, Shuran and Funkhouser, Thomas and Xiao, Jianxiong},
  journal={arXiv preprint arXiv:1506.03365},
  year={2015}
}

@inproceedings{karras2019style,
  title={A style-based generator architecture for generative adversarial networks},
  author={Karras, Tero and Laine, Samuli and Aila, Timo},
  booktitle={Proceedings of the IEEE/CVF conference on computer vision and pattern recognition},
  pages={4401--4410},
  year={2019}
}

@inproceedings{andriluka20142d,
title={2d human pose estimation: New benchmark and state of the art analysis},
author={Andriluka, Mykhaylo and Pishchulin, Leonid and Gehler, Peter and Schiele, Bernt},
booktitle={Proceedings of the IEEE Conference on computer Vision and Pattern Recognition},
pages={3686--3693},
year={2014}
}

@article{karras2017progressive,
  title={Progressive growing of gans for improved quality, stability, and variation},
  author={Karras, Tero and Aila, Timo and Laine, Samuli and Lehtinen, Jaakko},
  journal={arXiv preprint arXiv:1710.10196},
  year={2017}
}

@inproceedings{duan2019trb,
  title={Trb: a novel triplet representation for understanding 2d human body},
  author={Duan, Haodong and Lin, Kwan-Yee and Jin, Sheng and Liu, Wentao and Qian, Chen and Ouyang, Wanli},
  booktitle={Proceedings of the IEEE/CVF international conference on computer vision},
  pages={9479--9488},
  year={2019}
}

@article{karras2020training,
title={Training generative adversarial networks with limited data},
author={Karras, Tero and Aittala, Miika and Hellsten, Janne and Laine, Samuli and Lehtinen, Jaakko and Aila, Timo},
journal={Advances in Neural Information Processing Systems},
volume={33},
pages={12104--12114},
year={2020}
}

@inproceedings{choi2020stargan,
title={Stargan v2: Diverse image synthesis for multiple domains},
author={Choi, Yunjey and Uh, Youngjung and Yoo, Jaejun and Ha, Jung-Woo},
booktitle={Proceedings of the IEEE/CVF conference on computer vision and pattern recognition},
pages={8188--8197},
year={2020}
}

@article{wu2017ai,
  title={Ai challenger: A large-scale dataset for going deeper in image understanding},
  author={Wu, Jiahong and Zheng, He and Zhao, Bo and Li, Yixin and Yan, Baoming and Liang, Rui and Wang, Wenjia and Zhou, Shipei and Lin, Guosen and Fu, Yanwei and others},
  journal={arXiv preprint arXiv:1711.06475},
  year={2017}
}

@inproceedings{li2019crowdpose,
  title={Crowdpose: Efficient crowded scenes pose estimation and a new benchmark},
  author={Li, Jiefeng and Wang, Can and Zhu, Hao and Mao, Yihuan and Fang, Hao-Shu and Lu, Cewu},
  booktitle={Proceedings of the IEEE/CVF conference on computer vision and pattern recognition},
  pages={10863--10872},
  year={2019}
}

@inproceedings{zhang2019pose2seg,
  title={Pose2seg: Detection free human instance segmentation},
  author={Zhang, Song-Hai and Li, Ruilong and Dong, Xin and Rosin, Paul and Cai, Zixi and Han, Xi and Yang, Dingcheng and Huang, Haozhi and Hu, Shi-Min},
  booktitle={Proceedings of the IEEE/CVF conference on computer vision and pattern recognition},
  pages={889--898},
  year={2019}
}

@inproceedings{jin2020whole,
title={Whole-body human pose estimation in the wild},
author={Jin, Sheng and Xu, Lumin and Xu, Jin and Wang, Can and Liu, Wentao and Qian, Chen and Ouyang, Wanli and Luo, Ping},
booktitle={European Conference on Computer Vision},
pages={196--214},
year={2020},
organization={Springer}
}

@inproceedings{zhu2017unpaired,
  title={Unpaired image-to-image translation using cycle-consistent adversarial networks},
  author={Zhu, Jun-Yan and Park, Taesung and Isola, Phillip and Efros, Alexei A},
  booktitle={Proceedings of the IEEE international conference on computer vision},
  pages={2223--2232},
  year={2017}
}

@inproceedings{zhu2016face,
  title={Face alignment across large poses: A 3d solution},
  author={Zhu, Xiangyu and Lei, Zhen and Liu, Xiaoming and Shi, Hailin and Li, Stan Z},
  booktitle={Proceedings of the IEEE conference on computer vision and pattern recognition},
  pages={146--155},
  year={2016}
}

@inproceedings{wu2018look,
title={Look at boundary: A boundary-aware face alignment algorithm},
author={Wu, Wayne and Qian, Chen and Yang, Shuo and Wang, Quan and Cai, Yici and Zhou, Qiang},
booktitle={Proceedings of the IEEE conference on computer vision and pattern recognition},
pages={2129--2138},
year={2018}
}

@inproceedings{zhu2016unconstrained,
title={Unconstrained face alignment via cascaded compositional learning},
author={Zhu, Shizhan and Li, Cheng and Loy, Chen-Change and Tang, Xiaoou},
booktitle={Proceedings of the IEEE Conference on Computer Vision and Pattern Recognition},
pages={3409--3417},
year={2016}
}

@inproceedings{burgos2013robust,
title={Robust face landmark estimation under occlusion},
author={Burgos-Artizzu, Xavier P and Perona, Pietro and Doll{\'a}r, Piotr},
booktitle={Proceedings of the IEEE international conference on computer vision},
pages={1513--1520},
year={2013}
}

@article{wang2018mask,
title={Mask-pose cascaded cnn for 2d hand pose estimation from single color image},
author={Wang, Yangang and Peng, Cong and Liu, Yebin},
journal={IEEE Transactions on Circuits and Systems for Video Technology},
volume={29},
number={11},
pages={3258--3268},
year={2018},
publisher={IEEE}
}

@inproceedings{zimmermann2019freihand,
title={Freihand: A dataset for markerless capture of hand pose and shape from single rgb images},
author={Zimmermann, Christian and Ceylan, Duygu and Yang, Jimei and Russell, Bryan and Argus, Max and Brox, Thomas},
booktitle={Proceedings of the IEEE/CVF International Conference on Computer Vision},
pages={813--822},
year={2019}
}

@inproceedings{simon2017hand,
  title={Hand keypoint detection in single images using multiview bootstrapping},
  author={Simon, Tomas and Joo, Hanbyul and Matthews, Iain and Sheikh, Yaser},
  booktitle={Proceedings of the IEEE conference on Computer Vision and Pattern Recognition},
  pages={1145--1153},
  year={2017}
}

@article{jiang2022text2human,
  title={Text2human: Text-driven controllable human image generation},
  author={Jiang, Yuming and Yang, Shuai and Qiu, Haonan and Wu, Wayne and Loy, Chen Change and Liu, Ziwei},
  journal={ACM Transactions on Graphics (TOG)},
  volume={41},
  number={4},
  pages={1--11},
  year={2022},
  publisher={ACM New York, NY, USA}
}

@inproceedings{cao2019cross,
title={Cross-domain adaptation for animal pose estimation},
author={Cao, Jinkun and Tang, Hongyang and Fang, Hao-Shu and Shen, Xiaoyong and Lu, Cewu and Tai, Yu-Wing},
booktitle={Proceedings of the IEEE/CVF International Conference on Computer Vision},
pages={9498--9507},
year={2019}
}

@article{yu2021ap,
title={AP-10K: A Benchmark for Animal Pose Estimation in the Wild},
author={Yu, Hang and Xu, Yufei and Zhang, Jing and Zhao, Wei and Guan, Ziyu and Tao, Dacheng},
journal={arXiv preprint arXiv:2108.12617},
year={2021}
}

@inproceedings{mathis2021pretraining,
title={Pretraining boosts out-of-domain robustness for pose estimation},
author={Mathis, Alexander and Biasi, Thomas and Schneider, Steffen and Yuksekgonul, Mert and Rogers, Byron and Bethge, Matthias and Mathis, Mackenzie W},
booktitle={Proceedings of the IEEE/CVF Winter Conference on Applications of Computer Vision},
pages={1859--1868},
year={2021}
}

@article{labuguen2021macaquepose,
  title={MacaquePose: a novel “in the wild” macaque monkey pose dataset for markerless motion capture},
  author={Labuguen, Rollyn and Matsumoto, Jumpei and Negrete, Salvador Blanco and Nishimaru, Hiroshi and Nishijo, Hisao and Takada, Masahiko and Go, Yasuhiro and Inoue, Ken-ichi and Shibata, Tomohiro},
  journal={Frontiers in behavioral neuroscience},
  volume={14},
  pages={581154},
  year={2021},
  publisher={Frontiers Media SA}
}

@article{graving2019fast, 
title={Fast and robust animal pose estimation}, 
author={Graving, Jacob M and Chae, Daniel and Naik, Hemal and Li, Liang and Koger, Benjamin and Costelloe, Blair R and Couzin, Iain D}, 
journal={bioRxiv}, 
pages={620245}, 
year={2019}, 
publisher={Cold Spring Harbor Laboratory} 
}

@inproceedings{fang2017rmpe,
  title={Rmpe: Regional multi-person pose estimation},
  author={Fang, Hao-Shu and Xie, Shuqin and Tai, Yu-Wing and Lu, Cewu},
  booktitle={Proceedings of the IEEE international conference on computer vision},
  pages={2334--2343},
  year={2017}
}

@inproceedings{muller2018trackingnet,
  title={Trackingnet: A large-scale dataset and benchmark for object tracking in the wild},
  author={Muller, Matthias and Bibi, Adel and Giancola, Silvio and Alsubaihi, Salman and Ghanem, Bernard},
  booktitle={Proceedings of the European Conference on Computer Vision (ECCV)},
  pages={300--317},
  year={2018}
}

@ARTICLE{Huang2021,
author={Huang, Lianghua and Zhao, Xin and Huang, Kaiqi},
journal={IEEE Transactions on Pattern Analysis and Machine Intelligence}, 
title={GOT-10k: A Large High-Diversity Benchmark for Generic Object Tracking in the Wild}, 
year={2021},
volume={43},
number={5},
pages={1562-1577},
doi={10.1109/TPAMI.2019.2957464}
}

@article{milan2016mot16,
title={MOT16: A benchmark for multi-object tracking},
author={Milan, Anton and Leal-Taix{\'e}, Laura and Reid, Ian and Roth, Stefan and Schindler, Konrad},
journal={arXiv preprint arXiv:1603.00831},
year={2016}
}

@misc{ding2021object,
title={Object Detection in Aerial Images: A Large-Scale Benchmark and Challenges},
author={Jian Ding and Nan Xue and Gui-Song Xia and Xiang Bai and Wen Yang and Micheal Ying Yang and Serge Belongie and Jiebo Luo and Mihai Datcu and Marcello Pelillo and Liangpei Zhang},
year={2021},
eprint={2102.12219},
archivePrefix={arXiv},
primaryClass={cs.CV}
}

@inproceedings{liu2017high,
  title={A high resolution optical satellite image dataset for ship recognition and some new baselines},
  author={Liu, Zikun and Yuan, Liu and Weng, Lubin and Yang, Yiping},
  booktitle={International conference on pattern recognition applications and methods},
  volume={2},
  pages={324--331},
  year={2017},
  organization={SciTePress}
}

@inproceedings{zhu2015orientation,
  title={Orientation robust object detection in aerial images using deep convolutional neural network},
  author={Zhu, Haigang and Chen, Xiaogang and Dai, Weiqun and Fu, Kun and Ye, Qixiang and Jiao, Jianbin},
  booktitle={2015 IEEE International Conference on Image Processing (ICIP)},
  pages={3735--3739},
  year={2015},
  organization={IEEE}
}

@article{zhang2021sar,
  title={SAR ship detection dataset (SSDD): Official release and comprehensive data analysis},
  author={Zhang, Tianwen and Zhang, Xiaoling and Li, Jianwei and Xu, Xiaowo and Wang, Baoyou and Zhan, Xu and Xu, Yanqin and Ke, Xiao and Zeng, Tianjiao and Su, Hao and others},
  journal={Remote Sensing},
  volume={13},
  number={18},
  pages={3690},
  year={2021},
  publisher={MDPI}
}

@inproceedings{karatzas2013icdar,
title={ICDAR 2013 robust reading competition},
author={Karatzas, Dimosthenis and Shafait, Faisal and Uchida, Seiichi and Iwamura, Masakazu and i Bigorda, Lluis Gomez and Mestre, Sergi Robles and Mas, Joan and Mota, David Fernandez and Almazan, Jon Almazan and De Las Heras, Lluis Pere},
booktitle={2013 12th international conference on document analysis and recognition},
pages={1484--1493},
year={2013},
organization={IEEE}
}

@inproceedings{karatzas2015icdar,
title={ICDAR 2015 competition on robust reading},
author={Karatzas, Dimosthenis and Gomez-Bigorda, Lluis and Nicolaou, Anguelos and Ghosh, Suman and Bagdanov, Andrew and Iwamura, Masakazu and Matas, Jiri and Neumann, Lukas and Chandrasekhar, Vijay Ramaseshan and Lu, Shijian and others},
booktitle={2015 13th international conference on document analysis and recognition (ICDAR)},
pages={1156--1160},
year={2015},
organization={IEEE}
}

@inproceedings{wang2011end,
title={End-to-end scene text recognition},
author={Wang, Kai and Babenko, Boris and Belongie, Serge},
booktitle={2011 International conference on computer vision},
pages={1457--1464},
year={2011},
organization={IEEE}
}

@inproceedings{ch2017total,
title={Total-text: A comprehensive dataset for scene text detection and recognition},
author={Ch'ng, Chee Kheng and Chan, Chee Seng},
booktitle={2017 14th IAPR international conference on document analysis and recognition (ICDAR)},
volume={1},
pages={935--942},
year={2017},
organization={IEEE}
}

@inproceedings{yao2012detecting,
title={Detecting texts of arbitrary orientations in natural images},
author={Yao, Cong and Bai, Xiang and Liu, Wenyu and Ma, Yi and Tu, Zhuowen},
booktitle={2012 IEEE conference on computer vision and pattern recognition},
pages={1083--1090},
year={2012},
organization={IEEE}
}

@article{risnumawan2014robust,
  title={A robust arbitrary text detection system for natural scene images},
  author={Risnumawan, Anhar and Shivakumara, Palaiahankote and Chan, Chee Seng and Tan, Chew Lim},
  journal={Expert Systems with Applications},
  volume={41},
  number={18},
  pages={8027--8048},
  year={2014},
  publisher={Elsevier}
}

@inproceedings{mishra2012scene,
  title={Scene text recognition using higher order language priors},
  author={Mishra, Anand and Alahari, Karteek and Jawahar, CV},
  booktitle={BMVC-British machine vision conference},
  year={2012},
  organization={BMVA}
}


\end{document}